\documentclass{article}

\usepackage[main, final]{neurips_2026}
\usepackage{tcolorbox}

\usepackage[utf8]{inputenc} 
\usepackage[T1]{fontenc}    
\usepackage{hyperref}       
\usepackage{url}            
\usepackage{booktabs}       
\usepackage{amsfonts}       
\usepackage{nicefrac}       
\usepackage{microtype}      
\usepackage{xcolor}         
\usepackage[utf8]{inputenc} 
\usepackage[T1]{fontenc}    
\usepackage{hyperref}       
\usepackage{url}            
\usepackage{booktabs}       
\usepackage{amsfonts}       
\usepackage{nicefrac}       
\usepackage{microtype}      
\usepackage{xcolor}         
\usepackage[utf8]{inputenc}
\usepackage{algorithm}
\usepackage{algorithmic}
\usepackage{enumitem}
\usepackage{amsthm}
\newtheorem{remark}{Remark}
\usepackage{caption}
\usepackage{graphicx}
\usepackage{multirow}
\usepackage{subcaption}
\usepackage{amsmath}
\usepackage{dsfont}
\usepackage[table]{xcolor} 
\usepackage{arydshln} 
\usepackage{array}
\usepackage{amssymb}
\usepackage{pifont}            
\usepackage{makecell}

\newtheorem{theorem}{Theorem}[section]
\newtheorem{lemma}[theorem]{Lemma}

\newtheorem{definition}[theorem]{Definition}
\newtheorem{assumption}{Assumption}[section]

\newcommand\blfootnote[1]{%
  \begingroup
  \renewcommand\thefootnote{}\footnote{#1}%
  \addtocounter{footnote}{-1}%
  \endgroup
}

\definecolor{mathbg}{HTML}{E8F0FE}    
\definecolor{genbg}{HTML}{E6F4EA}     
\definecolor{allbg}{HTML}{FCE8E6}     
\definecolor{avgbg}{HTML}{F2F2F2}     

\title{Not All Tokens Should Be Treated Equally:\\ Context Credits Reassignment}

\author{%
  \textbf{Shujian Gao}\textsuperscript{1,2,*} \quad
  \textbf{Yuan Wang}\textsuperscript{4,*} \quad
  \textbf{Jiamei Yan}\textsuperscript{3} \quad
  \textbf{Penghao Zhou}\textsuperscript{3,\textdaggerdbl} \\
  \textbf{Qinglei Wang}\textsuperscript{3,\textdaggerdbl} \quad
  \textbf{Jiangtao Yan}\textsuperscript{5} \quad
  \textbf{Zuxuan Wu}\textsuperscript{1,2,\textdagger} \quad
  \textbf{Yu-Gang Jiang}\textsuperscript{1,\textdagger} \\
  \\
  \textsuperscript{1}Institute of Trustworthy Embodied AI, Fudan University \\
  \textsuperscript{2}Shanghai Innovation Institute \quad
  \textsuperscript{3}ByteDance \\
  \textsuperscript{4}Zhejiang University \quad
  \textsuperscript{5}Wuhan University \\
}

\begin{document}

\maketitle

\blfootnote{\textsuperscript{*}Equal contribution.\quad
            \textsuperscript{\textdaggerdbl}Project lead.\quad
            \textsuperscript{\textdagger}Corresponding authors.}

{\centering
\small Project page: \url{https://daryl-gsj.github.io/HiCCR/}\par
}

\begin{abstract}
Multimodal Reinforcement learning with verifiable rewards (RLVR) algorithms assign the same advantage to every token in a generated sequence, providing no differential signal for tokens with varying degrees of dependence on visual evidence. 
A simple experiment confirms this issue: \textbf{scaling token advantages by uninformative random noise already outperforms uniform credit}, suggesting that uniform credit is empirically suboptimal.
Beyond that, we observe RLVR tuning under uniform credit weakens visual grounding along three complementary axes: attention, gradient attribution, and functional dependence on image embeddings, \textbf{collectively indicating insufficient utilization of visual evidence}.
To understand this degradation, we provide a theoretical analysis showing that, under language-prior dominance and regularity conditions, uniform credit can amplify a pretrained text-favored gradient imbalance, contributing to weaker visual attention.
In response, we propose Hierarchical Context Credit Reassignment (\textbf{HiCCR}), which reassigns credit at two levels: a token-level weight amplifies the advantage for visually grounded tokens, and a trajectory-level weight upweights rollouts with stronger visual engagement. 
The entire mechanism \textbf{adds less than 3\% wall-clock overhead} and applies to mainstream RLVR algorithms without auxiliary models or additional data. 
On ten benchmarks spanning mathematical and general multimodal reasoning, HiCCR consistently improves upon its corresponding base algorithm and achieves state-of-the-art results among open-source models of comparable scale, while alleviating the weakening of visual grounding observed under RLVR training. 
\end{abstract}

\section{Introduction}
\label{sec:intro}


Reinforcement learning with verifiable rewards (RLVR) is widely recognized as an effective paradigm for improving reasoning in Multimodal Large Language Models (MLLMs)~\citep{chen2025r1v, visionr1, vppo_spotlight, racro}. 
Existing methods rely on a uniform token credit assumption where a sequence reward is distributed equally across all generated tokens~\citep{grpo, dapo}. 
However, multimodal reasoning inherently builds on the heterogeneous integration of visual evidence and textual priors~\citep{avar_narrow, Kong2025TokenRS}. 
Forcing all tokens to share a single scalar advantage $\hat{A}$ directly contradicts the intuition that token contributions differ significantly in multimodal tasks. 
This simplification fails to align with the mechanism of multimodal reasoning, as many tokens must extract information from the visual context instead of relying on language priors~\citep{tor-grpo, demystifyingllm}. \textit{Thus, it naturally raises the question: should all tokens in a trajectory really receive the same credit in multimodal RLVR?}

A simple experiment illustrates that uniform credit is overly coarse and empirically suboptimal. When we scale token advantages by \textbf{uninformative random noise} that carries no information about token quality, the resulting model already outperforms the base RLVR algorithm (Table~\ref{tab:table4_credit_assignment}). That is, \textit{even a marginal perturbation of uniformity yields gains over the default assignment. }

\begin{figure}[t]
    \centering
    \includegraphics[width=\linewidth]{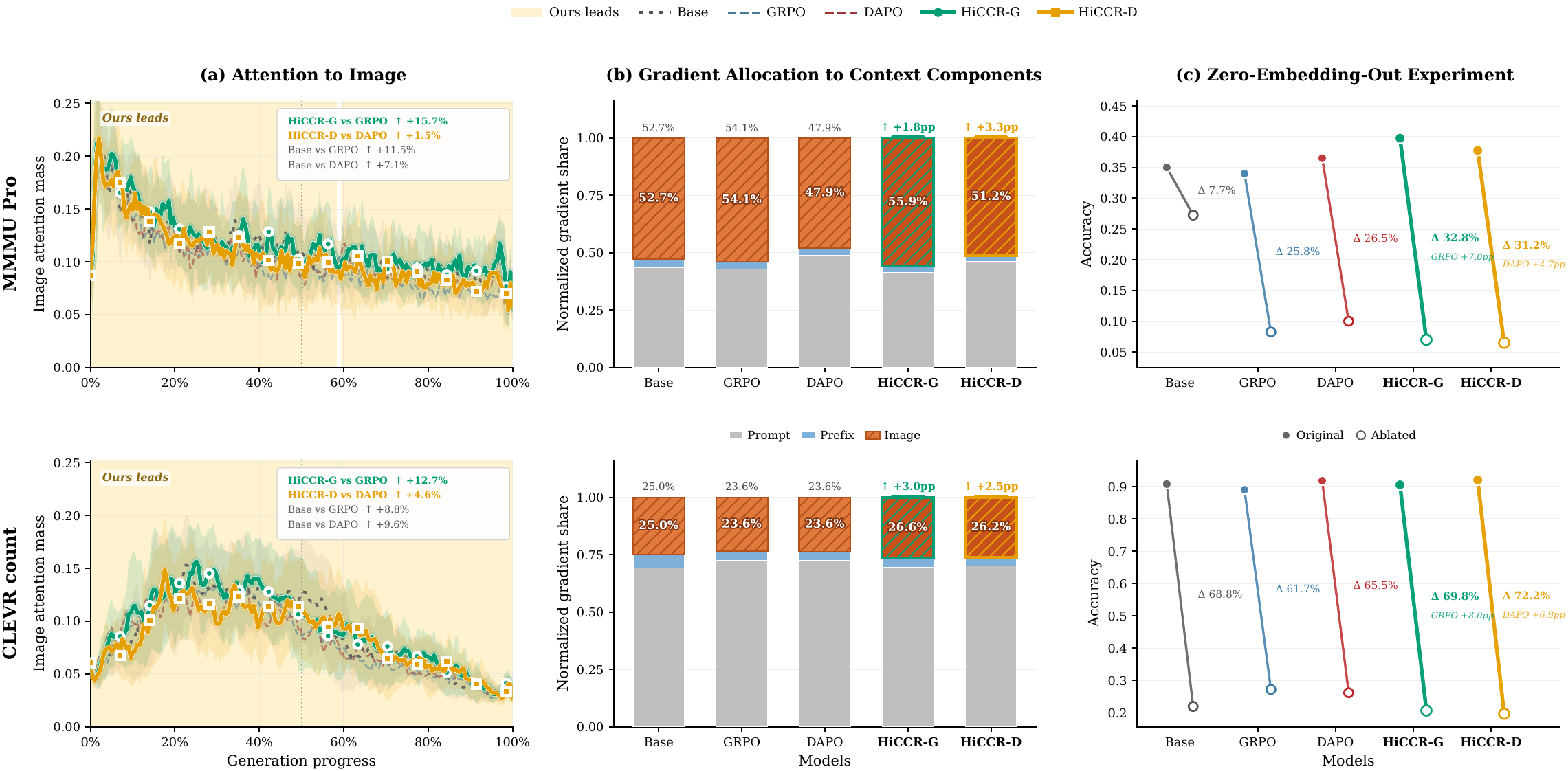}
    \caption{\textbf{RLVR fine-tuning weakens visual grounding in MLLMs, and HiCCR mitigates it.}}
    \label{fig:teaser}
\end{figure}
    
We begin by answering this question empirically.
Figure~\ref{fig:teaser} tracks three independent signatures of visual grounding on the same MLLM backbone~\citep{qwen3vl} before and after RLVR fine-tuning.
Panel~(a) tracks attention to image tokens across generation: the base model sustains higher image attention than GRPO and DAPO throughout, and the gap widens as generation proceeds.
Panel~(b) quantifies the gradient-level consequence via Input$\times$Gradient attribution: RLVR reduces rather than increases the gradient share routed to image tokens.
Panel~(c) probes functional dependence via a zero-embedding-out ablation: zeroing out image embeddings hurts RLVR baselines less than the base model, suggesting weaker dependence on visual input.
All three axes converge on the same finding: \textit{RLVR can improve answer accuracy while weakening the visual grounding that should underpin multimodal reasoning (full results in Appendix~\ref{app:empirical_grounding}).}

Why does uniform credit produce this effect? We provide a theoretical analysis offering a structural explanation.
Under a mild language-prior dominance condition (Appendix~\ref{sec:pg_theory}), we establish two results that answer this question.
\textbf{Theorem~\ref{thm:grad_bias}} shows that uniform positive-advantage assignment preserves and aggregates an existing text-favored local gradient imbalance, rather than correcting it: the expected policy-gradient magnitude through textual context exceeds that through visual context.
\textbf{Theorem~\ref{thm:drift}} shows that this per-step bias accumulates into a progressive drift of attention away from image tokens over generation, matching the trend in Figure~\ref{fig:teaser}(a).
Because language-prior dominance is a property of the pretrained model~\citep{demystifyingllm}, the bias is already present before fine-tuning, as reflected by the gradient gap in Figure~\ref{fig:teaser}(b); \textit{RLVR amplifies rather than creates it by reinforcing tokens without distinguishing their degree of visual grounding.}

Motivated by above analysis, we propose Hierarchical Context Credit Reassignment (\textbf{HiCCR}), a simple method that reallocates credit at two complementary levels. At the \textbf{token level}, HiCCR increases the effective learning signal for tokens that appear more visually engaged; at the \textbf{trajectory level}, it upweights rollouts whose overall generation exhibits stronger visual engagement. To instantiate this idea efficiently, we use the concentration of a token's visual attention as a lightweight forward-pass proxy, avoiding auxiliary models, additional supervision, or extra backward passes. HiCCR is designed as a drop-in modification to existing RLVR algorithms, and adds less than $3\%$ wall-clock overhead in our implementation.
Compared with task-agnostic heuristics~\citep{beyond28}, applying our formula to the text context, and raw attention energy (Table~\ref{tab:table4_credit_assignment}), concentration-based proxy improves both backbones and base algorithms, identifying the operative signal as concentration within the visual subset, not non-uniformity, not arbitrary context, not attention magnitude.

Together, our strategy closes the loop opened by our diagnosis: as shown in Figure~\ref{fig:teaser}, HiCCR mitigates attention drift, increases the gradient share routed to visual tokens, and yields stronger counterfactual evidence of visual dependence (Appendix.~\ref{app:empirical_grounding}).
Our findings suggest effective multimodal RLVR requires revisiting token-level credit assignment, and HiCCR provides a direct instantiation by reallocating outcome rewards toward visually engaged tokens and trajectories.
Our contributions are:

\begin{itemize}[leftmargin=1em,itemsep=2pt]
    \item \textbf{Finding.}
    We identify a visual grounding degradation in RLVR fine-tuned MLLMs on three independent axes (attention tracking, gradient attribution, and counterfactual visual ablation), already present in the pretrained base model and further amplified by RLVR (Appendix~\ref{app:empirical_grounding}).

    \item \textbf{Justification.}
    We attribute this finding to a structural interaction between uniform credit assignment and pretrained language-prior dominance: uniform advantage induces a systematic gradient bias favoring text over vision (Theorem~\ref{thm:grad_bias}), which compounds into visual-attention drift (Theorem~\ref{thm:drift}), explaining both the base-model gap and its amplification after RLVR training.

    \item \textbf{HiCCR.}
    We propose a hierarchical credit-reassignment method that mitigates the imbalance at both token and trajectory levels, using visual attention concentration as a lightweight proxy for visual dependence.

    \item \textbf{Strong results.}
    Applied to GRPO and DAPO on Qwen3-VL across ten multimodal reasoning benchmarks, HiCCR consistently improves its base algorithm and matches or exceeds open-source models several times its scale. Additional experiments on InternVL3~\citep{zhu2025internvl3} and the smaller Geo3K~\citep{geo3k} split further verify robustness across a different architecture and a lower-data regime.
\end{itemize}

\section{Method}
\label{sec:method}

We first formalize the token-level credit assignment problem in RLVR and then present \textbf{HiCCR}, a hierarchical framework that reassigns credit at both the token and trajectory levels to better align policy updates with visual dependence in multimodal reasoning.

\subsection{Preliminaries: Token-Level Credit Assignment in RLVR}
\label{sec:prelim}

Consider an MLLM parameterized by $\theta$ that generates a response $\mathbf{y}=(y_1,\dots,y_T)$ given a multimodal prompt $\mathbf{x}$ consisting of textual tokens $\mathbf{x}^{\text{text}}$ and visual tokens $\mathbf{x}^{\text{vis}}$. In RLVR, each complete trajectory $(\mathbf{x}, \mathbf{y})$ receives a scalar reward $R(\mathbf{x}, \mathbf{y}) \in \{0,1\}$ from a deterministic verifier.

Methods such as GRPO~\citep{grpo} and DAPO~\citep{dapo} optimize the policy using a trajectory-level advantage computed over a group of sampled responses $\{\mathbf{y}^{(i)}\}_{i=1}^{G}$:
\begin{equation}
    \hat{A}^{(i)} = \frac{R(\mathbf{x}, \mathbf{y}^{(i)}) - \mu_R}{\sigma_R + \epsilon},
    \label{eq:group_adv}
\end{equation}
where $\mu_R$ and $\sigma_R$ denote the mean and standard deviation of group rewards. The clipped objective is
\begin{equation}
    \mathcal{L}(\theta) = -\frac{1}{T} \sum_{t=1}^{T} \min\!\Big( r_t(\theta)\, \hat{A},\;\; \text{clip}\big(r_t(\theta), 1{-}\epsilon, 1{+}\epsilon\big)\, \hat{A} \Big),
    \label{eq:ppo_loss}
\end{equation}
where $r_t(\theta) = \pi_\theta(y_t \mid \mathbf{x}, \mathbf{y}_{<t}) \,/\, \pi_{\theta_{\text{old}}}(y_t \mid \mathbf{x}, \mathbf{y}_{<t})$ is the per-token importance ratio.

\paragraph{The uniform credit assignment problem.}
Equations~\eqref{eq:group_adv}--\eqref{eq:ppo_loss} imply that every token $y_t$ in a trajectory receives the same advantage signal $\hat{A}^{(i)}$. This uniform assignment treats all tokens as equally responsible for the trajectory outcome, even though, in multimodal reasoning, tokens may differ substantially in their reliance on visual evidence.

To make this explicit, the policy gradient for a single token takes the form
\begin{equation}
    \nabla_\theta \mathcal{L}_t \propto \hat{A}\,\nabla_\theta \log \pi_\theta(y_t \mid \mathbf{x}, \mathbf{y}_{<t}).
    \label{eq:pg_token}
\end{equation}
Under standard RLVR, the same scalar advantage modulates every token update regardless of the degree of visual dependence. This mismatch motivates the need for a credit assignment mechanism that better reflects visual dependence in multimodal reasoning.

\begin{figure}[t]
    \centering
    \includegraphics[width=\textwidth]{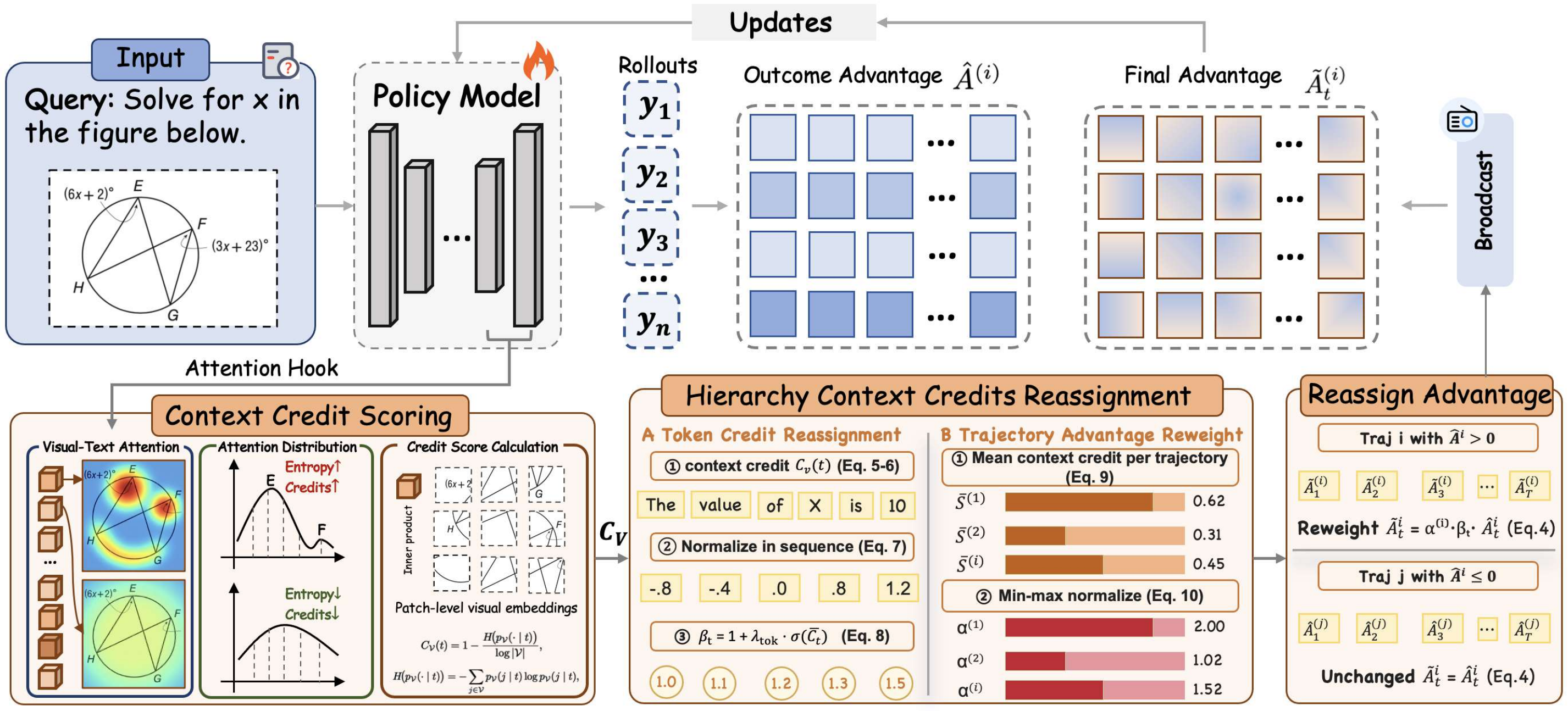}
\caption{\textbf{Overview of HiCCR.}
An attention hook on the policy model produces a per-token context credit $C_V(t)$ from intra-visual attention entropy.
Token- and trajectory-level weights $\beta_t,\alpha^{(i)}$ are derived from $C_V$ and reassign the outcome advantage as $\tilde A^{(i)}_t=\alpha^{(i)}\beta_t\hat A^{(i)}$.
The reassignment is applied only to positive-advantage trajectories.}
    \label{fig:pepiline}
\end{figure}

\subsection{HiCCR: Hierarchical Context Credit Reassignment}
\label{sec:hiccr}

To address this token-level credit mismatch, we propose HiCCR, which augments the standard RLVR pipeline with two complementary mechanisms operating at different granularities:
\begin{enumerate}
    \item \textbf{Token Credit Reassignment} (\S\ref{sec:token_credit}), a per-token weight $\beta_t$ that amplifies the policy gradient for tokens with stronger estimated dependence on visual context.
    \item \textbf{Trajectory Advantage Reweighting} (\S\ref{sec:traj_reweight}), a per-trajectory scaling factor $\alpha$ that preferentially reinforces trajectories exhibiting stronger overall visual engagement.
\end{enumerate}
Both mechanisms derive their signals from a lightweight \emph{context credit score} (\S\ref{sec:credit_score}) computed from the internal attention patterns of the model, without requiring an additional reward model, value network, or external supervision. The reassigned advantage is
\begin{equation}
    \tilde{A}_t^{(i)} = 
    \begin{cases}
        \alpha^{(i)} \cdot \beta_t^{(i)} \cdot \hat{A}^{(i)}, & \text{if } \hat{A}^{(i)} > 0, \\[4pt]
        \hat{A}^{(i)}, & \text{otherwise},
    \end{cases}
    \label{eq:hiccr_adv}
\end{equation}
where reassignment is applied only to positive-advantage trajectories. For correct trajectories, the goal is to differentially reinforce tokens and trajectories that exhibit stronger reliance on visual evidence. For incorrect trajectories, the negative advantage already provides a uniform discouragement signal, and further modulation is unnecessary. A discussion of asymmetric shaping is provided in Table~\ref{tab:asymmetric_shaping}.

\subsubsection{Context Credit Score}
\label{sec:credit_score}

The foundation of both mechanisms is a per-token \emph{context credit score} that serves as a lightweight proxy for the degree to which the generation of token $y_t$ depends on visual context. We define this score as visual information concentration, namely the complement of the normalized entropy of the attention distribution restricted to visual positions.

\paragraph{Setup.}
For a target layer $\ell \in \mathcal{L}_{\text{target}}$, let $\mathbf{q}_t^{(h)}, \mathbf{k}_j^{(h)} \in \mathbb{R}^{d}$ denote the query and key vectors at positions $t$ and $j$ for head $h$, and let $\mathcal{V} \subset \{1, \dots, |\mathbf{x}|\}$ be the set of visual token positions. We compute the intra-visual attention distribution by restricting the softmax to visual positions only:
\begin{equation}
    p_{\mathcal{V}}(j \mid t, h) = \frac{\exp\!\big(e_{t,j}^{(h)} / \tau\big)}{\sum_{j' \in \mathcal{V}} \exp\!\big(e_{t,j'}^{(h)} / \tau\big)}, \qquad e_{t,j}^{(h)} = \frac{\mathbf{q}_t^{(h)\top} \mathbf{k}_j^{(h)}}{\sqrt{d}},
    \label{eq:intra_vis_attn}
\end{equation}
where $\tau > 0$ is a temperature parameter. The resulting context credit score is
\begin{equation}
    C_{\mathcal{V}}(t) = 1 - \frac{H\!\big(p_{\mathcal{V}}(\cdot \mid t)\big)}{\log |\mathcal{V}|}, \qquad H\!\big(p_{\mathcal{V}}(\cdot \mid t)\big) = -\sum_{j \in \mathcal{V}} p_{\mathcal{V}}(j \mid t) \log p_{\mathcal{V}}(j \mid t),
    \label{eq:concentration}
\end{equation}
averaged over heads $h$ and layers $\ell$, with indices omitted for clarity. Dividing by $\log |\mathcal{V}|$, the maximum entropy of a uniform distribution over $|\mathcal{V}|$ positions, normalizes $C_{\mathcal{V}}(t)$ to the range $[0,1]$ regardless of the number of visual tokens.

This definition admits a clear interpretation. A large value of $C_{\mathcal{V}}(t)$ indicates that token $t$ concentrates its visual attention on a small subset of image positions, whereas a small value indicates a more diffuse attention pattern~\citep{zhai2023stabilizing}. Because the softmax in Eq.~\eqref{eq:intra_vis_attn} is computed only over visual positions, $C_{\mathcal{V}}(t)$ is invariant to global additive shifts in the energy scale across layers and training stages~\citep{liu2026sinktrack}.

\subsubsection{Token Credit Reassignment (TCR)}
\label{sec:token_credit}

Given the context credit scores $\{C_{\mathcal{V}}(t)\}_{t=1}^{T}$ for a trajectory, we normalize them to zero mean and unit variance within each sequence:
\begin{equation}
    \bar{c}_t = \frac{C_{\mathcal{V}}(t) - \mu_c}{\sigma_c + \epsilon},
    \label{eq:norm_score}
\end{equation}
where $\mu_c$ and $\sigma_c$ are the per-sequence mean and standard deviation. The resulting token-level weight is
\begin{equation}
    \beta_t = 1 + \lambda_{\text{tok}} \cdot \sigma(\bar{c}_t),
    \label{eq:beta}
\end{equation}
where $\sigma(\cdot)$ denotes the sigmoid function and $\lambda_{\text{tok}} > 0$ controls the maximum amplification. Since $\sigma(\bar{c}_t) \in (0, 1)$, the weight satisfies $\beta_t \in (1,\; 1 + \lambda_{\text{tok}})$. Tokens with above-average visual concentration therefore receive stronger reinforcement, while those with below-average concentration remain close to unity. No token is suppressed. The mechanism does not reverse the sign of any per-token update; it only modulates the relative magnitude of positive reinforcement.

\subsubsection{Trajectory Advantage Reweighting (TAR)}
\label{sec:traj_reweight}

Token credit reassignment operates within a single trajectory, whereas trajectory reweighting provides a complementary signal across trajectories. Among correct trajectories, those with stronger visual engagement are more likely to reflect solutions that make substantive use of visual evidence and should therefore receive proportionally larger reinforcement.

For each trajectory $i$, we compute the mean context credit score:
\begin{equation}
    \bar{S}^{(i)} = \frac{1}{T_i} \sum_{t=1}^{T_i} C_{\mathcal{V}}^{(i)}(t).
    \label{eq:traj_score}
\end{equation}
The trajectory-level weight is obtained by min-max normalization across the current batch:
\begin{equation}
    \alpha^{(i)} = 1 + \frac{\bar{S}^{(i)} - \min_k \bar{S}^{(k)}}{\max_k \bar{S}^{(k)} - \min_k \bar{S}^{(k)} + \epsilon},
    \label{eq:alpha}
\end{equation}
Under this normalization, the reweighting is strictly non-diminishing, which helps preserve the stability of the underlying RL objective. The procedure and design principles are summarized in~\ref{app:algorithm}.



\begin{table*}[t]
\caption{Performance on Qwen3-VL-4B and 8B across multimodal reasoning benchmarks. Within each backbone we place the \textbf{Ours vs.\ matched base RL} pairs at the bottom. Best in \textbf{bold}.}
\label{tab:table1_rlvr}
\centering
\small
\setlength{\tabcolsep}{3.6pt}
\renewcommand{\arraystretch}{1.00}

\definecolor{ourbg}{HTML}{EDF2FB}    
\definecolor{pairbg}{HTML}{F7F7F2}   
\definecolor{sechead}{HTML}{E9ECEF}  
\definecolor{ctxhead}{HTML}{FAFAFA}  

\newcommand{\up}[1]{\,{\scriptsize\textcolor{red!80!black}{(+#1)}}}
\newcommand{\upb}[1]{\,{\footnotesize\textcolor{red!80!black}{\textbf{(+#1)}}}}
\newcommand{\dn}[1]{\,{\scriptsize\textcolor{blue!65!black}{($-$#1)}}}

\resizebox{\textwidth}{!}{%
\begin{tabular}{@{}l ccccc c cccc c c@{}}
\toprule
\multirow{2.5}{*}{\textbf{Method}} &
\multicolumn{6}{c}{\textbf{Mathematical Reasoning}} &
\multicolumn{5}{c}{\textbf{General \& Logical Reasoning}} &
\multirow{2.5}{*}{\textbf{AVG@8}} \\
\cmidrule(lr){2-7} \cmidrule(lr){8-12}
& Geo3K & MVerse & MVista & WeMath & MV-V & \textit{Avg.}
& MMK12 & Clevr & Logic & MMMU & \textit{Avg.}
& \\
\midrule

\multicolumn{13}{c}{\cellcolor{sechead}\textit{\textbf{Qwen3-VL-4B} Backbone}} \\[1pt]

Base Model (Instruct)
  & 45.42 & 43.39 & 65.80 & 53.39 & 40.90 & \textit{49.78}
  & 39.44 & 90.00 & 42.73 & 20.52 & \textit{48.17}
  & 49.07 \\
\midrule

\multicolumn{13}{c}{\cellcolor{ctxhead}\scriptsize\textit{Other RLVR baselines (context)}} \\[1pt]
SAPO
  & 40.47 & 54.17 & 48.70 & 62.99 & 52.80 & \textit{51.83}
  & 54.02 & 63.50 & 50.43 & 42.89 & \textit{52.71}
  & 52.22 \\
GSPO
  & 59.07 & 70.69 & \textbf{74.90} & 74.89 & 67.13 & \textit{69.34}
  & 56.86 & 91.00 & 53.47 & 41.04 & \textit{60.59}
  & 65.45 \\
CISPO
  & 54.08 & 64.72 & 70.20 & 68.51 & 59.10 & \textit{63.32}
  & 51.38 & 88.50 & 50.11 & 41.33 & \textit{57.83}
  & 60.88 \\
Reinforce++
  & 55.24 & 71.19 & 69.00 & 74.66 & 66.13 & \textit{67.24}
  & 67.37 & 83.00 & 54.81 & 42.54 & \textit{61.93}
  & 64.88 \\
PAPO
  & 52.94 & 69.29 & 68.60 & 73.97 & 64.13 & \textit{65.80}
  & 65.28 & 83.00 & 53.20 & 41.87 & \textit{60.84}
  & 63.59 \\
DVRP
  & 59.43 & 73.18 & 70.80 & 75.42 & 69.13 & \textit{69.60}
  & 61.35 & 90.00 & 54.17 & 43.26 & \textit{61.95}
  & 66.20 \\
VPPO
  & 57.08 & 70.31 & 70.20 & 73.12 & 66.44 & \textit{67.39}
  & 58.22 & 90.50 & 53.17 & 42.06 & \textit{60.99}
  & 64.19 \\
\midrule

\rowcolor{pairbg}
GRPO
  & 53.41 & 67.16 & 68.70 & 68.45 & 63.15 & \textit{64.17}
  & 55.97 & 87.50 & 52.35 & 40.87 & \textit{59.17}
  & 61.95 \\
\rowcolor{ourbg}
\textbf{HiCCR-G} (Ours)
  & 61.73 & \textbf{77.34} & 71.70
  & 78.62 & \textbf{73.85}
  & \textit{\textbf{72.65}\upb{8.48}}
  & \textbf{68.64} & \textbf{94.00}
  & 57.05 & \textbf{46.53}
  & \textit{\textbf{66.55}\upb{7.38}}
  & \textbf{69.94}\upb{7.99} \\
\midrule

\rowcolor{pairbg}
DAPO
  & 60.57 & 74.40 & 72.30 & \textbf{79.48} & 70.26 & \textit{71.40}
  & 60.56 & 92.00 & 54.59 & 44.05 & \textit{62.80}
  & 67.58 \\
\rowcolor{ourbg}
\textbf{HiCCR-D} (Ours)
  & \textbf{63.23} & 73.21 & 73.50
  & 75.69 & 71.18
  & \textit{71.36\dn{0.04}}
  & 65.54 & 91.50
  & \textbf{60.40} & 46.30
  & \textit{65.93\upb{3.13}}
  & 68.95\upb{1.37} \\

\midrule
\multicolumn{13}{c}{\cellcolor{sechead}\textit{\textbf{Qwen3-VL-8B} Backbone}} \\[1pt]

Base Model (Instruct)
  & 51.58 & 53.99 & 65.70 & 60.29 & 50.08 & \textit{56.33}
  & 45.91 & 89.50 & 47.87 & 36.47 & \textit{54.94}
  & 55.71 \\
\midrule

\multicolumn{13}{c}{\cellcolor{ctxhead}\scriptsize\textit{Other RLVR baselines (context)}} \\[1pt]
SAPO
  & 53.35 & 60.21 & 57.20 & 63.47 & 56.62 & \textit{58.17}
  & 50.36 & 72.05 & 52.11 & 41.84 & \textit{54.09}
  & 56.35 \\
GSPO
  & 65.22 & 76.10 & \textbf{76.70} & 78.97 & 72.40 & \textit{73.88}
  & 63.61 & 90.50 & 57.49 & 47.86 & \textit{64.86}
  & 69.87 \\
CISPO
  & 57.57 & 66.15 & 67.70 & 69.89 & 62.23 & \textit{64.71}
  & 53.76 & 91.50 & 52.35 & 45.26 & \textit{60.72}
  & 62.93 \\
Reinforce++
  & 65.89 & 76.59 & 72.80 & 79.20 & 77.22 & \textit{74.94}
  & 70.23 & 90.50 & \textbf{63.98} & \textbf{50.08} & \textit{\textbf{69.95}}
  & 72.32 \\
PAPO
  & 63.23 & 76.03 & 71.40 & 75.84 & 75.06 & \textit{72.31}
  & 70.75 & 89.00 & 60.98 & 50.01 & \textit{67.68}
  & 70.25 \\
DVRP
  & 64.42 & 76.15 & 72.30 & 78.06 & 74.41 & \textit{73.06}
  & 69.53 & 89.00 & 59.45 & 48.54 & \textit{66.63}
  & 70.20 \\
VPPO
  & 60.41 & 69.28 & 72.00 & 73.66 & 67.52 & \textit{68.17}
  & 60.73 & 90.50 & 55.38 & 47.21 & \textit{63.46}
  & 65.58 \\
\midrule

\rowcolor{pairbg}
GRPO
  & 56.74 & 65.23 & 68.40 & 68.91 & 61.93 & \textit{64.24}
  & 54.81 & 90.50 & 52.57 & 43.35 & \textit{60.31}
  & 62.49 \\
\rowcolor{ourbg}
\textbf{HiCCR-G}\textsuperscript{\,1} (Ours)
  & 58.40 & 72.29 & 73.50
  & 76.49 & 68.27
  & \textit{69.79\upb{5.55}}
  & 62.11 & 90.00
  & 57.49 & 46.18
  & \textit{63.94\upb{3.63}}
  & 67.19\upb{4.70} \\
\midrule

\rowcolor{pairbg}
DAPO
  & 64.23 & 73.39 & 71.10 & 77.13 & 68.88 & \textit{70.95}
  & 63.99 & 89.00 & 57.49 & 49.36 & \textit{64.96}
  & 68.29 \\
\rowcolor{ourbg}
\textbf{HiCCR-D} (Ours)
  & \textbf{68.39} & \textbf{78.94} & 75.70
  & \textbf{81.03} & \textbf{75.99}
  & \textit{\textbf{76.01}\upb{5.06}}
  & \textbf{70.58} & \textbf{92.00}
  & 60.63 & 51.56
  & \textit{68.69\upb{3.73}}
  & \textbf{72.76}\upb{4.47} \\

\bottomrule
\end{tabular}%
}

\vspace{3pt}
\begin{minipage}{\textwidth}
\scriptsize\setlength{\parskip}{0pt}
\textcolor{red!80!black}{(+$\Delta$)} / \textcolor{blue!65!black}{($-\Delta$)} are gains against the matched base RL row above. HiCCR wins all 4 AVG@8 aggregates. All baseline experiments are produced in identical environment with same conditions for fair comparison.\\[1pt]
\end{minipage}
\end{table*}
\begin{table*}[t]
\caption{Token-level credit assignment across various reasoning benchmarks under GRPO and DAPO.}
\label{tab:table4_credit_assignment}
\vspace{-1mm}
\centering
\footnotesize
\setlength{\tabcolsep}{2.8pt}
\renewcommand{\arraystretch}{1.02}

\definecolor{ourbg}{HTML}{EDF2FB}     
\definecolor{ourabl}{HTML}{DAE6F2}    
\definecolor{textctxbg}{HTML}{C8D9EC} 
\definecolor{pairbg}{HTML}{F7F7F2}    
\definecolor{sechead}{HTML}{E9ECEF}   

\newcommand{\up}[1]{\,{\scriptsize\textcolor{red!80!black}{(+#1)}}}
\newcommand{\upb}[1]{\,{\footnotesize\textcolor{red!80!black}{\textbf{(+#1)}}}}
\newcommand{\dn}[1]{\,{\scriptsize\textcolor{blue!65!black}{($-$#1)}}}

\resizebox{\textwidth}{!}{%
\begin{tabular}{@{}l ccccc cccc c c@{}}
\toprule
\multirow{2.5}{*}{\textbf{Strategy}} &
\multicolumn{5}{c}{\textbf{Mathematical Reasoning}} &
\multicolumn{4}{c}{\textbf{General \& Logical Reasoning}} &
\multirow{2.5}{*}{\textbf{AVG@8}} &
\multirow{2.5}{*}{$\boldsymbol{\Delta}$} \\
\cmidrule(lr){2-6} \cmidrule(lr){7-10}
& Geo3K & MVerse & MVista & WeMath & MV-V
& MMK12 & Clevr & Logic & MMMU & & \\
\midrule

\multicolumn{12}{c}{\cellcolor{sechead}\textit{\textbf{Qwen3-VL-4B} Backbone}} \\[0.5pt]
Base Model (Instruct)
  & 45.42 & 43.39 & 65.80 & 53.39 & 40.90
  & 39.44 & 90.00 & 42.73 & 20.52
  & 49.07 & \textcolor{gray}{--} \\
\midrule

\multicolumn{12}{l}{\textit{GRPO-based}} \\[0.5pt]
\rowcolor{pairbg}
\quad Uniform Allocation \textcolor{gray}{(Baseline)}
  & 53.41 & 67.16 & 68.70 & 68.45 & 63.15
  & 55.97 & 87.50 & 52.35 & 40.87
  & 61.95 & \textcolor{gray}{--} \\
\quad Random Allocation
  & 56.08 & 70.33 & 66.80 & 68.80 & 64.44
  & 58.19 & 85.00 & 53.39 & 40.60
  & 62.63 & \up{0.68} \\
\quad Time-decay ($\gamma$-decay)
  & 56.40 & 73.19 & 70.80 & 69.43 & 62.86
  & 55.54 & 87.00 & 52.55 & 39.55
  & 63.04 & \up{1.09} \\
\quad Entropy-mask (Top-20\%)
  & 55.91 & 70.62 & 69.30 & 66.64 & 65.22
  & 59.27 & 89.50 & 53.64 & 41.36
  & 63.50 & \up{1.55} \\
\rowcolor{textctxbg}
\quad Text-Context-based (HiCCR-T)
  & 54.07 & 66.51 & 69.25 & 68.12 & 60.78
  & 54.37 & 86.50 & 53.70 & 40.16
  & 61.50 & \dn{0.45} \\
\rowcolor{ourabl}
\quad Raw Attention Energy \textit{(w/o Eq.~\ref{eq:concentration})}
  & 58.42 & 73.67 & 70.20 & 75.19 & 69.58
  & 64.83 & 91.50 & 53.94 & 42.17
  & 66.61 & \up{4.66} \\
\rowcolor{ourbg}
\quad \textbf{Context-based (HiCCR-G, Ours)}
  & \textbf{61.73} & \textbf{77.34} & \textbf{71.70} & \textbf{78.62} & \textbf{73.85}
  & \textbf{68.64} & \textbf{94.00} & \textbf{57.05} & \textbf{46.53}
  & \textbf{69.94} & \up{7.99} \\
\cmidrule(l){1-12}

\multicolumn{12}{l}{\textit{DAPO-based}} \\[0.5pt]
\rowcolor{pairbg}
\quad Uniform Allocation \textcolor{gray}{(Baseline)}
  & 60.57 & \textbf{74.40} & 72.30 & \textbf{79.48} & 70.26
  & 60.56 & \textbf{92.00} & 54.59 & 44.05
  & 67.58 & \textcolor{gray}{--} \\
\quad Random Allocation
  & 63.06 & 74.27 & 72.60 & 75.80 & 70.48
  & \textbf{65.76} & \textbf{92.00} & 58.84 & 44.45
  & 68.58 & \up{1.00} \\
\quad Time-decay ($\gamma$-decay)
  & 57.74 & 70.09 & 71.30 & 74.37 & 68.35
  & 61.01 & 78.50 & 54.36 & 40.23
  & 64.00 & \dn{3.58} \\
\quad Entropy-mask (Top-20\%)
  & 61.06 & 72.84 & 69.00 & 77.39 & 68.88
  & 61.67 & 90.50 & 58.97 & 39.71
  & 66.67 & \dn{0.91} \\
\rowcolor{textctxbg}
\quad Text-Context-based (HiCCR-T)
  & 61.90 & 73.86 & 72.70 & 77.05 & 69.36
  & 59.29 & 91.00 & 56.27 & 43.29
  & 67.19 & \dn{0.39} \\
\rowcolor{ourabl}
\quad Raw Attention Energy \textit{(w/o Eq.~\ref{eq:concentration})}
  & 62.97 & 73.01 & 73.20 & 75.34 & 70.96
  & 65.17 & 91.00 & 60.14 & 46.21
  & 68.67 & \up{1.09} \\
\rowcolor{ourbg}
\quad \textbf{Context-based (HiCCR-D, Ours)}
  & \textbf{63.23} & 73.21 & \textbf{73.50} & 75.69 & \textbf{71.18}
  & 65.54 & 91.50 & \textbf{60.40} & \textbf{46.30}
  & \textbf{68.95} & \upb{1.37} \\

\midrule

\multicolumn{12}{c}{\cellcolor{sechead}\textit{\textbf{Qwen3-VL-8B} Backbone}} \\[0.5pt]
Base Model (Instruct)
  & 51.58 & 53.99 & 65.70 & 60.29 & 50.08
  & 45.91 & 89.50 & 47.87 & 36.47
  & 55.71 & \textcolor{gray}{--} \\
\midrule

\multicolumn{12}{l}{\textit{GRPO-based}} \\[0.5pt]
\rowcolor{pairbg}
\quad Uniform Allocation \textcolor{gray}{(Baseline)}
  & 56.74 & 65.23 & 68.40 & 68.91 & 61.93
  & 54.81 & \textbf{90.50} & 52.57 & 43.35
  & 62.49 & \textcolor{gray}{--} \\
\quad Random Allocation
  & 57.32 & 66.17 & 69.30 & 67.75 & 62.94
  & 55.86 & 90.00 & 52.23 & 42.87
  & 62.72 & \up{0.23} \\
\quad Time-decay ($\gamma$-decay)
  & 57.06 & 65.19 & 71.20 & 69.31 & 67.24
  & 57.09 & 88.50 & 48.39 & 43.17
  & 63.02 & \up{0.53} \\
\quad Entropy-mask (Top-20\%)
  & 55.98 & 67.46 & 69.65 & 71.72 & 63.29
  & 58.69 & 90.00 & 53.61 & 43.25
  & 63.74 & \up{1.25} \\
\rowcolor{textctxbg}
\quad Text-Context-based (HiCCR-T)
  & 53.24 & 70.13 & 70.60 & 73.85 & 66.82
  & 58.96 & 89.00 & 54.59 & 44.39
  & 64.62 & \up{2.13} \\
\rowcolor{ourabl}
\quad Raw Attention Energy \textit{(w/o Eq.~\ref{eq:concentration})}
  & 57.64 & 70.18 & 72.00 & 73.96 & 65.82
  & 60.43 & 89.50 & 55.27 & 44.71
  & 65.50 & \up{3.01} \\
\rowcolor{ourbg}
\quad \textbf{Context-based (HiCCR-G, Ours)}
  & \textbf{58.40} & \textbf{72.29} & \textbf{73.50} & \textbf{76.49} & \textbf{68.27}
  & \textbf{62.11} & 90.00 & \textbf{57.49} & \textbf{46.18}
  & \textbf{67.19} & \upb{4.70} \\
\cmidrule(l){1-12}

\multicolumn{12}{l}{\textit{DAPO-based}} \\[0.5pt]
\rowcolor{pairbg}
\quad Uniform Allocation \textcolor{gray}{(Baseline)}
  & 64.23 & 73.39 & 71.10 & 77.13 & 68.88
  & 63.99 & 89.00 & 57.49 & 49.36
  & 68.29 & \textcolor{gray}{--} \\
\quad Random Allocation
  & 66.39 & 76.53 & 74.60 & 80.55 & 73.84
  & 68.20 & \textbf{93.00} & 60.19 & 49.27
  & 71.40 & \up{3.11} \\
\quad Time-decay ($\gamma$-decay)
  & 62.90 & 71.92 & 72.70 & 76.97 & 73.99
  & 62.34 & 88.50 & 57.72 & 49.54
  & 68.51 & \up{0.22} \\
\quad Entropy-mask (Top-20\%)
  & 65.72 & 77.03 & 74.40 & 79.48 & 73.35
  & 66.92 & 91.00 & 56.60 & 47.98
  & 70.28 & \up{1.99} \\
\rowcolor{textctxbg}
\quad Text-Context-based (HiCCR-T)
  & 62.93 & 74.50 & 73.90 & 77.28 & 70.71
  & 66.21 & 89.50 & 58.96 & 48.61
  & 69.18 & \up{0.89} \\
\rowcolor{ourabl}
\quad Raw Attention Energy \textit{(w/o Eq.~\ref{eq:concentration})}
  & 67.64 & 78.13 & 75.20 & 80.36 & 75.17
  & 69.87 & 91.50 & 59.94 & 50.89
  & 72.08 & \up{3.79} \\
\rowcolor{ourbg}
\quad \textbf{Context-based (HiCCR-D, Ours)}
  & \textbf{68.39} & \textbf{78.94} & \textbf{75.70} & \textbf{81.03} & \textbf{75.99}
  & \textbf{70.58} & 92.00 & \textbf{60.63} & \textbf{51.56}
  & \textbf{72.76} & \upb{4.47} \\

\bottomrule
\end{tabular}
}
\vspace{2pt}
\begin{minipage}{\textwidth}
\scriptsize\setlength{\parskip}{0pt}
Best per framework in \textbf{bold}; $\Delta$ is improvement over Uniform (red = gain, blue = drop).
\colorbox{ourabl}{Raw Attention Energy} is our ablation without Eq.~\ref{eq:concentration}, \colorbox{ourbg}{Context-based} is the full HiCCR score.
\colorbox{textctxbg}{Text-Context-based (HiCCR-T)} replaces the visual reference set $\mathcal{V}$ in Eq.~\ref{eq:concentration} with the non-visual context $\mathcal{X}_{\text{text}}\cup y_{<t}$ (prompt text plus previously generated tokens), isolating the effect of restricting the reference set to $\mathcal{V}$.
\end{minipage}
\end{table*}
\vspace{-1em}
\begin{table*}[t]
\caption{Ablation of the two core components (\textbf{TCR} and \textbf{TAR}) in HiCCR on Qwen3-VL-4B and 8B.\textcolor{red!80!black}{(+$\Delta$)} / \textcolor{blue!65!black}{($-\Delta$)} are per-cell differences against the matched base RL row (grey).}
\label{tab:table3_ablation_study}
\vspace{1mm}
\centering
\small
\setlength{\tabcolsep}{3.6pt}
\renewcommand{\arraystretch}{1.15}

\definecolor{ourbg}{HTML}{EDF2FB}     
\definecolor{ourabl}{HTML}{DAE6F2}    
\definecolor{pairbg}{HTML}{F7F7F2}    
\definecolor{sechead}{HTML}{E9ECEF}   

\newcommand{\up}[1]{\,{\scriptsize\textcolor{red!80!black}{(+#1)}}}
\newcommand{\upb}[1]{\,{\footnotesize\textcolor{red!80!black}{\textbf{(+#1)}}}}
\newcommand{\dn}[1]{\,{\scriptsize\textcolor{blue!65!black}{($-$#1)}}}

\newcommand{\cmark}{\ding{51}}%
\newcommand{\xmark}{\ding{55}}%

\resizebox{\textwidth}{!}{%
\begin{tabular}{@{}l cc ccccc cccc c@{}}
\toprule
\multirow{2.5}{*}{\textbf{Method}} &
\multicolumn{2}{c}{\textbf{Components}} &
\multicolumn{5}{c}{\textbf{Mathematical Reasoning}} &
\multicolumn{4}{c}{\textbf{General \& Logical Reasoning}} &
\multirow{2.5}{*}{\textbf{AVG@8}} \\
\cmidrule(lr){2-3} \cmidrule(lr){4-8} \cmidrule(lr){9-12}
& TCR & TAR
& Geo3K & MVerse & MVista & WeMath & MV-V
& MMK12 & Clevr & Logic & MMMU & \\
\midrule

\multicolumn{13}{c}{\cellcolor{sechead}\textit{\textbf{Qwen3-VL-4B} Backbone}} \\[1pt]
Base Model (Instruct) & &
  & 45.42 & 43.39 & 65.80 & 53.39 & 40.90
  & 39.44 & 90.00 & 42.73 & 20.52
  & 49.07 \\
\midrule

\rowcolor{pairbg}
DAPO
  & \xmark & \xmark
  & 60.57 & 74.40 & 72.30 & 79.48 & 70.26
  & 60.56 & 92.00 & 54.59 & 44.05
  & 67.58 \\
\rowcolor{ourabl}
\quad + TCR
  & \cmark & \xmark
  & 62.45\up{1.88} & 75.12\up{0.72} & 72.80\up{0.50} & 79.56\up{0.08} & 71.34\up{1.08}
  & 63.89\up{3.33} & 91.50\dn{0.50} & 57.23\up{2.64} & 45.67\up{1.62}
  & 68.84\upb{1.26} \\
\rowcolor{ourabl}
\quad + TAR
  & \xmark & \cmark
  & 61.78\up{1.21} & 74.65\up{0.25} & 72.55\up{0.25} & 78.23\dn{1.25} & 70.89\up{0.63}
  & 62.34\up{1.78} & 91.85\dn{0.15} & 56.12\up{1.53} & 44.92\up{0.87}
  & 68.15\upb{0.57} \\
\rowcolor{ourbg}
\textbf{HiCCR-D} (Ours)
  & \cmark & \cmark
  & 63.23\up{2.66} & 73.21\dn{1.19} & \textbf{73.50}\up{1.20} & 75.69\dn{3.79} & 71.18\up{0.92}
  & 65.54\up{4.98} & 91.50\dn{0.50} & \textbf{60.40}\up{5.81} & 46.30\up{2.25}
  & \textbf{68.95}\upb{1.37} \\

\midrule

\multicolumn{13}{c}{\cellcolor{sechead}\textit{\textbf{Qwen3-VL-8B} Backbone}} \\[1pt]
Base Model (Instruct) & &
  & 51.58 & 53.99 & 65.70 & 60.29 & 50.08
  & 45.91 & 89.50 & 47.87 & 36.47
  & 55.71 \\
\midrule

\rowcolor{pairbg}
GRPO
  & \xmark & \xmark
  & 56.74 & 65.23 & 68.40 & 68.91 & 61.93
  & 54.81 & 90.50 & 52.57 & 43.35
  & 62.49 \\
\rowcolor{ourabl}
\quad + TCR
  & \cmark & \xmark
  & 58.76\up{2.02} & 71.34\up{6.11} & 71.50\up{3.10} & 75.23\up{6.32} & 67.45\up{5.52}
  & 61.12\up{6.31} & 90.50\up{0.00} & 56.34\up{3.77} & 46.78\up{3.43}
  & 66.56\upb{4.07} \\
\rowcolor{ourabl}
\quad + TAR
  & \xmark & \cmark
  & 56.23\dn{0.51} & 69.85\up{4.62} & 70.50\up{2.10} & 72.89\up{3.98} & 65.12\up{3.19}
  & 58.76\up{3.95} & 89.75\dn{0.75} & 54.91\up{2.34} & 44.56\up{1.21}
  & 64.73\upb{2.24} \\
\rowcolor{ourbg}
\textbf{HiCCR-G} (Ours)
  & \cmark & \cmark
  & 58.40\up{1.66} & 72.29\up{7.06} & 73.50\up{5.10} & 76.49\up{7.58} & 68.27\up{6.34}
  & 62.11\up{7.30} & 90.00\dn{0.50} & 57.49\up{4.92} & 46.18\up{2.83}
  & \textbf{67.19}\upb{4.70} \\

\bottomrule
\end{tabular}%
}
\end{table*}
\section{Experiments}
\label{sec:experiments}
\subsection{Experimental Setup and Details}
\label{sec:exp_setup}
\paragraph{Data, baselines, benchmarks, and evaluation.}
We train on \textbf{ViRL39K}~\citep{vl-rethinker}, a set of 39K multimodal questions with verifiable answers spanning geometry, algebra, and word problems. We compare against \textbf{nine RLVR algorithms reproduced under identical conditions} (GRPO~\citep{grpo}, DAPO~\citep{dapo}, SAPO~\citep{sapo}, GSPO~\citep{gspo}, CISPO~\citep{cispo}, Reinforce++~\citep{reinforce++}, PAPO~\citep{papo}, DVRP~\citep{dvrp}, VPPO~\citep{vppo_spotlight}), a broad set of proprietary and open-weight multimodal models, and alternative token-level credit assignment strategies. Evaluation spans ten benchmarks across three categories: Mathematical Reasoning, including Geometry3K~\citep{geo3k}, MathVerse~\citep{zhang2024mathverse}, MathVista~\citep{lu2024mathvista}, WeMath~\citep{wemath}, and MathVerse-Vision~\citep{zhang2024mathverse}; General and Logical Reasoning, including MMK12~\citep{mmeurka}, CLEVR-Count~\citep{clevrcount}, LogicVista~\citep{logicvista}, and MMMU~\citep{mmmu}; and Hallucination Robustness, including HallusionBench~\citep{guan2024hallusionbench}. We report accuracy (\%) on each benchmark, averaged over eight independent forward passes (AVG@8). The final training checkpoint is selected for evaluation to ensure fairness. For further details, please refer to Appendix~\ref{sec:details_train_eval}.

\subsection{Main Experiments}
\label{sec:exp_comp}

\paragraph{Comparison with RLVR algorithms and Computation Efficiency.}
HiCCR consistently attains the best or near-best accuracy across both backbones while remaining highly efficient, under identical-condition comparisons with nine RLVR baselines.
Table~\ref{tab:table1_rlvr} shows HiCCR-G reaches \textbf{69.94\%} AVG@8 on 4B and HiCCR-D \textbf{72.76\%} on 8B, surpassing the strongest baselines on both scales and improving their respective base algorithms, with the largest gains on inherently vision-dependent benchmarks such as MathVerse, which is consistent with our hypothesis that uniform credit assignment disproportionately harms visually grounded reasoning.
These gains come at negligible cost: HiCCR adds only marginal wall-clock and peak-memory overhead, far below perturbation-based alternatives such as PAPO~\citep{papo} (Computation Cost Details in Appendix~\ref{app:overhead}).
\vspace{-1em}

\paragraph{Comparison with multimodal models.}
HiCCR-D matches or exceeds several larger open-weight and proprietary baselines on the evaluated benchmark aggregate, demonstrating strong performance beyond the controlled RLVR setting. As reported in Table~\ref{tab:table2_model}(Appendix~\ref{app:more_results}), HiCCR-D achieves \textbf{72.76\%}, surpassing the 72B Qwen2.5-VL-Instruct~\citep{qwen25vl}, underscoring HiCCR's effectiveness in unlocking the multimodal reasoning potential of smaller models.

\vspace{-1em}
\paragraph{Analysis of token-level credit computation strategies.}
The gains of HiCCR come from the concentration structure of visual attention, not from non-uniform weighting in general.
Table~\ref{tab:table4_credit_assignment} compares our score against five alternatives under GRPO and DAPO: Uniform, Random ($w_t\!\sim\!\mathcal{N}(1,\sigma^2)$, $\sigma{=}0.2$), Time-decay ($w_t\!=\!\gamma^{T-t}$), Entropy-mask (top-20\%~\citep{beyond28}), HiCCR-T (same formula over the text context $\mathcal{X}_{\text{text}}\cup y_{<t}$), and Raw Energy (logsumexp of $e_{t,j}^{(h)}$ over $\mathcal{V}$, no intra-visual normalization).
The results form a clear ladder.
Random beats Uniform on every configuration, so uniform credit is not even locally optimal.
Yet non-uniformity alone is insufficient: Time-decay and Entropy-mask can hurt DAPO, and HiCCR-T often falls below Uniform although it uses the same formula as ours.
Concentration helps only when computed over the visual subset.
Within that subset, concentration outperforms magnitude: Raw Energy beats every heuristic but trails our full score across all four configurations.
These comparisons suggest that, among the tested alternatives, the most effective signal is concentration within the visual subset rather than generic non-uniformity, raw attention magnitude, or applying the same formula to textual context.

\paragraph{Generalization to Perception and Hallucination Benchmarks.}
\textbf{HiCCR also improves performance on perception- and hallucination-oriented benchmarks.}
On MMStar~\cite{mmstar} and HalluBench~\cite{guan2024hallusionbench}, HiCCR-G consistently outperforms all baselines across both backbones (Fig.~\ref{fig:mmstar_hallu}, Appendix~\ref{app:more_results}), reaching 71.33\%/72.20\% on MMStar and 59.73\%/62.67\% on HalluBench.
This pattern is consistent with HiCCR's design: by up-weighting visually engaged tokens during training, it encourages more faithful use of image evidence and reduces reliance on spurious textual priors.
\vspace{-1em}

\paragraph{Data size verification and different mllm architecture generalization.}
Restricting training to the small Geo3K split (Appendix~\ref{sec:data_scaling}) and transplanting the full pipeline to a different MLLM family with InternVL3-8B~\citep{internvl25} (Appendix~\ref{sec:arch_generalization}), HiCCR continues to outperform its base RL algorithms in both settings. This indicates that the gains are robust to data scarcity and transfer across distinct vision encoders, connectors, and language priors, supporting HiCCR as a data-efficient and architecture-agnostic credit-reassignment mechanism.

\begin{figure}[t]
    \centering
    \includegraphics[width=\textwidth]{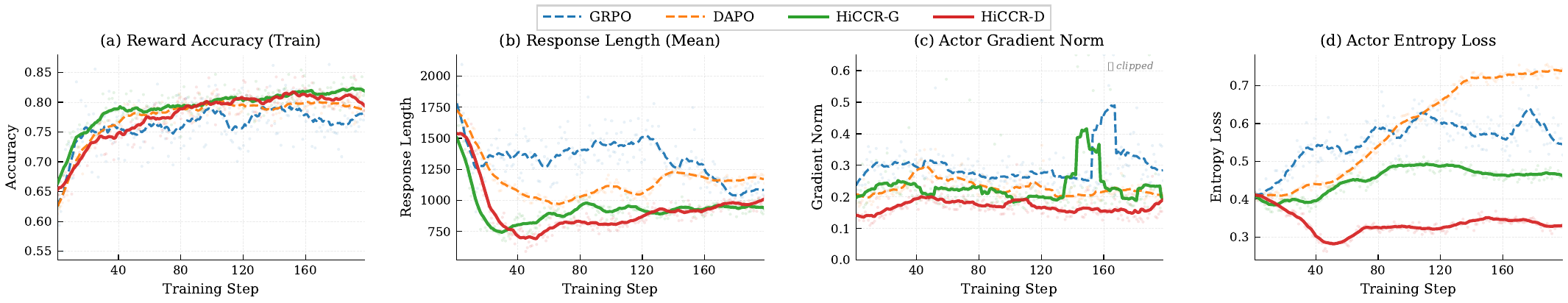}
    \caption{Training dynamics on Qwen3-VL-4B. \textbf{(a)}~Reward accuracy.
    \textbf{(b)}~Mean response length. \textbf{(c)}~Actor gradient norm. \textbf{(d)}~Policy entropy.
    Faint dots show per-step values; solid/dashed lines are smoothed averages.
    HiCCR converges to higher accuracy with shorter responses, stable gradients,
    and lower entropy, indicating efficient and confident visual reasoning.}
    \label{fig:training_dynamics}
\end{figure}

\subsection{Ablation Study and Extented Discussion}
\label{sec:exp_ablation}
\paragraph{Component Contribution and Training Log.}
We incrementally add TCR and TAR to DAPO (4B) and GRPO (8B) to isolate their contributions. Table~\ref{tab:table3_ablation_study} shows the two modules are complementary: on 8B, TCR or TAR alone lifts GRPO's AVG@8 from 62.49\% to 66.56\% and 64.73\%, while combining both in HiCCR-G yields the best overall score, confirming that token- and trajectory-level signals capture distinct aspects of reasoning. Consistently, Figure~\ref{fig:training_dynamics} shows HiCCR reaches higher reward accuracy with shorter responses, smaller gradient norms, and lower entropy than GRPO/DAPO, indicating more stable and confident optimization (Appendix~\ref{sec:training_dynamics}).

\begin{figure*}[t]
    \centering
    \includegraphics[width=\textwidth]{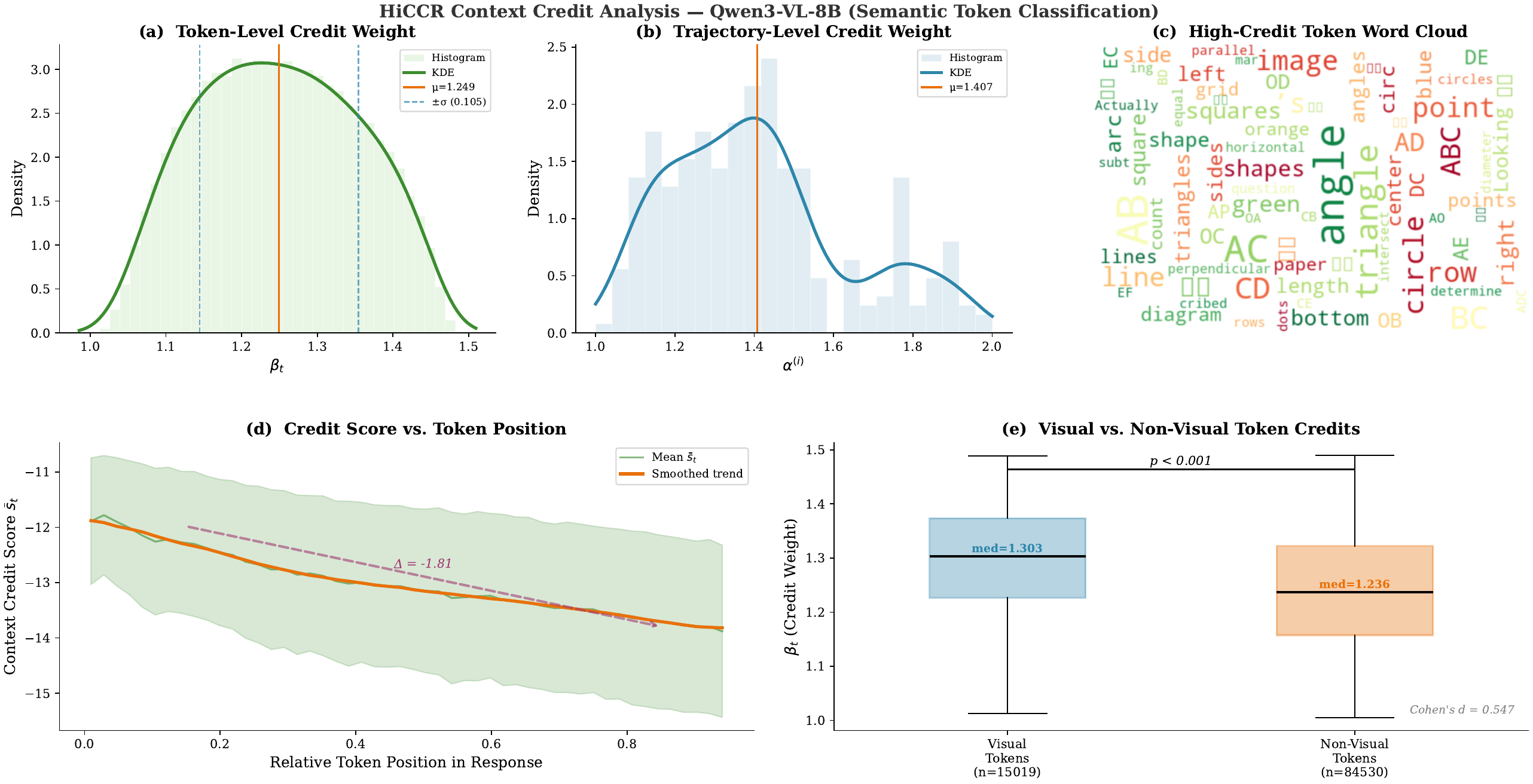}
    \caption{\textbf{HiCCR context credit analysis on Qwen3-VL-8B.}
    (a)~Token-level credit weight $\beta_t$ distribution.
    (b)~Trajectory-level credit weight $\alpha^{(i)}$ distribution.
    (c)~Word cloud of high-credit tokens, sized by average $\beta_t \times$ frequency.
    (d)~Context credit score $\bar{s}_t$ vs.\ relative token position, showing monotonic decay ($\Delta{=}{-}1.81$).
    (e)~Credit weight comparison between semantically visual and non-visual tokens ($p < 0.001$, Cohen's $d{=}0.547$).}
    \label{fig:credit_analysis}
\end{figure*}
\vspace{-1em}

\paragraph{Visualization: How HiCCR redistributes credit.}
A fine-grained credit analysis (Fig.~\ref{fig:credit_analysis}) shows that HiCCR reallocates optimization pressure in a distinctly vision-aware manner rather than uniformly across the response.
Its token-level weights are smooth rather than binary, its trajectory-level weights favor visually engaged rollouts, and its highest-credit tokens are dominated by geometry and diagram semantics.
Moreover, credit is concentrated earlier in generation and decays over token position, suggesting that visual evidence is used most heavily when establishing the reasoning scaffold.
Consistently, semantically visual tokens receive significantly higher credit than non-visual ones, indicating that HiCCR assigns higher weights to tokens that are semantically associated with visual reasoning, without token-level supervision. Full analysis in~Appendix~\ref{sec:context_credits_analysis}.

\paragraph{Ablation: Credit Distribution Strategy.}
Motivated by prior work that updates only the top-$k$\% of token gradients~\citep{vppo_spotlight, beyond28}, we compare \textbf{hard masking} against our \textbf{soft weighting} under identical credit scores (Table~\ref{tab:credit_dapo}, Appendix~\ref{app:more_results}).
Hard masking consistently underperforms: Top-20\% degrades both backbones, and Top-40\% barely matches Uniform ($+0.58$), as discarding sub-threshold tokens forfeits gradient mass from partially informative positions.
The continuous weighting of HiCCR preserves the full token spectrum while amplifying visually grounded steps, and further eliminates the need to tune the sensitive masking ratio $k$, making it both more effective and more practical to deploy.

\paragraph{Ablation: Target Layers for Credit Estimation.}
\textbf{Later transformer layers yield progressively more informative credit signals.}
We ablate which layers supply the visual-attention proxy for HiCCR-G on Qwen3-VL-4B (Table~\ref{tab:layer_ablation_grpo_4b}, Appendix~\ref{app:more_results}). Accuracy improves monotonically from the first to the middle to the last two layers, with the largest margins on visually demanding tasks (MMK12~$+6.86$).
This reflects a representational hierarchy: early layers capture local visual-text alignment but lack task specificity, middle layers begin cross-modal fusion, and the last layers, where the model commits to its reasoning trajectory, best indicate whether attention to the image is causally consequential~\citep{yu2025multimodalllmssolveimage, neo2025interpretingvisualinformationprocessing}.
Credit assignment should therefore be grounded in where the model acts on visual information, not merely where it attends to it.

\paragraph{Ablation: Asymmetric Advantage Shaping}
\textbf{Credit reassignment should be restricted to positive-advantage trajectories.}
HiCCR modulates only correct rollouts ($\hat{A}>0$), leaving negative-advantage trajectories unchanged (Eq.~\ref{eq:hiccr_adv}).
We compare this \textbf{asymmetric} design against a \textbf{symmetric} variant that reshapes both signs under HiCCR-G at 4B and 8B scales (Table~\ref{tab:asymmetric_shaping}, Appendix~\ref{app:more_results}).
Asymmetric shaping yields substantially larger gains, with the widest margins on geometry- and diagram-intensive tasks (Geo3K, WeMath, MV-V).
The mechanism is intuitive: symmetric reshaping amplifies negative gradients for incorrect rollouts that happen to attend strongly to the image, thereby penalising visual attention indiscriminately and conflating visual engagement with correctness.
Restricting reassignment to positive-advantage trajectories avoids this failure mode, ensuring that only correct reasoning paths receive visually guided credit amplification.

\paragraph{Sensitivity to Token-Level Scaling Magnitude $\lambda_{\mathrm{tok}}$}
\textbf{Moderate scaling yields the best trade-off; both under- and over-amplification hurt.}
The hyper-parameter $\lambda_{\mathrm{tok}}$ in Eq.~\ref{eq:beta} controls the maximum amplitude of the per-token credit multiplier $\beta_t = 1 + \lambda_{\mathrm{tok}} \cdot \sigma(\bar{s}_t)$.
We sweep $\lambda_{\mathrm{tok}} \in \{0.1, 0.2, 0.5, 0.8, 1.5\}$ under HiCCR-D on Qwen3-VL-4B (Table~\ref{tab:lambda_tok}, Appendix~\ref{app:more_results}).
Small values ($0.1$--$0.2$) under-differentiate tokens and approach the uniform baseline ($+0.25$/$+0.57$);
$\lambda_{\mathrm{tok}}=0.8$ achieves the best AVG ($69.31$, $+1.73$);
$\lambda_{\mathrm{tok}}=1.5$ over-amplifies the credit gap and destabilises training ($-1.34$), excessively down-weighting tokens that are useful but not directly attending to the image.
Our default $\lambda_{\mathrm{tok}}=0.5$ ($+1.39$) trades a marginal $0.34$ in AVG for greater stability, and performance remains strong across the $[0.5, 0.8]$ range, indicating low sensitivity to exact tuning.

\paragraph{Ablation: Matched Random Controls}
\textbf{Visual credit alignment matters beyond generic non-uniform weighting.}
To disentangle these effects, we introduce two controls.
\emph{Matched Shuffle} permutes HiCCR-derived weights, preserving their empirical distribution and scale while disrupting their correspondence with visual engagement.
\emph{Mean-Preserving HiCCR} retains visual ranking but normalizes the mean positive-advantage multiplier to one.
Table~\ref{tab:matched_controls} shows that HiCCR outperforms Matched Shuffle, while Mean-Preserving HiCCR also improves over Uniform and Random.
These results support the value of visually aligned credit: neither the weight distribution nor average positive-update amplification alone explains the gains.

\begin{table}[t]
\caption{\textbf{Matched random controls.}
AVG@8 (\%) averaged over five training seeds (different from main experiments).
4B-D and 8B-D denote the Qwen3-VL-4B and 8B settings with DAPO, respectively.}
\label{tab:matched_controls}
\centering
\small
\setlength{\tabcolsep}{12pt}
\renewcommand{\arraystretch}{1.08}
\definecolor{ourhighlight}{HTML}{EDF2FB}
\begin{tabular}{@{}lcc@{}}
\toprule
\textbf{Credit Rule} & \textbf{4B-D} & \textbf{8B-D} \\
\midrule
Uniform                  & 67.51 & 68.21 \\
Gaussian Random          & 68.53 & 71.31 \\
Matched Shuffle          & 68.47 & 71.48 \\
Mean-Preserving HiCCR     & 68.84 & 72.24 \\
\rowcolor{ourhighlight}
\textbf{HiCCR (Full)}     & \textbf{69.51} & \textbf{72.65} \\
\bottomrule
\end{tabular}
\end{table}

\paragraph{Grounding--Accuracy Correlation}
\textbf{Grounding recovery is positively associated with accuracy gains.}
We compute Spearman correlations between accuracy improvements and changes in three grounding diagnostics, each measured as HiCCR minus its matched GRPO/DAPO baseline.
Across ten benchmarks and four settings, Table~\ref{tab:grounding_accuracy} shows positive pooled correlations for visual-gradient share ($0.42$), image-attention mass ($0.45$), and zero-out accuracy drop ($0.50$).
The associations vary across settings and are weaker for 8B-G, indicating that grounding recovery does not map one-to-one to accuracy gains.
These results are consistent with grounding contributing to the improvements, but do not establish causal mediation or exclude other mechanisms.

\begin{table}[t]
\caption{\textbf{Grounding--accuracy correlations.}
Spearman $\rho$ between matched accuracy gains and grounding changes.
Each setting contains ten benchmark-level pairs; Pooled combines all 40 pairs.
Image-attention mass is summarized by its area under the generation curve (AUC).
Zero-out drop measures the accuracy decrease after zeroing image embeddings.}
\label{tab:grounding_accuracy}
\centering
\small
\setlength{\tabcolsep}{6pt}
\renewcommand{\arraystretch}{1.08}
\begin{tabular}{@{}lccccc@{}}
\toprule
\textbf{Grounding Diagnostic}
& \textbf{4B-G} & \textbf{4B-D}
& \textbf{8B-G} & \textbf{8B-D} & \textbf{Pooled} \\
\midrule
Visual-gradient share
& 0.64 & 0.36 & 0.15 & 0.31 & \textbf{0.42} \\
Image-attention mass (AUC)
& 0.58 & 0.39 & 0.20 & 0.44 & \textbf{0.45} \\
Zero-out accuracy drop
& 0.71 & 0.47 & 0.18 & 0.51 & \textbf{0.50} \\
\bottomrule
\end{tabular}
\end{table}

\section{Conclusion}

We studied the token-level credit mismatch in multimodal RLVR and showed that uniform credit assignment can bias learning toward textual rather than visual context. To address this issue, we proposed HiCCR, which reassigns credit at both the token and trajectory levels using a lightweight context attention concentration score. HiCCR requires no auxiliary models or additional data, adds less than 3\% wall-clock overhead, and integrates easily into group-relative RLVR algorithms. Experiments on Qwen3-VL-4B and 8B across ten benchmarks show consistent improvements over base RL baselines, suggesting that effective multimodal RL benefits from credit signals that better reflect visual dependence.

\section*{Acknowledgments and Disclosure of Funding}
This work is supported by the National Natural Science Foundation of China (Grant No. 62521004) and the Science and Technology Commission of Shanghai Municipality (No. 24511103100). The authors declare no competing interests.

\small
\bibliographystyle{plain}   
\bibliography{ref}   

\newpage
\appendix

\section*{Appendix Roadmap}
\label{app:roadmap}

This appendix provides supplementary experiments, theory, proxy validation,
mechanistic diagnostics, reproducibility details, qualitative examples,
limitations, and broader impacts.

\vspace{0.6em}

\begingroup
\small
\setlength{\parindent}{0pt}

\definecolor{roadmapblue}{HTML}{1F4E79}
\definecolor{roadmapgray}{HTML}{5F6368}
\definecolor{roadmaplight}{HTML}{EEF3F8}

\newcommand{\appblock}[4]{%
    \vspace{0.55em}
    \noindent
    \colorbox{roadmaplight}{%
        \parbox{\dimexpr\linewidth-2\fboxsep}{%
            \textcolor{roadmapblue}{\textbf{Appendix~\ref*{#1}}}
            \quad
            \hyperref[#1]{\textbf{#2}}
            \hfill
            {\footnotesize\textsc{#3}}
        }%
    }%
    \vspace{0.25em}

    {\color{roadmapgray}#4}
    \vspace{0.35em}
}

\newcommand{\appsubitem}[3]{%
    \item \hyperref[#1]{\textbf{\ref*{#1}~#2}}:
    #3
}


\appblock{app:more_exp_results}
{Additional Experimental Evidence}
{Experiments}
{Additional results on efficiency, scaling, architecture transfer, training
dynamics, benchmarks, and context-credit quality.}

\begin{itemize}[leftmargin=2.2em,itemsep=0.15em,topsep=0.15em]
    \appsubitem{app:overhead}
    {Computational Overhead}
    {wall-clock and memory cost.}

    \appsubitem{sec:data_scaling}
    {Data Scaling Verification}
    {Geo3K-only low-data training.}

    \appsubitem{sec:arch_generalization}
    {Model Architecture Generalization}
    {transfer to InternVL3-8B.}

    \appsubitem{sec:training_dynamics}
    {Training Dynamics}
    {accuracy, length, gradient norm, and entropy.}

    \appsubitem{app:more_results}
    {Remaining Results}
    {additional benchmarks and ablations.}

    \appsubitem{sec:context_credits_analysis}
    {Context Credit Analysis}
    {credit distributions and semantic alignment.}
\end{itemize}

\appblock{sec:pg_theory}
{Theoretical Analysis: From Uniform Credit to Attention Drift}
{Theory}
{Formalizes text-favored gradient bias under uniform credit and explains how
HiCCR mitigates the resulting visual attention drift.}

\appblock{sec:saliency_proxy}
{Context Credit as a Lightweight Visual-Dependence Proxy}
{Proxy Validation}
{Justifies visual-attention concentration as an efficient proxy for visual
dependence in RLVR training.}

\appblock{app:empirical_grounding}
{Mechanistic Interpretation for Visual Grounding}
{Diagnostics}
{Reports gradient attribution, counterfactual visual ablation, and attention
dynamics across training stages.}

\appblock{sec:related_work}
{Related Work}
{Positioning}
{Discusses multimodal reasoning, RLVR, attention-based grounding, and
token-level credit assignment.}

\appblock{sec:details_train_eval}
{Training and Evaluation Details}
{Reproducibility}
{Provides implementation details, hyperparameters, evaluation protocol, and the
HiCCR algorithm pipeline.}

\appblock{sec:case_study}
{Qualitative Case Studies}
{Case Analysis}
{Shows representative examples of visually grounded credit reassignment.}

\appblock{sec:limitation}
{Limitations}
{Scope}
{Discusses assumptions, scope, and possible extensions.}

\appblock{sec:broader_impacts}
{Broader Impacts}
{Impact}
{Discusses potential benefits, risks, and responsible-use considerations.}

\endgroup

\section{More Experiments Results}
\label{app:more_exp_results}

\subsection{Computational Overhead}
\label{app:overhead}

A practical credit-reassignment method must not introduce prohibitive overhead into the already expensive RLVR pipeline.
Table~\ref{tab:compute_overhead} compares the computational cost of HiCCR against uniform baselines (GRPO, DAPO) and the perturbation-based method PAPO under identical hardware and training configurations (8$\times$ High-Performance GPUs, same batch size, rollout length, and total training steps).

HiCCR computes the context credit score via forward-pass attention hooks on the last two decoder layers, requiring no additional backward pass and no attention-weight materialization.
As shown in Table~\ref{tab:compute_overhead}, the per-step wall-clock overhead is only 2.7\% relative to the uniform baselines, well within the $<\!3\%$ budget claimed in \S\ref{sec:intro}.
Peak GPU memory increases by 4.5\%, as the only additional storage is the pre-softmax attention logits at visual key positions for the target layers.
In contrast, PAPO requires counterfactual perturbation passes, incurring 26.3\% wall-clock overhead and 12.0\% additional memory for comparable or lower accuracy.

\begin{table*}[t]
\caption{\textbf{Computational overhead comparison (Qwen3-VL-8B).} All methods are trained on 8$\times$ High-Performance GPUs with identical batch size, rollout configuration, and total training steps. \textit{Rel.\ Overhead} = wall-clock increase vs.\ the matched uniform baseline (GRPO for HiCCR-G, DAPO for HiCCR-D); \textcolor{orange!80!black}{(+$\Delta$)} on overhead columns marks cost increase (lower is better), \textcolor{red!80!black}{(+$\Delta$)} on AVG@8 marks accuracy gain. HiCCR adds only 2.7\% wall-clock overhead while delivering +4.70 / +4.47 AVG@8; PAPO requires $>$20\% overhead for a smaller gain and an extra backward pass.}
\label{tab:compute_overhead}
\vspace{-2mm}
\centering
\small
\setlength{\tabcolsep}{3.6pt}
\renewcommand{\arraystretch}{1.15}

\definecolor{ourbg}{HTML}{EDF2FB}     
\definecolor{ourabl}{HTML}{DAE6F2}    
\definecolor{pairbg}{HTML}{F7F7F2}    
\definecolor{sechead}{HTML}{E9ECEF}   

\newcommand{\up}[1]{\,{\scriptsize\textcolor{red!80!black}{(+#1)}}}
\newcommand{\upb}[1]{\,{\footnotesize\textcolor{red!80!black}{\textbf{(+#1)}}}}
\newcommand{\dn}[1]{\,{\scriptsize\textcolor{blue!65!black}{($-$#1)}}}
\newcommand{\cost}[1]{\,{\scriptsize\textcolor{orange!85!black}{(+#1)}}}

\resizebox{0.92\textwidth}{!}{%
\begin{tabular}{@{}l ccccc c@{}}
\toprule
\multirow{2.5}{*}{\textbf{Method}} &
\makecell{Step Time\\(sec)} & \makecell{Rel.\ Overhead\\(\%)} &
\makecell{Peak Mem\\(GB)} & \makecell{Mem Overhead\\(\%)} &
\makecell{Extra\\Backward} &
\multirow{2.5}{*}{\textbf{AVG@8}} \\
\midrule

\rowcolor{pairbg}
GRPO
  & 619 & \textcolor{gray}{--}
  & 115.8 & \textcolor{gray}{--}
  & \ding{55}
  & 62.49 \\
\rowcolor{ourbg}
\textbf{HiCCR-G} (Ours)
  & 636\cost{2.7} & +2.7
  & 121.0\cost{4.5} & +4.5
  & \ding{55}
  & 67.19\upb{4.70} \\
\cmidrule(l){1-7}

\rowcolor{pairbg}
DAPO
  & 634 & \textcolor{gray}{--}
  & 118.1 & \textcolor{gray}{--}
  & \ding{55}
  & 68.29 \\
\rowcolor{ourbg}
\textbf{HiCCR-D} (Ours)
  & 651\cost{2.7} & +2.7
  & 123.4\cost{4.5} & +4.5
  & \ding{55}
  & \textbf{72.76}\upb{4.47} \\
\cmidrule(l){1-7}

\rowcolor{ourabl}
PAPO \textit{\small(perturbation-based, context)}
  & 782\cost{26.3} & +26.3
  & 129.7\cost{12.0} & +12.0
  & \ding{51}
  & 70.25\up{1.96} \\

\bottomrule
\end{tabular}%
}
\end{table*}

\subsection{Data Scaling Verification}
\label{sec:data_scaling}

To validate data efficiency, we restrict training to the Geo3K training split alone, a minimal corpus covering only geometry problems. As shown in Table~\ref{tab:data_scaling}, HiCCR consistently outperforms its base RL algorithm across both backbones: HiCCR-G exceeds GRPO by +2.69 (4B) and +3.18 (8B), while HiCCR-D surpasses DAPO by +3.18 (4B) and +2.83 (8B). These gains confirm that the effectiveness of HiCCR is algorithm-driven and robust to data scarcity, rather than an artifact of large-scale data memorization.

\begin{table*}[!t]
\caption{Data scaling verification. Models are fine-tuned exclusively on the Geo3K training split to evaluate data efficiency. \textcolor{red!80!black}{(+$\Delta$)} / \textcolor{blue!65!black}{($-\Delta$)} are per-cell differences against the matched base RL row (grey). HiCCR variants consistently surpass base RL algorithms across both model scales under this limited data regime; best AVG@8 per backbone in \textbf{bold}.}
\label{tab:data_scaling}
\vspace{1mm}
\centering
\small
\setlength{\tabcolsep}{3.6pt}
\renewcommand{\arraystretch}{1.15}

\definecolor{ourbg}{HTML}{EDF2FB}     
\definecolor{pairbg}{HTML}{F7F7F2}    
\definecolor{sechead}{HTML}{E9ECEF}   

\newcommand{\up}[1]{\,{\scriptsize\textcolor{red!80!black}{(+#1)}}}
\newcommand{\upb}[1]{\,{\footnotesize\textcolor{red!80!black}{\textbf{(+#1)}}}}
\newcommand{\dn}[1]{\,{\scriptsize\textcolor{blue!65!black}{($-$#1)}}}

\resizebox{\textwidth}{!}{%
\begin{tabular}{@{}l ccccc cccc c@{}}
\toprule
\multirow{2.5}{*}{\textbf{Method}} &
\multicolumn{5}{c}{\textbf{Mathematical Reasoning}} &
\multicolumn{4}{c}{\textbf{General \& Logical Reasoning}} &
\multirow{2.5}{*}{\textbf{AVG@8}} \\
\cmidrule(lr){2-6} \cmidrule(lr){7-10}
& Geo3K & MVerse & MVista & WeMath & MV-V
& MMK12 & Clevr & Logic & MMMU & \\
\midrule

\multicolumn{11}{c}{\cellcolor{sechead}\textit{\textbf{Qwen3-VL-4B} Backbone (Trained on Geo3K Train)}} \\[1pt]
Base Model (Instruct)
  & 45.42 & 43.39 & 65.80 & 53.39 & 40.90
  & 39.44 & 90.00 & 42.73 & 20.52
  & 49.07 \\
\midrule

\rowcolor{pairbg}
GRPO
  & 52.91 & 53.85 & 65.40 & 62.76 & 47.71
  & 45.35 & 87.50 & 44.74 & 25.84
  & 54.01 \\
\rowcolor{ourbg}
\textbf{HiCCR-G} (Ours)
  & 54.81\up{1.90} & 56.93\up{3.08} & 68.50\up{3.10} & 67.18\up{4.42} & 52.06\up{4.35}
  & 49.28\up{3.93} & 88.50\up{1.00} & 45.64\up{0.90} & 27.34\up{1.50}
  & \textbf{56.69}\upb{2.68} \\
\midrule

\rowcolor{pairbg}
DAPO
  & 48.42 & 49.63 & 64.60 & 59.22 & 45.18
  & 39.66 & 88.00 & 42.73 & 25.84
  & 51.48 \\
\rowcolor{ourbg}
\textbf{HiCCR-D} (Ours)
  & 51.75\up{3.33} & 53.35\up{3.72} & 67.20\up{2.60} & 64.25\up{5.03} & 49.62\up{4.44}
  & 45.35\up{5.69} & 89.00\up{1.00} & 43.62\up{0.89} & 27.75\up{1.91}
  & 54.65\upb{3.17} \\

\midrule

\multicolumn{11}{c}{\cellcolor{sechead}\textit{\textbf{Qwen3-VL-8B} Backbone (Trained on Geo3K Train)}} \\[1pt]
Base Model (Instruct)
  & 51.58 & 53.99 & 65.70 & 60.29 & 50.08
  & 45.91 & 89.50 & 47.87 & 36.47
  & 55.71 \\
\midrule

\rowcolor{pairbg}
GRPO
  & 54.58 & 59.86 & 69.20 & 67.53 & 55.05
  & 49.00 & 86.00 & 50.56 & 38.21
  & 58.89 \\
\rowcolor{ourbg}
\textbf{HiCCR-G} (Ours)
  & 59.90\up{5.32} & 65.05\up{5.19} & 69.30\up{0.10} & 70.06\up{2.53} & 58.64\up{3.59}
  & 53.10\up{4.10} & 87.00\up{1.00} & 53.91\up{3.35} & 41.68\up{3.47}
  & \textbf{62.07}\upb{3.18} \\
\midrule

\rowcolor{pairbg}
DAPO
  & 55.41 & 58.72 & 65.80 & 65.06 & 52.98
  & 49.00 & 89.50 & 50.78 & 39.94
  & 58.58 \\
\rowcolor{ourbg}
\textbf{HiCCR-D} (Ours)
  & 60.12\up{4.71} & 62.97\up{4.25} & 67.40\up{1.60} & 68.74\up{3.68} & 55.12\up{2.14}
  & 52.36\up{3.36} & 89.50\up{0.00} & 53.66\up{2.88} & 42.78\up{2.84}
  & 61.41\upb{2.83} \\

\bottomrule
\end{tabular}%
}
\end{table*}

\subsection{Model Architecture Generalization}
\label{sec:arch_generalization}

To verify that HiCCR's benefits are not tied to a single MLLM family, we transplant the full pipeline to InternVL3-8B~\cite{zhu2025internvl3}, whose vision encoder, connector, and language backbone differ substantially from Qwen3-VL, while keeping the ViRL39K training set, rollout configuration ($G=5$, temperature $1.0$), and HiCCR hyperparameters unchanged. As shown in Table~\ref{tab:arch_generalization}, HiCCR-G improves GRPO by $+3.21$ AVG@8 points, and HiCCR-D improves DAPO by $+3.23$ points, with consistent gains across mathematical reasoning, visual counting, and logical reasoning benchmarks. These results indicate that HiCCR provides additional benefits on top of both GRPO and DAPO, and that its concentration-based credit signal transfers across distinct VLM architectures rather than relying on architecture-specific tuning.

\begin{table}[!t]
\caption{Model architecture generalization. We replace the Qwen3-VL backbone with InternVL3-8B and re-run the full RLVR pipeline on ViRL39K under identical hyperparameters. \textcolor{red!80!black}{(+$\Delta$)} / \textcolor{blue!65!black}{($-\Delta$)} are per-cell differences against the matched base RL row (grey). HiCCR variants consistently surpass base RL algorithms across both frameworks, confirming that the gains transfer across VLM families with distinct vision encoders, connectors, and language priors; best AVG@8 in \textbf{bold}.}
\label{tab:arch_generalization}
\vspace{1mm}
\centering
\small
\setlength{\tabcolsep}{3.6pt}
\renewcommand{\arraystretch}{1.15}

\definecolor{ourbg}{HTML}{EDF2FB}
\definecolor{pairbg}{HTML}{F7F7F2}
\definecolor{sechead}{HTML}{E9ECEF}

\providecommand{\up}[1]{\,{\scriptsize\textcolor{red!80!black}{(+#1)}}}
\providecommand{\upb}[1]{\,{\footnotesize\textcolor{red!80!black}{\textbf{(+#1)}}}}
\providecommand{\dn}[1]{\,{\scriptsize\textcolor{blue!65!black}{($-$#1)}}}

\resizebox{\textwidth}{!}{%
\begin{tabular}{@{}l ccccc cccc c@{}}
\toprule
\multirow{2.5}{*}{\textbf{Method}} &
\multicolumn{5}{c}{\textbf{Mathematical Reasoning}} &
\multicolumn{4}{c}{\textbf{General \& Logical Reasoning}} &
\multirow{2.5}{*}{\textbf{AVG@8}} \\
\cmidrule(lr){2-6} \cmidrule(lr){7-10}
& Geo3K & MVerse & MVista & WeMath & MV-V
& MMK12 & Clevr & Logic & MMMU & \\
\midrule

\multicolumn{11}{c}{\cellcolor{sechead}\textit{\textbf{InternVL3-8B} Backbone (Trained on ViRL39K)}} \\[1pt]
Base Model (Instruct)
  & 36.77 & 48.17 & 54.70 & 54.77 & 42.35
  & 48.29 & 71.50 & 40.94 & 29.77
  & 47.47 \\
\midrule

\rowcolor{pairbg}
GRPO
  & 42.28 & 58.62 & 63.70 & 63.68 & 52.28
  & 45.96 & 82.00 & 51.39 & 39.31
  & 55.47 \\
\rowcolor{ourbg}
\textbf{HiCCR-G} (Ours)
  & 45.87\up{3.59} & 63.46\up{4.84} & 66.15\up{2.45} & 66.21\up{2.53} & 54.69\up{2.41}
  & 48.54\up{2.58} & 85.50\up{3.50} & 55.92\up{4.53} & 41.78\up{2.47}
  & 58.68\upb{3.21} \\
\midrule

\rowcolor{pairbg}
DAPO
  & 44.16 & 60.55 & 65.50 & 65.91 & 54.12
  & 48.03 & 84.00 & 53.64 & 41.42
  & 57.48 \\
\rowcolor{ourbg}
\textbf{HiCCR-D} (Ours)
  & 48.74\up{4.58} & 63.21\up{2.66} & 69.05\up{3.55} & 69.47\up{3.56} & 56.55\up{2.43}
  & 51.71\up{3.68} & 86.50\up{2.50} & 56.12\up{2.48} & 45.07\up{3.65}
  & \textbf{60.71}\upb{3.23} \\

\bottomrule
\end{tabular}%
}
\end{table}

\subsection{Training Dynamics}
\label{sec:training_dynamics}

Figure~\ref{fig:training_dynamics} traces four diagnostic metrics across training for all methods on the Qwen3-VL-4B backbone.
\textbf{(a)~Reward accuracy:} HiCCR variants converge faster and stabilize at a higher plateau ($\sim$0.80--0.82) with markedly lower variance, whereas GRPO oscillates persistently in the 0.72--0.80 range.
\textbf{(b)~Response length:} Both baselines exhibit monotonic length inflation---a hallmark of reward hacking---while HiCCR \emph{shortens} responses over training (HiCCR-D reaches $\sim$800 tokens), achieving higher accuracy with more concise reasoning chains.
\textbf{(c)~Gradient norm:} GRPO gradients remain near-zero throughout, signaling weak optimization signal; DAPO norms escalate sharply in later stages, risking instability. HiCCR maintains gradient norms in a healthy, bounded regime (0.15--0.25), consistent with the bounded-modulation design ($\alpha\!\geq\!1,\;\beta\!\geq\!1$) that amplifies without destabilizing.
\textbf{(d)~Entropy:} Baseline entropies rise continuously, reflecting growing policy uncertainty. HiCCR-D entropy \emph{decreases} to $\sim$0.30, indicating that the model learns a confident, visually grounded policy rather than hedging between textual and visual cues.
Taken together, these dynamics show that HiCCR's credit reassignment not only improves final accuracy but also yields a qualitatively healthier optimization trajectory: shorter outputs, stable gradients, and decisive policies.

\subsection{Remaining Results}
\label{app:more_results}
We present remaining experiment results according to the main content order.

\begin{figure}[t]
    \centering
    \includegraphics[width=\linewidth]{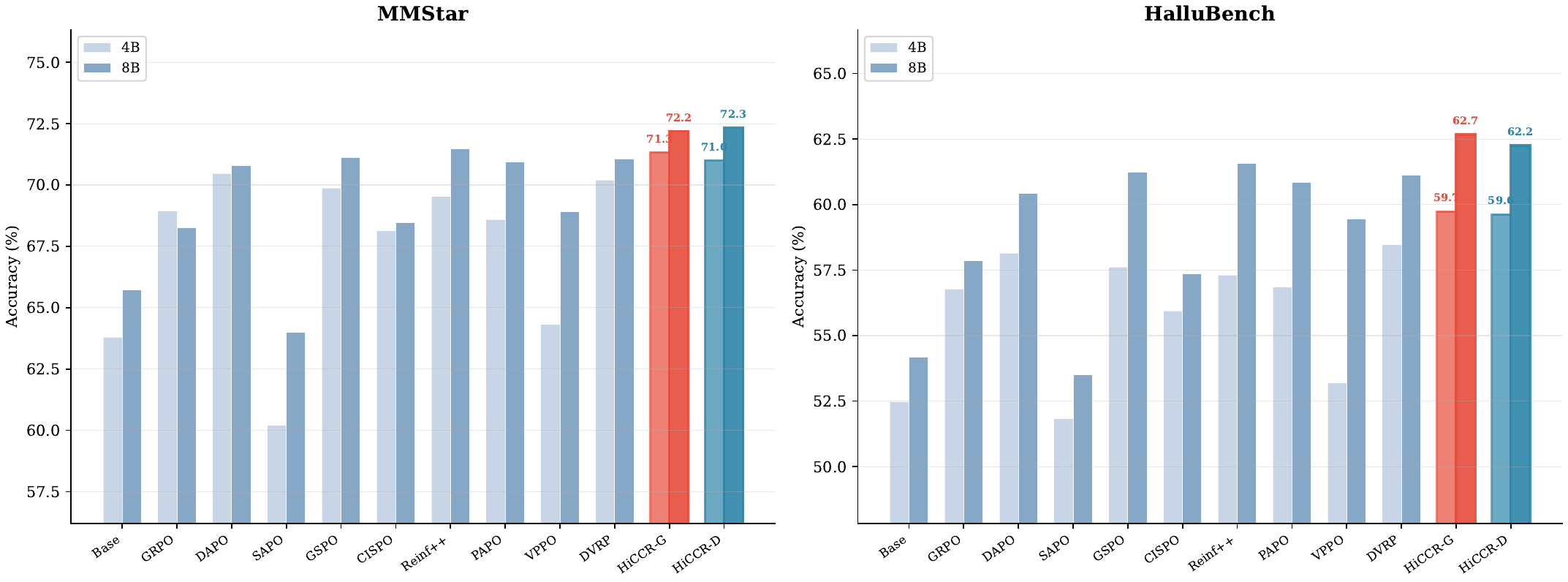}
    \caption{\textbf{Performance on MMStar and HalluBench.} Grouped bars compare all methods on Qwen3-VL-4B and 8B backbones. HiCCR variants (red/blue) consistently achieve the highest accuracy on both benchmarks, indicating improved visual perception and reduced hallucination.}
    \label{fig:mmstar_hallu}
\end{figure}

\begin{table*}[t]
\caption{Comprehensive comparison with multimodal models. Benchmarks are grouped into \textbf{Mathematical Reasoning} (with sub-average) and \textbf{General \& Logical Reasoning} (with sub-average). Missing scores are denoted by ``--''; sub-averages are computed over available entries only. Subscripts $_D$ and $_G$ denote HiCCR built on DAPO and GRPO. \textcolor{red!80!black}{(+$\Delta$)} marks absolute gain over the matched Base Model (grey). Best within each backbone in \textbf{bold}.}
\label{tab:table2_model}
\vspace{-1mm}
\centering
\small
\setlength{\tabcolsep}{3.6pt}
\renewcommand{\arraystretch}{1.15}

\definecolor{ourbg}{HTML}{EDF2FB}     
\definecolor{pairbg}{HTML}{F7F7F2}    
\definecolor{sechead}{HTML}{E9ECEF}   

\newcommand{\up}[1]{\,{\scriptsize\textcolor{red!80!black}{(+#1)}}}
\newcommand{\upb}[1]{\,{\footnotesize\textcolor{red!80!black}{\textbf{(+#1)}}}}
\newcommand{\dn}[1]{\,{\scriptsize\textcolor{blue!65!black}{($-$#1)}}}

\resizebox{\textwidth}{!}{%
\begin{tabular}{@{}l ccccc c cccc c c@{}}
\toprule
\multirow{2.5}{*}{\textbf{Method}} &
\multicolumn{6}{c}{\textbf{Mathematical Reasoning}} &
\multicolumn{5}{c}{\textbf{General \& Logical Reasoning}} &
\multirow{2.5}{*}{\textbf{AVG@8}} \\
\cmidrule(lr){2-7} \cmidrule(lr){8-12}
& Geo3K & MVista & WeMath & MVerse & MV-V & \textit{Avg.}
& MMK12 & Clevr & Logic & MMMU & \textit{Avg.}
& \\
\midrule

\multicolumn{13}{c}{\cellcolor{sechead}\textit{Proprietary Models}} \\[1pt]
GPT-4o
  & 42.31 & 64.70 & 62.84 & 50.15 & 53.77 & \textit{54.75}
  & 55.80 & 95.00 & 64.83 & 59.12 & \textit{68.69}
  & 60.95 \\
GPT-4o-mini
  & 34.58 & 59.90 & 56.27 & 42.36 & 45.14 & \textit{47.65}
  & 51.90 & 88.00 & 54.37 & 46.19 & \textit{60.12}
  & 53.19 \\
Gemini-2.0-flash
  & 48.86 & 70.40 & 47.45 & 47.79 & 48.62 & \textit{52.62}
  & 65.20 & 96.00 & 67.94 & 60.31 & \textit{72.36}
  & 61.40 \\
\midrule

\multicolumn{13}{c}{\cellcolor{sechead}\textit{General-Purpose Multimodal VLMs}} \\[1pt]
Qwen2.5-VL-72B
  & 45.41 & 74.20 & 49.08 & 47.34 & 48.55 & \textit{52.92}
  & 70.50 & 92.50 & 65.82 & 61.47 & \textit{72.57}
  & 61.65 \\
InternVL2.5-8B
  & 33.19 & 64.90 & 44.86 & 37.04 & 40.15 & \textit{44.03}
  & 46.80 & 85.00 & 51.76 & 42.09 & \textit{56.41}
  & 49.53 \\
InternVL2.5-78B
  & 44.57 & 64.90 & 44.88 & 36.95 & 40.24 & \textit{46.31}
  & 59.80 & 90.00 & 61.23 & 54.81 & \textit{66.46}
  & 55.26 \\
LLaVA-OV-7B
  & 32.11 & 58.50 & 44.17 & -- & -- & \textit{44.93}
  & 43.89 & 82.00 & 46.21 & 37.94 & \textit{52.51}
  & 49.26 \\
LLaVA-OV-72B
  & 41.32 & 67.10 & 31.97 & 27.24 & 30.08 & \textit{39.54}
  & 54.27 & 88.50 & 54.83 & 51.62 & \textit{62.31}
  & 49.66 \\
LLaVA-OV-1.5-8B
  & 36.85 & 69.60 & 61.47 & -- & -- & \textit{55.97}
  & 48.55 & 86.00 & 53.79 & 43.15 & \textit{57.87}
  & 57.06 \\
LLaVA-Critic-R1-7B
  & 35.43 & 68.70 & 62.58 & 58.85 & 53.14 & \textit{55.74}
  & 57.40 & 87.50 & 55.83 & 42.67 & \textit{60.85}
  & 58.01 \\
R1-OneVision-7B
  & 30.56 & 64.90 & 55.23 & 61.72 & 44.27 & \textit{51.34}
  & 43.30 & 84.00 & 52.19 & 39.71 & \textit{54.80}
  & 52.88 \\
\midrule

\multicolumn{13}{c}{\cellcolor{sechead}\textit{Domain-Specific Multimodal VLMs}} \\[1pt]
MM-Eureka-7B
  & 36.38 & 59.10 & 45.34 & 57.55 & 56.46 & \textit{50.97}
  & 60.60 & 80.00 & 49.12 & 36.24 & \textit{56.49}
  & 53.42 \\
MM-Eureka-7B-CPGD
  & 37.64 & 64.20 & 64.28 & 63.66 & 59.25 & \textit{57.81}
  & 64.70 & 81.50 & 48.73 & 37.51 & \textit{58.11}
  & 57.94 \\
ADORA-7B
  & 41.17 & 61.10 & 53.06 & 45.18 & 41.83 & \textit{48.47}
  & 49.80 & 78.00 & 44.62 & 33.95 & \textit{51.59}
  & 49.86 \\
R1-VL-7B
  & 31.86 & 63.50 & 56.14 & 42.05 & 43.18 & \textit{47.35}
  & 55.30 & 82.50 & 48.16 & 39.27 & \textit{56.31}
  & 51.33 \\
VLAA-Thinker-7B
  & 24.23 & 67.40 & 65.87 & 47.92 & 51.96 & \textit{51.48}
  & 63.20 & 83.00 & 49.88 & 40.54 & \textit{59.16}
  & 54.89 \\
VL-Rethinker-7B
  & 33.64 & 61.30 & 66.46 & 63.95 & 60.83 & \textit{57.24}
  & 59.80 & 84.50 & 54.12 & 41.73 & \textit{60.04}
  & 58.48 \\
RACRO-7B-CRO-GRPO
  & 41.42 & 61.70 & 68.87 & 65.73 & 61.68 & \textit{59.88}
  & 70.50 & 85.00 & 53.81 & 44.15 & \textit{63.37}
  & 61.43 \\
ThinkLite-7B-VL
  & 34.36 & 68.90 & 63.52 & 49.48 & 46.05 & \textit{52.46}
  & 56.20 & 81.00 & 50.36 & 38.82 & \textit{56.60}
  & 54.30 \\
AVAR-Thinker-7B
  & 32.95 & 58.80 & 49.66 & 53.42 & 52.16 & \textit{49.40}
  & 46.92 & 80.00 & 41.48 & 38.34 & \textit{51.69}
  & 50.41 \\
\midrule

\multicolumn{13}{c}{\cellcolor{sechead}\textit{Ours}} \\[1pt]

\rowcolor{pairbg}
Base Model \textit{\small(Qwen3-VL-4B)}
  & 45.42 & 65.80 & 53.39 & 43.39 & 40.90 & \textit{49.78}
  & 39.44 & 90.00 & 42.73 & 20.52 & \textit{48.17}
  & 49.07 \\
\rowcolor{ourbg}
\quad + \textbf{HiCCR}$_D$
  & \textbf{63.23}\up{17.81} & 73.50\up{7.70} & 75.69\up{22.30}
  & 73.21\up{29.82} & 71.18\up{30.28} & \textit{71.36\upb{21.58}}
  & 65.54\up{26.10} & 91.50\up{1.50} & \textbf{60.40}\up{17.67}
  & 46.30\up{25.78} & \textit{65.93\upb{17.76}}
  & 68.95\upb{19.88} \\
\rowcolor{ourbg}
\quad + \textbf{HiCCR}$_G$
  & 61.73\up{16.31} & 71.70\up{5.90} & 78.62\up{25.23}
  & \textbf{77.34}\up{33.95} & \textbf{73.85}\up{32.95} & \textit{\textbf{72.65}\upb{22.87}}
  & \textbf{68.64}\up{29.20} & \textbf{94.00}\up{4.00} & 57.05\up{14.32}
  & \textbf{46.53}\up{26.01} & \textit{\textbf{66.55}\upb{18.38}}
  & \textbf{69.94}\upb{20.87} \\
\cmidrule(l){1-13}

\rowcolor{pairbg}
Base Model \textit{\small(Qwen3-VL-8B)}
  & 51.58 & 65.70 & 60.29 & 53.99 & 50.08 & \textit{56.33}
  & 45.91 & 89.50 & 47.87 & 36.47 & \textit{54.94}
  & 55.71 \\
\rowcolor{ourbg}
\quad + \textbf{HiCCR}$_D$
  & \textbf{68.39}\up{16.81} & \textbf{75.70}\up{10.00} & \textbf{81.03}\up{20.74}
  & \textbf{78.94}\up{24.95} & \textbf{75.99}\up{25.91} & \textit{\textbf{76.01}\upb{19.68}}
  & \textbf{70.58}\up{24.67} & \textbf{92.00}\up{2.50} & \textbf{60.63}\up{12.76}
  & \textbf{51.56}\up{15.09} & \textit{\textbf{68.69}\upb{13.75}}
  & \textbf{72.76}\upb{17.05} \\
\rowcolor{ourbg}
\quad + \textbf{HiCCR}$_G$
  & 58.40\up{6.82} & 73.50\up{7.80} & 76.49\up{16.20}
  & 72.29\up{18.30} & 68.27\up{18.19} & \textit{69.79\upb{13.46}}
  & 62.11\up{16.20} & 90.00\up{0.50} & 57.49\up{9.62}
  & 46.18\up{9.71} & \textit{63.94\upb{9.00}}
  & 67.19\upb{11.48} \\

\bottomrule
\end{tabular}%
}
\end{table*}

\begin{table*}[t]
\caption{Ablation of credit \emph{distribution} strategies under \textbf{DAPO}.
All non-uniform rows use the same credit scores from our visual-attention proxy; only the score-to-weight mapping differs.
\textit{Hard}: tokens below threshold receive zero weight.
\textit{Soft}: all tokens are rescaled continuously.
$\Delta$: change vs.\ Uniform.  Best per backbone in \textbf{bold}.}
\label{tab:credit_dapo}
\vspace{-2mm}
\centering
\small
\setlength{\tabcolsep}{4.5pt}
\renewcommand{\arraystretch}{1.08}

\definecolor{ourhighlight}{HTML}{EDF2FB}
\definecolor{headgray}{HTML}{F5F5F5}

\resizebox{\textwidth}{!}{%
\begin{tabular}{@{}l l ccccc cccc c c@{}}
\toprule
\multirow{2.5}{*}{\textbf{Category}} &
\multirow{2.5}{*}{\textbf{Strategy}} &
\multicolumn{5}{c}{\textbf{Mathematical Reasoning}} &
\multicolumn{4}{c}{\textbf{General \& Logical Reasoning}} &
\multirow{2.5}{*}{\textbf{AVG}} &
\multirow{2.5}{*}{$\boldsymbol{\Delta}$} \\
\cmidrule(lr){3-7} \cmidrule(lr){8-11}
& & Geo3K & MVerse & MVista & WeMath & MV-V
  & MMK12 & CLEVR & Logic & MMMU & & \\
\midrule

\multicolumn{13}{c}{\cellcolor{headgray}\textit{\textbf{Qwen3-VL-4B}}} \\[2pt]

-- & Uniform \textcolor{gray}{(Baseline)}
  & 60.57 & 74.40 & 72.30 & \textbf{79.48} & 70.26
  & 60.56 & \textbf{92.00} & 54.59 & 44.05
  & 67.58 & \textcolor{gray}{--} \\
\cmidrule(l){1-13}

\multirow{2}{*}{\textit{Hard}}
& Top-20\% Mask
  & 59.07 & 70.92 & 73.20 & 74.20 & 67.43
  & 59.46 & 85.50 & 55.70 & 41.45
  & 65.21 & -2.37 \\
& Top-40\% Mask
  & 63.19 & 73.50 & 73.30 & 76.18 & 70.26
  & 61.28 & 90.00 & 59.73 & 41.39
  & 67.65 & +0.07 \\
\cmidrule(l){1-13}

\rowcolor{ourhighlight}
\textit{Soft} & \textbf{HiCCR (Ours)}
  & \textbf{63.23} & \textbf{73.21} & \textbf{73.50} & 75.69 & \textbf{71.18}
  & \textbf{65.54} & 91.50 & \textbf{60.40} & \textbf{46.30}
  & \textbf{68.95} & {\small \textbf{+1.37}} \\

\midrule

\multicolumn{13}{c}{\cellcolor{headgray}\textit{\textbf{Qwen3-VL-8B}}} \\[2pt]

-- & Uniform \textcolor{gray}{(Baseline)}
  & 64.23 & 73.39 & 71.10 & 77.13 & 68.88
  & 63.99 & 89.00 & 57.49 & 49.36
  & 68.29 & \textcolor{gray}{--} \\
\cmidrule(l){1-13}

\multirow{2}{*}{\textit{Hard}}
& Top-20\% Mask
  & 63.06 & 72.48 & 74.60 & 76.78 & 68.43
  & 61.28 & 89.00 & 58.17 & 47.57
  & 67.93 & -0.36 \\
& Top-40\% Mask
  & 62.40 & 74.40 & 73.70 & 77.18 & 70.03
  & 64.66 & \textbf{92.50} & 58.39 & 46.59
  & 68.87 & +0.58 \\
\cmidrule(l){1-13}

\rowcolor{ourhighlight}
\textit{Soft} & \textbf{HiCCR (Ours)}
  & \textbf{68.39} & \textbf{78.94} & \textbf{75.70} & \textbf{81.03} & \textbf{75.99}
  & \textbf{70.58} & 92.00 & \textbf{60.63} & \textbf{51.56}
  & \textbf{72.76} & {\small \textbf{+4.47}} \\

\bottomrule
\end{tabular}%
}
\end{table*}

\begin{table*}[t]
\caption{Ablation of \textbf{target layers} used for context credit estimation under \textbf{GRPO} on the \textbf{Qwen3-VL-4B} backbone.
We report the original GRPO baseline and HiCCR-G with different target-layer choices.
All HiCCR variants share the same training recipe; the only difference is which two transformer layers are used to extract the visual-attention proxy.}
\label{tab:layer_ablation_grpo_4b}
\vspace{-2mm}
\centering
\small
\setlength{\tabcolsep}{4.5pt}
\renewcommand{\arraystretch}{1.08}

\definecolor{ourhighlight}{HTML}{EDF2FB}

\resizebox{\textwidth}{!}{%
\begin{tabular}{@{}l ccccc cccc c c@{}}
\toprule
\multirow{2.5}{*}{\textbf{Method / Target Layers}} &
\multicolumn{5}{c}{\textbf{Mathematical Reasoning}} &
\multicolumn{4}{c}{\textbf{General \& Logical Reasoning}} &
\multirow{2.5}{*}{\textbf{AVG@8}} &
\multirow{2.5}{*}{$\boldsymbol{\Delta}$} \\
\cmidrule(lr){2-6} \cmidrule(lr){7-10}
& Geo3K & MVerse & MVista & WeMath & MV-V
& MMK12 & Clevr & Logic & MMMU & & \\
\midrule

Original GRPO \textcolor{gray}{(Baseline)}
& 53.41 & 67.16 & 68.70 & 68.45 & 63.15
& 55.97 & 87.50 & 52.35 & 40.87
& 61.95 & \textcolor{gray}{--} \\

\midrule

HiCCR-G w/ First Two Layers
& 61.56 & 72.94 & 70.10 & 76.32 & 70.34
& 61.78 & 91.00 & 55.96 & 44.74
& 67.19 & +5.24 \\

HiCCR-G w/ Middle Two Layers
& 60.43 & 76.56 & 71.20 & 76.66 & 72.34
& 65.92 & 91.50 & 55.92 & 45.86
& 68.49 & +6.54 \\

\rowcolor{ourhighlight}
\textbf{HiCCR-G w/ Last Two Layers (Ours)}
& \textbf{61.73} & \textbf{77.34} & \textbf{71.70} & \textbf{78.62} & \textbf{73.85}
& \textbf{68.64} & \textbf{94.00} & \textbf{57.05} & \textbf{46.53}
& \textbf{69.94} & {\small \textbf{+7.99}} \\

\bottomrule
\end{tabular}%
}
\end{table*}

\begin{table*}[t]
\caption{Ablation of \textbf{asymmetric} \emph{vs.}\ \textbf{symmetric} advantage shaping under \textbf{GRPO} on two model scales.
\textbf{Asymmetric} applies HiCCR credit reassignment only to positive-advantage trajectories;
\textbf{Symmetric} applies it to both positive and negative advantages.}
\label{tab:asymmetric_shaping}
\vspace{-2mm}
\centering
\small
\setlength{\tabcolsep}{4.5pt}
\renewcommand{\arraystretch}{1.08}

\definecolor{ourhighlight}{HTML}{EDF2FB}
\definecolor{headgray}{HTML}{F5F5F5}

\resizebox{\textwidth}{!}{%
\begin{tabular}{@{}l ccccc cccc c c@{}}
\toprule
\multirow{2.5}{*}{\textbf{Method}} &
\multicolumn{5}{c}{\textbf{Mathematical Reasoning}} &
\multicolumn{4}{c}{\textbf{General \& Logical Reasoning}} &
\multirow{2.5}{*}{\textbf{AVG@8}} &
\multirow{2.5}{*}{$\boldsymbol{\Delta}$} \\
\cmidrule(lr){2-6} \cmidrule(lr){7-10}
& Geo3K & MVerse & MVista & WeMath & MV-V
& MMK12 & Clevr & Logic & MMMU & & \\
\midrule

\multicolumn{12}{c}{\cellcolor{headgray}\textit{\textbf{Qwen3-VL-4B} Backbone}} \\[2pt]

Original GRPO \textcolor{gray}{(Baseline)}
& 53.41 & 67.16 & 68.70 & 68.45 & 63.15
& 55.97 & 87.50 & 52.35 & 40.87
& 61.95 & \textcolor{gray}{--} \\
\cmidrule(l){1-12}

HiCCR-G w/ Symmetric
& 55.37 & 71.03 & 69.50 & 70.39 & 66.94
& 62.46 & 89.50 & 55.64 & 42.38
& 64.80 & {\small +2.85} \\

\rowcolor{ourhighlight}
\textbf{HiCCR-G w/ Asymmetric (Ours)}
& \textbf{61.73} & \textbf{77.34} & \textbf{71.70} & \textbf{78.62} & \textbf{73.85}
& \textbf{68.64} & \textbf{94.00} & \textbf{57.05} & \textbf{46.53}
& \textbf{69.94} & {\small \textbf{+7.99}} \\

\midrule

\multicolumn{12}{c}{\cellcolor{headgray}\textit{\textbf{Qwen3-VL-8B} Backbone}} \\[2pt]

Original GRPO \textcolor{gray}{(Baseline)}
& 56.74 & 65.23 & 68.40 & 68.91 & 61.93
& 54.81 & 90.50 & 52.57 & 43.35
& 62.49 & \textcolor{gray}{--} \\
\cmidrule(l){1-12}

HiCCR-G w/ Symmetric
& 55.31 & 67.93 & 69.10 & 66.78 & 64.87
& 57.67 & 89.00 & 53.96 & 44.11
& 63.19 & {\small +0.70} \\

\rowcolor{ourhighlight}
\textbf{HiCCR-G w/ Asymmetric (Ours)}
& \textbf{58.40} & \textbf{72.29} & \textbf{73.50} & \textbf{76.49} & \textbf{68.27}
& \textbf{62.11} & \textbf{90.00} & \textbf{57.49} & \textbf{46.18}
& \textbf{67.19} & {\small \textbf{+4.70}} \\

\bottomrule
\end{tabular}%
}
\end{table*}

\begin{table*}[t]
\caption{Sensitivity of $\lambda_{\mathrm{tok}}$ (token-level scaling magnitude) under \textbf{DAPO} on the \textbf{Qwen3-VL-4B} backbone.
$\lambda_{\mathrm{tok}}$ controls the maximum amplitude of the per-token credit multiplier $\beta_t$.
$\Delta$: improvement over the DAPO baseline.
Best in \textbf{bold}. Default is the setting used in other experiments.}
\label{tab:lambda_tok}
\vspace{-2mm}
\centering
\small
\setlength{\tabcolsep}{4.5pt}
\renewcommand{\arraystretch}{1.08}
\definecolor{ourhighlight}{HTML}{EDF2FB}
\definecolor{headgray}{HTML}{F5F5F5}
\resizebox{\textwidth}{!}{%
\begin{tabular}{@{}l ccccc cccc c c@{}}
\toprule
\multirow{2.5}{*}{\textbf{Method / $\lambda_{\mathrm{tok}}$}} &
\multicolumn{5}{c}{\textbf{Mathematical Reasoning}} &
\multicolumn{4}{c}{\textbf{General \& Logical Reasoning}} &
\multirow{2.5}{*}{\textbf{AVG@8}} &
\multirow{2.5}{*}{$\boldsymbol{\Delta}$} \\
\cmidrule(lr){2-6} \cmidrule(lr){7-10}
& Geo3K & MVerse & MVista & WeMath & MV-V
& MMK12 & Clevr & Logic & MMMU & & \\
\midrule
\multicolumn{12}{c}{\cellcolor{headgray}\textit{\textbf{Qwen3-VL-4B} Backbone}} \\[2pt]
Original DAPO \textcolor{gray}{(Baseline)}
& 60.59 & \textbf{74.42} & 72.30 & \textbf{79.45} & 70.29
& 60.58 & \textbf{92.00} & 54.55 & 44.02
& 67.58 & \textcolor{gray}{--} \\
\cmidrule(l){1-12}
HiCCR-D w/ $\lambda_{\mathrm{tok}}=0.1$
& 61.18 & 74.28 & 72.25 & 78.92 & 70.48
& 61.29 & 91.95 & 55.68 & 44.48
& 67.83 & +0.25 \\
HiCCR-D w/ $\lambda_{\mathrm{tok}}=0.2$
& 61.89 & 73.91 & 72.55 & 78.07 & 70.72
& 62.71 & 91.75 & 56.82 & 44.91
& 68.15 & +0.57 \\
\rowcolor{ourhighlight}
\textbf{HiCCR-D w/ $\lambda_{\mathrm{tok}}=0.5$ (Default)}
& 63.27 & 73.26 & 73.50 & 75.62 & 71.15
& \textbf{65.58} & 91.50 & \textbf{60.47} & \textbf{46.37}
& 68.97 & +1.39 \\
HiCCR-D w/ $\lambda_{\mathrm{tok}}=0.8$
& \textbf{63.48} & 74.37 & \textbf{73.85} & 77.28 & \textbf{71.59}
& 65.32 & 91.80 & 59.81 & 46.28
& \textbf{69.31} & +1.73 \\
HiCCR-D w/ $\lambda_{\mathrm{tok}}=1.5$
& 59.82 & 73.08 & 71.75 & 75.19 & 69.41
& 59.87 & 90.75 & 52.83 & 43.49
& 66.24 & -1.34 \\
\bottomrule
\end{tabular}%
}
\end{table*}

\subsection{Context Credit Quantization and Quality Analysis}
\label{sec:context_credits_analysis}

To understand how HiCCR redistributes optimization pressure across tokens and
trajectories, we conduct a fine-grained credit analysis on 500 samples from the
ViRL training set using a Qwen3-VL-8B checkpoint trained with HiCCR-G.
Figure~\ref{fig:credit_analysis} visualizes the resulting credit distributions
at both the token and trajectory levels.

\paragraph{Token-level credit weight $\beta_t$.}
As shown in Figure~\ref{fig:credit_analysis}(a), the token-level credit weight
$\beta_t = 1 + \lambda_{\mathrm{tok}} \cdot \sigma(\bar{s}_t)$ follows an
approximately Gaussian distribution centered at $\mu{=}1.249$ with
$\sigma{=}0.105$, spanning the effective range $[1.0, 1.5]$. This confirms that
the context credit mechanism produces a smooth, well-spread weighting rather
than a degenerate or binary assignment, enabling fine-grained differentiation
among tokens based on their visual grounding.

\paragraph{Trajectory-level credit weight $\alpha^{(i)}$.}
Figure~\ref{fig:credit_analysis}(b) reveals a right-skewed distribution of
trajectory-level weights, with the majority of trajectories receiving moderate
weights ($\alpha \in [1.1, 1.5]$) and a long tail extending toward
$\alpha{=}2.0$. This indicates that HiCCR selectively amplifies gradients from
trajectories whose reasoning chains exhibit strong visual engagement, while
down-weighting trajectories that rely less on visual context---a desirable
property for multimodal RLVR.

\paragraph{High-credit token semantics.}
The word cloud in Figure~\ref{fig:credit_analysis}(c) displays tokens with the
highest average credit scores. Dominant terms include geometric primitives
(\emph{triangle}, \emph{circle}, \emph{angle}, \emph{square}), spatial
relations (\emph{parallel}, \emph{perpendicular}, \emph{row}, \emph{side}),
diagram references (\emph{image}, \emph{diagram}, \emph{point}), and geometric
vertex labels (\emph{AB}, \emph{BC}, \emph{AC}, \emph{AD}). This qualitatively
demonstrates that HiCCR assigns higher credit to tokens that are semantically
grounded in visual content.

\paragraph{Credit decay along generation.}
Figure~\ref{fig:credit_analysis}(d) plots the mean context credit score
$\bar{s}_t$ against relative token position in the response. A monotonic decay
of $\Delta{=}{-}1.81$ is observed from the beginning to the end of generation.
This pattern is consistent with the intuition that early response tokens---which
typically establish the visual interpretation and set up the reasoning
scaffold---attend more strongly to the image, whereas later tokens progressively
shift toward symbolic computation and answer formatting. And this finding aligns with our attention pattern experiments ~\ref{app:attention_dynamics}.

\paragraph{Visual vs.\ non-visual token credits.}
Finally, we classify each response token as \emph{visual} or \emph{non-visual}
based on an independent semantic lexicon of ${\sim}$150 geometry and spatial
terms. As shown in
Figure~\ref{fig:credit_analysis}(e), visual tokens receive a significantly
higher median credit weight ($\tilde{\beta}{=}1.303$) than non-visual tokens
($\tilde{\beta}{=}1.236$), with $p < 0.001$ under a Mann--Whitney $U$ test and
a medium effect size of Cohen's $d{=}0.547$. This provides direct quantitative
evidence that HiCCR's credit mechanism successfully up-weights visually grounded
tokens without any explicit supervision on token-level labels.

\section{Theoretical Analysis: From Uniform Credit to Attention Drift}
\label{sec:pg_theory}

We analyze how uniform credit assignment can bias multimodal RLVR updates toward non-visual context.
The goal of this section is not to prove a global training guarantee, but to isolate a local first-order mechanism: under mild structural conditions, the same trajectory-level advantage can allocate a larger gradient budget to non-visual tokens and induce a local drift of attention away from visual tokens.
We then show how HiCCR changes the corresponding diagonal update through token- and trajectory-level reweighting.

Throughout this section, we analyze a single attention head in a single layer $\ell$ and suppress layer/head superscripts unless needed.
The extension to the full transformer is discussed at the end as a local decomposition argument rather than a separate global theorem.

\subsection{Setup and Notation}
\label{sec:theory:setup}

Let $x=(x_{\mathrm{vis}},x_{\mathrm{text}})$ denote the multimodal prompt.
At generation step $t$, let $\mathcal V\subset\{1,\ldots,|x|\}$ index visual tokens, and let
$\mathcal T_t\subset\{1,\ldots,|x|+t-1\}$ index all non-visual positions, including textual prompt tokens and the generated prefix $y_{<t}$.
Thus, throughout this section, ``textual'' should be read as ``non-visual'' unless otherwise specified.
We write $\mathcal T$ for $\mathcal T_t$ when $t$ is clear.

Let
\[
   z_j^{(\ell)}=\mathrm{LN}\!\left(h_j^{(\ell)}\right)
\]
be the normalized hidden state used by a pre-norm attention block.
The attention branch computes
\begin{equation}
   a_t^{(\ell)}
   =
   \sum_{j=1}^{|x|+t-1}
   \alpha_{tj} W_V z_j^{(\ell)},
   \qquad
   \alpha_{tj}=\mathrm{softmax}_j(e_{tj}),
   \qquad
   e_{tj}=\frac{q_t^\top k_j}{\sqrt{d_h}},
   \label{eq:attn_branch}
\end{equation}
where
\[
   q_t=W_Q z_t^{(\ell)},\qquad
   k_j=W_K z_j^{(\ell)}.
\]
The residual branch is omitted from the derivation because it does not depend on $W_Q$ and therefore does not affect the $W_Q$-gradient considered below.
For notational economy, we write $h_j^{(\ell)}$ for the normalized state $z_j^{(\ell)}$ in the rest of the analysis.

The visual and non-visual attention masses are
\begin{equation}
   \alpha_t^{\mathcal V}
   =
   \sum_{j\in\mathcal V}\alpha_{tj},
   \qquad
   \alpha_t^{\mathcal T}
   =
   \sum_{j\in\mathcal T}\alpha_{tj}.
   \label{eq:attn_mass}
\end{equation}
Since $\mathcal V\cup\mathcal T$ covers the attended positions, $\alpha_t^{\mathcal V}+\alpha_t^{\mathcal T}=1$.

We focus on the local unclipped regime of group-relative RLVR, where the importance ratio satisfies $r_t(\theta)\approx 1$ and the clipped objective reduces locally to the standard policy-gradient term.
For one trajectory with scalar advantage $\hat A$, the local objective is
\begin{equation}
   \mathcal L(\theta)
   =
   -\hat A
   \sum_{t=1}^{T}
   \log\pi_\theta(y_t\mid x,y_{<t}).
   \label{eq:uniform_pg}
\end{equation}
This isolates the credit-assignment structure; clipping gates the magnitude of updates but does not change the fact that standard RLVR assigns the same $\hat A$ to every token in the trajectory.
Unless otherwise stated, the drift results below are conditional on $\hat A>0$.
For group-relative objectives, the full batch-level update additionally depends on the correlation between $\hat A^{(i)}$ and the local drift coefficients across sampled trajectories.

\subsection{Gradient Decomposition for Attention Parameters}
\label{sec:theory:grad_decomp}

Let
\[
   d_t
   \triangleq
   \frac{\partial \log \pi_\theta(y_t\mid x,y_{<t})}
        {\partial h_t^{(\ell+1)}}
   \in\mathbb R^d,
   \qquad
   v_j=W_V h_j^{(\ell)}.
\]
By the softmax Jacobian,
\begin{equation}
   \frac{\partial\alpha_{tj}}{\partial e_{tk}}
   =
   \alpha_{tj}(\mathbf 1\{j=k\}-\alpha_{tk}),
   \label{eq:softmax_jac}
\end{equation}
the gradient of the token log-probability with respect to $W_Q$ can be written as
\begin{equation}
   \frac{\partial \log\pi_\theta(y_t\mid x,y_{<t})}{\partial W_Q}
   =
   \sum_j
   \alpha_{tj}
   \left(d_t^\top v_j-\bar s_t\right)
   \frac{\partial e_{tj}}{\partial W_Q},
   \label{eq:full_grad_WQ}
\end{equation}
where
\begin{equation}
   \bar s_t
   =
   \sum_k \alpha_{tk} d_t^\top v_k
   =
   \alpha_t^{\mathcal V}\mu_{\mathcal V,t}
   +
   \alpha_t^{\mathcal T}\mu_{\mathcal T,t},
   \label{eq:sbar}
\end{equation}
with
\begin{equation}
   \mu_{\mathcal S,t}
   =
   \frac{1}{\alpha_t^{\mathcal S}}
   \sum_{j\in\mathcal S}
   \alpha_{tj} d_t^\top v_j,
   \qquad
   \mathcal S\in\{\mathcal V,\mathcal T\}.
   \label{eq:modality_mean}
\end{equation}

\begin{definition}[Per-modality gradient contribution]
\label{def:grad_contrib}
For $\mathcal S\in\{\mathcal V,\mathcal T\}$, define
\begin{equation}
   G_t^{\mathcal S}
   =
   \sum_{j\in\mathcal S}
   \alpha_{tj}
   \left(d_t^\top v_j-\bar s_t\right)
   \frac{\partial e_{tj}}{\partial W_Q}.
   \label{eq:grad_contrib}
\end{equation}
Then
\[
   \frac{\partial \log\pi_\theta(y_t\mid x,y_{<t})}{\partial W_Q}
   =
   G_t^{\mathcal V}+G_t^{\mathcal T}.
\]
\end{definition}

\subsection{Structural Conditions}
\label{sec:theory:dominance}

We use two types of conditions.
The first captures the language-prior asymmetry of pretrained VLMs.
The remaining conditions are local regularity assumptions needed to convert activation asymmetry into gradient and attention-update statements.

\begin{assumption}[Language-prior dominance]
\label{asm:lpd}
\leavevmode
\begin{enumerate}[label=(\alph*),leftmargin=*,nosep]
\item \textbf{Absolute-scale dominance.}
For each generation step $t$, there exists $\kappa_t>1$ such that
\begin{equation}
   \mathbb E_{j\sim\mathrm{Unif}(\mathcal T)}
   \left[|d_t^\top v_j|\right]
   \ge
   \kappa_t
   \mathbb E_{j\sim\mathrm{Unif}(\mathcal V)}
   \left[|d_t^\top v_j|\right].
   \label{eq:lpd_abs}
\end{equation}

\item \textbf{Signed alignment gap.}
For a subset $\mathcal H$ of generation steps, there exists $\delta_t>0$ such that
\begin{equation}
   \mu_{\mathcal T,t}-\mu_{\mathcal V,t}
   \ge
   \delta_t,
   \qquad t\in\mathcal H.
   \label{eq:lpd_signed}
\end{equation}
We call such steps \emph{text-dominant} steps.
\end{enumerate}
\end{assumption}

The two parts are logically distinct.
Assumption~\ref{asm:lpd}(a) states that non-visual value activations have larger absolute scale on average.
Assumption~\ref{asm:lpd}(b) states that, on text-dominant steps, the attention-weighted signed activation is larger on the non-visual side.
The empirical analysis below supports the absolute-scale component; the signed component is used as a local condition for the drift result.

\begin{assumption}[Low cancellation]
\label{asm:low_cancel}
For $\mathcal S\in\{\mathcal V,\mathcal T\}$, there exists $c_{\mathcal S}\in(0,1]$ such that
\begin{equation}
   \left\|
       \sum_{j\in\mathcal S}
       \alpha_{tj}
       \left(d_t^\top v_j-\bar s_t\right)
       \frac{\partial e_{tj}}{\partial W_Q}
   \right\|_F
   \ge
   c_{\mathcal S}
   \sum_{j\in\mathcal S}
   \alpha_{tj}
   \left|
       d_t^\top v_j-\bar s_t
   \right|
   \left\|
       \frac{\partial e_{tj}}{\partial W_Q}
   \right\|_F.
   \label{eq:low_cancel}
\end{equation}
This excludes adversarial cancellation among same-modality gradient directions.
\end{assumption}

\begin{assumption}[Jacobian-scale regularity]
\label{asm:jacobian_reg}
There exist constants $0<c_J\le C_J<\infty$ and a step-dependent scale $J_t>0$ such that for all attended positions $j$,
\begin{equation}
   c_J J_t
   \le
   \left\|
       \frac{\partial e_{tj}}{\partial W_Q}
   \right\|_F
   \le
   C_J J_t.
   \label{eq:jacobian_reg}
\end{equation}
Since
\[
   \frac{\partial e_{tj}}{\partial W_Q}
   =
   \frac{k_j h_t^{(\ell)\top}}{\sqrt{d_h}},
\]
this condition corresponds to comparable key-vector norms at the analyzed layer.
\end{assumption}

\begin{assumption}[Local key-metric alignment]
\label{asm:key_metric}
Let
\[
   G_t=G_t^{\mathcal V}+G_t^{\mathcal T}.
\]
For text-dominant steps $t\in\mathcal H$, the key-induced metric does not reverse the logit-space direction of the signed gap:
\begin{equation}
   -\left\langle
       \frac{\partial \alpha_t^{\mathcal V}}{\partial W_Q},
       G_t
   \right\rangle_F
   \ge
   \zeta\,
   C_t\,
   \alpha_t^{\mathcal V}\alpha_t^{\mathcal T}
   \left(\mu_{\mathcal T,t}-\mu_{\mathcal V,t}\right),
   \qquad
   \zeta>0,
   \label{eq:key_metric}
\end{equation}
where
\begin{equation}
   C_t
   =
   \frac{\|h_t^{(\ell)}\|^2}{d_h}
   \sum_j \alpha_{tj}\|k_j\|^2
   >0.
   \label{eq:Ct_def}
\end{equation}
This assumption is used only to sign the same-step diagonal attention update.
It is weaker and more explicit than requiring a global attention-dynamics guarantee.
\end{assumption}

\paragraph{Empirical support for absolute-scale LPD.}
We measure
\[
   \kappa_{\mathrm{emp}}
   =
   \frac{
      \mathbb E_{j\in\mathcal T}
      \left[|d_t^\top v_j|\right]
   }{
      \mathbb E_{j\in\mathcal V}
      \left[|d_t^\top v_j|\right]
   }
\]
on Qwen3-VL and InternVL3 using full forward--backward passes.
This experiment supports Assumption~\ref{asm:lpd}(a), but does not claim to verify all regularity assumptions.

\begin{figure}[t]
   \centering
   \begin{subfigure}[t]{0.48\textwidth}
       \centering
       \includegraphics[width=\textwidth]{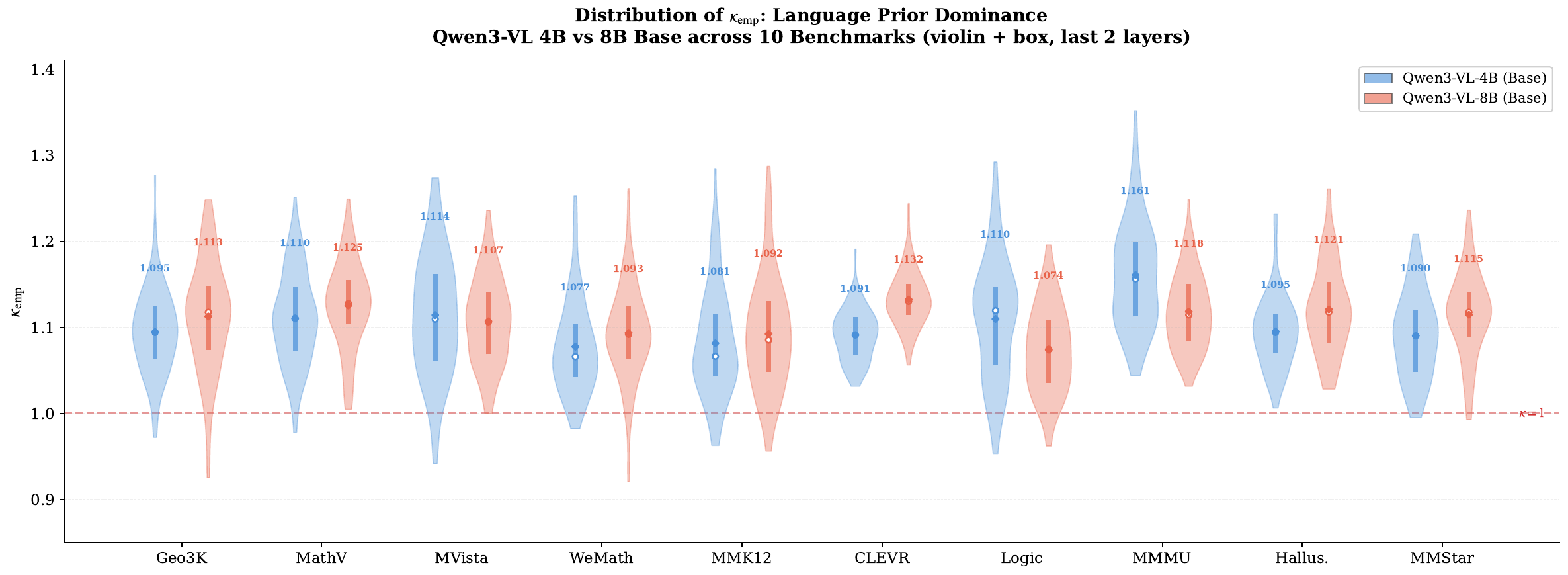}
       \caption{Qwen3-VL per-benchmark $\kappa_{\mathrm{emp}}$ (last two layers).}
       \label{fig:kappa_violin_qwen}
   \end{subfigure}
   \hfill
   \begin{subfigure}[t]{0.48\textwidth}
       \centering
       \includegraphics[width=\textwidth]{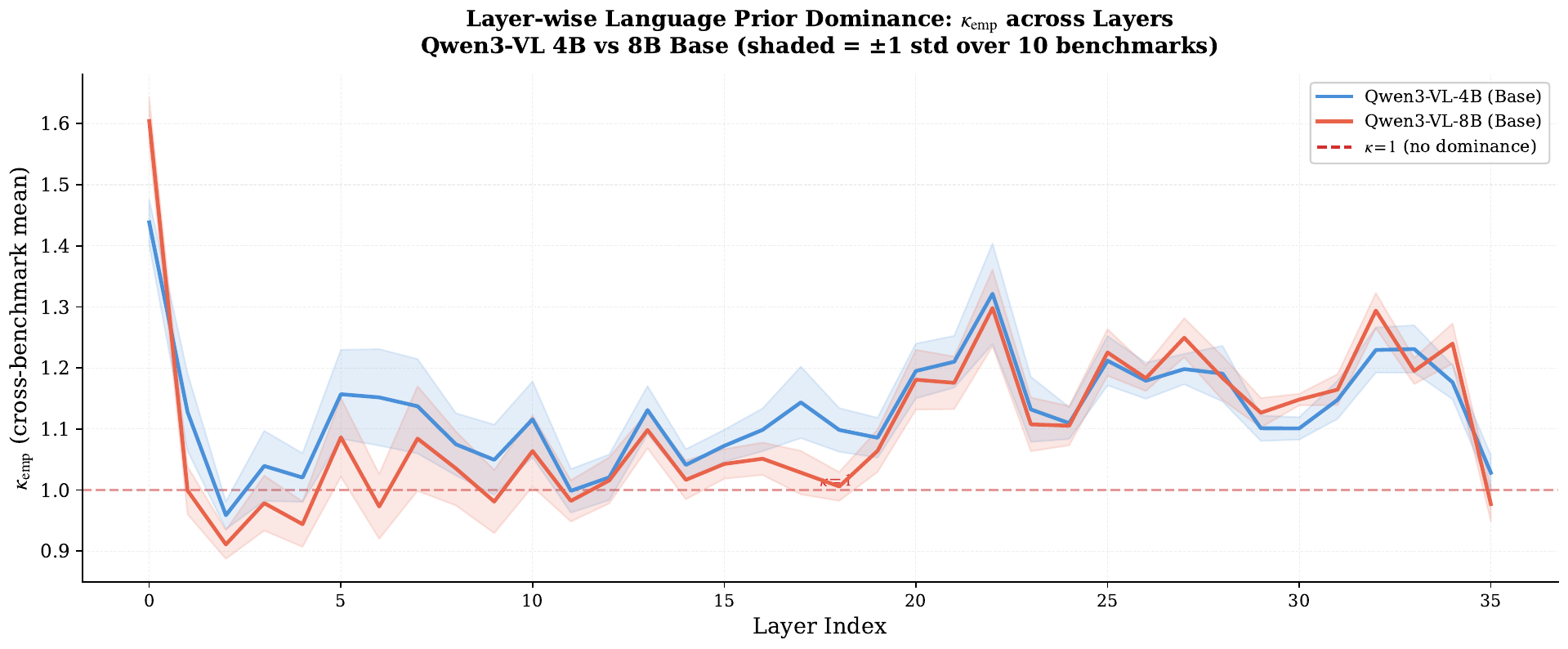}
       \caption{Qwen3-VL layer-wise $\kappa_{\mathrm{emp}}$.}
       \label{fig:kappa_layerwise_qwen}
   \end{subfigure}

   \vspace{0.3em}

   \begin{subfigure}[t]{0.48\textwidth}
       \centering
       \includegraphics[width=\textwidth]{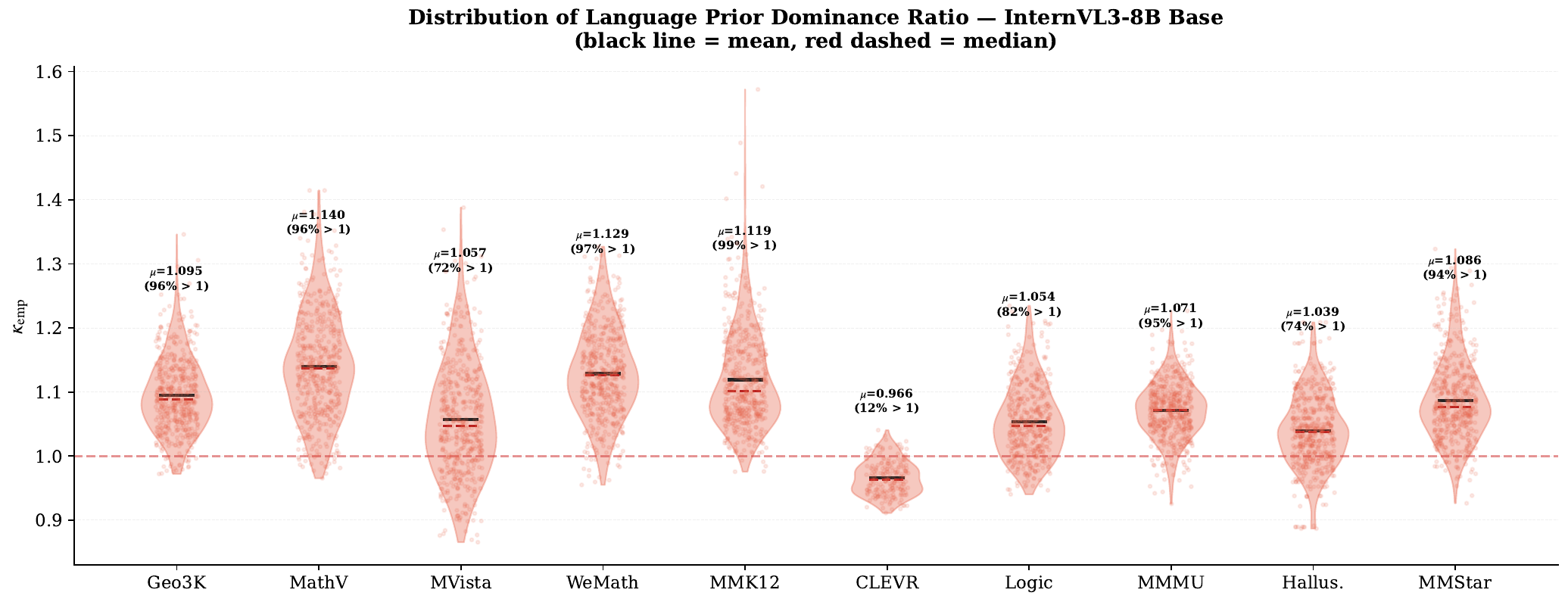}
       \caption{InternVL3-8B per-benchmark $\kappa_{\mathrm{emp}}$ (last two layers).}
       \label{fig:kappa_violin_internvl}
   \end{subfigure}
   \hfill
   \begin{subfigure}[t]{0.48\textwidth}
       \centering
       \includegraphics[width=\textwidth]{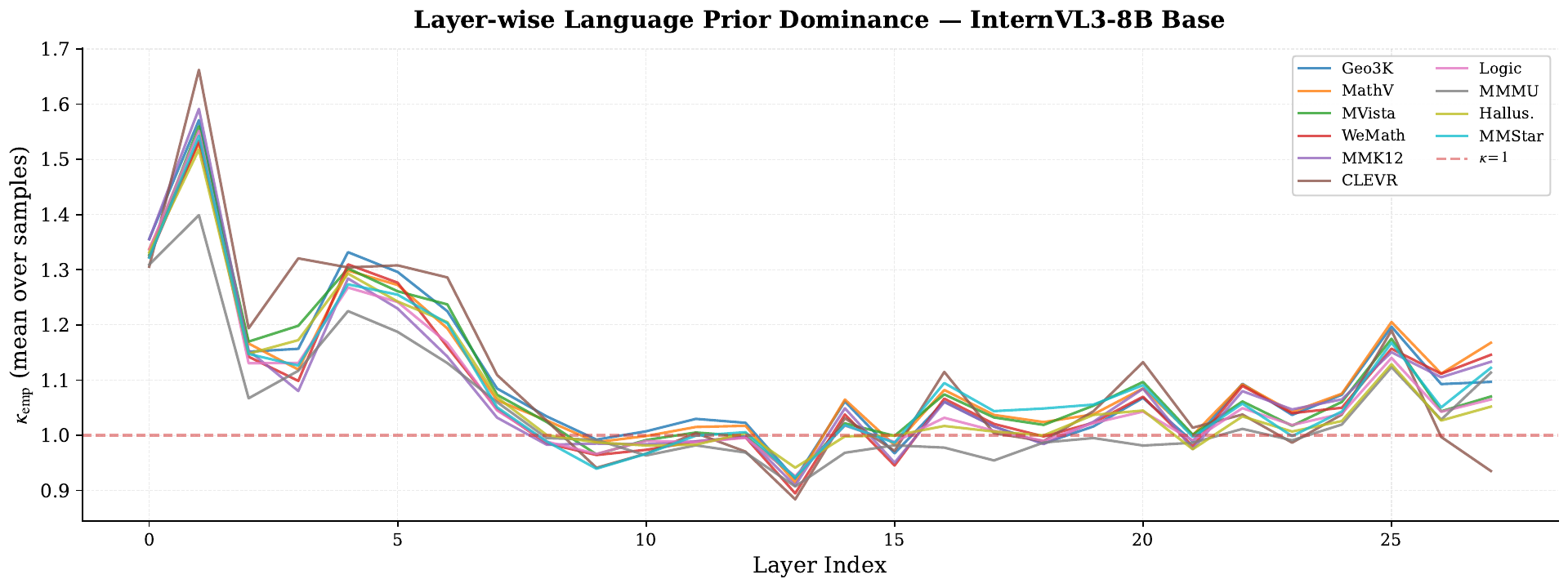}
       \caption{InternVL3-8B layer-wise $\kappa_{\mathrm{emp}}$.}
       \label{fig:kappa_layerwise_internvl}
   \end{subfigure}
   \caption{
   Empirical support for the absolute-scale component of language-prior dominance.
   Across the evaluated benchmarks and model families, $\kappa_{\mathrm{emp}}$ is consistently above one in the measured layers, suggesting that non-visual context often carries larger downstream-projected activation magnitude than visual context.
   }
   \label{fig:kappa_emp}
\end{figure}

\subsection{Main Results}
\label{sec:theory:main}

We first relate the signed gap to the centered quantities that appear in the attention-gradient decomposition.

\begin{lemma}[Centered activation transfer]
\label{lem:centred_lpd}
Fix a text-dominant step $t\in\mathcal H$.
Define
\[
   \sigma_{\mathcal V,t}
   =
   \frac{1}{\alpha_t^{\mathcal V}}
   \sum_{j\in\mathcal V}
   \alpha_{tj}
   \left|
       d_t^\top v_j-\mu_{\mathcal V,t}
   \right|.
\]
Then
\begin{equation}
   \rho_t
   \triangleq
   \frac{
       \sum_{j\in\mathcal T}
       \alpha_{tj}
       |d_t^\top v_j-\bar s_t|
   }{
       \sum_{j\in\mathcal V}
       \alpha_{tj}
       |d_t^\top v_j-\bar s_t|
   }
   \ge
   \gamma_t,
   \label{eq:rho_gamma}
\end{equation}
where
\begin{equation}
   \gamma_t
   =
   \frac{
      \alpha_t^{\mathcal T}
      (\mu_{\mathcal T,t}-\mu_{\mathcal V,t})
   }{
      \sigma_{\mathcal V,t}
      +
      \alpha_t^{\mathcal T}
      (\mu_{\mathcal T,t}-\mu_{\mathcal V,t})
   }.
   \label{eq:gamma_def}
\end{equation}
\end{lemma}

\begin{proof}
Since $t\in\mathcal H$,
\[
   \mu_{\mathcal T,t}\ge \mu_{\mathcal V,t}+\delta_t.
\]
Using
\[
   \bar s_t
   =
   \alpha_t^{\mathcal V}\mu_{\mathcal V,t}
   +
   \alpha_t^{\mathcal T}\mu_{\mathcal T,t},
\]
we have
\[
   \mu_{\mathcal T,t}-\bar s_t
   =
   \alpha_t^{\mathcal V}
   (\mu_{\mathcal T,t}-\mu_{\mathcal V,t}),
   \qquad
   \bar s_t-\mu_{\mathcal V,t}
   =
   \alpha_t^{\mathcal T}
   (\mu_{\mathcal T,t}-\mu_{\mathcal V,t}).
\]
By Jensen's inequality,
\begin{align}
   \sum_{j\in\mathcal T}
   \alpha_{tj}
   |d_t^\top v_j-\bar s_t|
   &\ge
   \left|
      \sum_{j\in\mathcal T}
      \alpha_{tj}
      (d_t^\top v_j-\bar s_t)
   \right|  \nonumber\\
   &=
   \alpha_t^{\mathcal T}\alpha_t^{\mathcal V}
   (\mu_{\mathcal T,t}-\mu_{\mathcal V,t}).
   \label{eq:text_center_lower}
\end{align}
For the visual side, the triangle inequality gives
\begin{align}
   \sum_{j\in\mathcal V}
   \alpha_{tj}
   |d_t^\top v_j-\bar s_t|
   &\le
   \sum_{j\in\mathcal V}
   \alpha_{tj}
   |d_t^\top v_j-\mu_{\mathcal V,t}|
   +
   \sum_{j\in\mathcal V}
   \alpha_{tj}
   |\mu_{\mathcal V,t}-\bar s_t| \nonumber\\
   &=
   \alpha_t^{\mathcal V}
   \sigma_{\mathcal V,t}
   +
   \alpha_t^{\mathcal V}\alpha_t^{\mathcal T}
   (\mu_{\mathcal T,t}-\mu_{\mathcal V,t}).
   \label{eq:vis_center_upper}
\end{align}
Dividing~\eqref{eq:text_center_lower} by~\eqref{eq:vis_center_upper} yields~\eqref{eq:rho_gamma}.
\end{proof}

Lemma~\ref{lem:centred_lpd} shows that a signed non-visual gap yields a nonzero lower bound on the centered text-to-vision activation ratio.
The bound is intentionally conservative; the next theorem uses the actual centered ratio $\rho_t$ as the sufficient condition.

\begin{theorem}[Gradient-budget bias under uniform credit]
\label{thm:grad_bias}
Consider a positive-advantage local update with uniform credit $\hat A>0$.
Under Assumptions~\ref{asm:low_cancel} and~\ref{asm:jacobian_reg}, if a step $t$ satisfies
\begin{equation}
   \rho_t
   >
   \frac{C_J}{c_{\mathcal T}c_J},
   \label{eq:rho_threshold}
\end{equation}
then the non-visual gradient contribution dominates the visual contribution:
\begin{equation}
   \frac{
      \|G_t^{\mathcal T}\|_F
   }{
      \|G_t^{\mathcal V}\|_F
   }
   \ge
   \frac{c_{\mathcal T}c_J}{C_J}\rho_t
   >1.
   \label{eq:grad_bias}
\end{equation}
Consequently, for any subset $\mathcal H_\rho$ of steps satisfying~\eqref{eq:rho_threshold},
\begin{equation}
   \sum_{t\in\mathcal H_\rho}
   \|\hat A G_t^{\mathcal T}\|_F
   \ge
   \underline\kappa
   \sum_{t\in\mathcal H_\rho}
   \|\hat A G_t^{\mathcal V}\|_F,
   \qquad
   \underline\kappa
   =
   \min_{t\in\mathcal H_\rho}
   \frac{c_{\mathcal T}c_J}{C_J}\rho_t
   >1.
   \label{eq:grad_bias_sum}
\end{equation}
\end{theorem}

\begin{proof}
By Assumption~\ref{asm:low_cancel},
\begin{equation}
   \|G_t^{\mathcal T}\|_F
   \ge
   c_{\mathcal T}
   \sum_{j\in\mathcal T}
   \alpha_{tj}
   |d_t^\top v_j-\bar s_t|
   \left\|
       \frac{\partial e_{tj}}{\partial W_Q}
   \right\|_F.
   \label{eq:text_grad_lower}
\end{equation}
By the triangle inequality,
\begin{equation}
   \|G_t^{\mathcal V}\|_F
   \le
   \sum_{j\in\mathcal V}
   \alpha_{tj}
   |d_t^\top v_j-\bar s_t|
   \left\|
       \frac{\partial e_{tj}}{\partial W_Q}
   \right\|_F.
   \label{eq:vis_grad_upper}
\end{equation}
Using Assumption~\ref{asm:jacobian_reg}, the numerator in~\eqref{eq:text_grad_lower} is lower-bounded by
$c_{\mathcal T}c_JJ_t\sum_{j\in\mathcal T}\alpha_{tj}|d_t^\top v_j-\bar s_t|$,
while the denominator in~\eqref{eq:vis_grad_upper} is upper-bounded by
$C_JJ_t\sum_{j\in\mathcal V}\alpha_{tj}|d_t^\top v_j-\bar s_t|$.
Therefore
\[
   \frac{\|G_t^{\mathcal T}\|_F}{\|G_t^{\mathcal V}\|_F}
   \ge
   \frac{c_{\mathcal T}c_J}{C_J}\rho_t.
\]
Condition~\eqref{eq:rho_threshold} gives the strict dominance.
Multiplying by the same positive advantage $\hat A$ and summing over $\mathcal H_\rho$ gives~\eqref{eq:grad_bias_sum}.
\end{proof}

\begin{remark}[Interpretation]
\label{rem:grad_bias_interp}
Theorem~\ref{thm:grad_bias} is a sufficient-condition result.
It does not require the conservative lower bound $\gamma_t$ itself to exceed the threshold.
Instead, $\gamma_t$ explains how a signed non-visual gap contributes to the centered ratio $\rho_t$, while the actual gradient-budget dominance is controlled by $\rho_t$ together with cancellation and Jacobian-scale regularity.
\end{remark}

\begin{theorem}[Local visual-attention drift]
\label{thm:drift}
Consider a gradient-descent update on~\eqref{eq:uniform_pg} with $\hat A>0$:
\begin{equation}
   \Delta W_Q
   =
   -\eta\nabla_{W_Q}\mathcal L
   =
   \eta\hat A
   \sum_{t'=1}^{T}
   G_{t'}.
   \label{eq:wq_update}
\end{equation}
Define the same-step diagonal drift coefficient
\begin{equation}
   D_t
   \triangleq
   -
   \left\langle
       \frac{\partial\alpha_t^{\mathcal V}}{\partial W_Q},
       G_t
   \right\rangle_F,
   \label{eq:Dt_def}
\end{equation}
and the cross-step remainder
\begin{equation}
   R_t
   \triangleq
   \sum_{t'\ne t}
   \left\langle
       \frac{\partial\alpha_t^{\mathcal V}}{\partial W_Q},
       G_{t'}
   \right\rangle_F.
   \label{eq:Rt_def}
\end{equation}
Then the first-order change in visual attention mass is
\begin{equation}
   \Delta\alpha_t^{\mathcal V}
   =
   -\eta\hat A D_t
   +
   \eta\hat A R_t
   +
   O(\|\Delta W_Q\|_F^2).
   \label{eq:drift}
\end{equation}
If $t\in\mathcal H$ and Assumption~\ref{asm:key_metric} holds, then
\begin{equation}
   D_t
   \ge
   \zeta
   C_t
   \alpha_t^{\mathcal V}\alpha_t^{\mathcal T}
   (\mu_{\mathcal T,t}-\mu_{\mathcal V,t})
   >0.
   \label{eq:Dt_positive}
\end{equation}
Therefore, for positive-advantage trajectories, visual attention mass locally decreases whenever
\begin{equation}
   |R_t|<D_t.
   \label{eq:local_decrease_condition}
\end{equation}
\end{theorem}

\begin{proof}
A first-order Taylor expansion gives
\[
   \Delta\alpha_t^{\mathcal V}
   =
   \left\langle
       \frac{\partial\alpha_t^{\mathcal V}}{\partial W_Q},
       \Delta W_Q
   \right\rangle_F
   +
   O(\|\Delta W_Q\|_F^2).
\]
Substituting~\eqref{eq:wq_update} and separating the $t'=t$ term from the $t'\ne t$ terms yields
\[
   \Delta\alpha_t^{\mathcal V}
   =
   \eta\hat A
   \left\langle
       \frac{\partial\alpha_t^{\mathcal V}}{\partial W_Q},
       G_t
   \right\rangle_F
   +
   \eta\hat A
   \sum_{t'\ne t}
   \left\langle
       \frac{\partial\alpha_t^{\mathcal V}}{\partial W_Q},
       G_{t'}
   \right\rangle_F
   +
   O(\|\Delta W_Q\|_F^2).
\]
Using the definitions of $D_t$ and $R_t$ gives~\eqref{eq:drift}.
For $t\in\mathcal H$, Assumption~\ref{asm:key_metric} gives~\eqref{eq:Dt_positive}.
If $|R_t|<D_t$, the first-order term in~\eqref{eq:drift} is strictly negative, so $\alpha_t^{\mathcal V}$ decreases for sufficiently small effective update size.
\end{proof}

\begin{remark}[Scope of the drift result]
\label{rem:drift_scope}
Theorem~\ref{thm:drift} is a local first-order statement.
It does not require the cross-step term $R_t$ to vanish; it only states a sufficient regime in which the same-step diagonal drift dominates cross-step interference.
For $\hat A<0$, the first-order sign reverses.
In group-relative RLVR, the batch-level drift depends on the correlation between trajectory advantages and the local coefficients $D_t$ across sampled rollouts.
\end{remark}

\begin{remark}[Second-order term]
\label{rem:second_order}
The remainder in~\eqref{eq:drift} is controlled by the effective parameter displacement $\|\Delta W_Q\|_F$.
Thus the result should be interpreted in the standard local-update regime of policy-gradient analysis.
We do not claim that the first-order expression characterizes large-step or long-horizon training dynamics globally.
\end{remark}

\subsection{Effect of HiCCR Reweighting}
\label{sec:theory:correction}

HiCCR replaces the uniform positive-trajectory advantage by
\[
   \widetilde A_t^{(i)}
   =
   \alpha^{(i)}\beta_t^{(i)}\hat A^{(i)},
   \qquad
   \hat A^{(i)}>0,
\]
while leaving non-positive-advantage trajectories unmodified.
For a positive-advantage trajectory, the same first-order expansion gives
\begin{equation}
   \Delta\alpha_t^{\mathcal V}\big|_{\mathrm{HiCCR}}
   =
   -\eta
   \alpha^{(i)}
   \beta_t
   \hat A
   D_t
   +
   \eta
   \alpha^{(i)}
   \hat A
   R_t^{(\beta)}
   +
   O(\|\Delta W_Q\|_F^2),
   \label{eq:hiccr_drift}
\end{equation}
where
\begin{equation}
   R_t^{(\beta)}
   =
   \sum_{t'\ne t}
   \beta_{t'}
   \left\langle
       \frac{\partial\alpha_t^{\mathcal V}}{\partial W_Q},
       G_{t'}
   \right\rangle_F.
   \label{eq:hiccr_remainder}
\end{equation}
Thus, at the diagonal level, HiCCR changes the aggregate within-trajectory coefficient from
\begin{equation}
   S_1
   =
   \sum_t D_t
   \qquad
   \text{to}
   \qquad
   S_\beta
   =
   \sum_t \beta_t D_t.
   \label{eq:S_beta}
\end{equation}
Uniform credit corresponds to $\beta_t=1$.

Define
\[
   \mathcal N=\{t:D_t>0\},
   \qquad
   \mathcal P=\{t:D_t<0\}.
\]
Steps in $\mathcal N$ contribute to decreasing visual attention under positive advantage, whereas steps in $\mathcal P$ contribute in the opposite direction.
Under the diagonal approximation, HiCCR reverses the aggregate visual-attention drift if
\begin{equation}
   \sum_{t\in\mathcal P}
   \beta_t |D_t|
   \ge
   \sum_{t\in\mathcal N}
   \beta_t D_t.
   \label{eq:hiccr_reversal}
\end{equation}
It mitigates the magnitude of a negative uniform drift whenever
\begin{equation}
   0
   \le
   S_\beta
   <
   S_1.
   \label{eq:hiccr_mitigation}
\end{equation}

HiCCR is designed to make such mitigation more likely by assigning larger $\beta_t$ to tokens with stronger visual attention concentration.
This proxy does not assert that attention concentration is a causal saliency measure for every token; rather, it provides an efficient operational signal for redistributing positive credit toward visually engaged generation steps.
The trajectory-level factor $\alpha^{(i)}$ is constant within a trajectory and therefore does not change the sign of $S_\beta$ for that trajectory.
Its role is batch-level: among positive-advantage rollouts, it increases the contribution of trajectories with stronger overall visual engagement.

\begin{remark}[Empirical justification of the $\beta_t$ proxy]
\label{rem:beta_design}
The analysis in~\eqref{eq:S_beta}--\eqref{eq:hiccr_mitigation} shows that any reweighting that assigns larger $\beta_t$ to steps with larger positive $D_t$ (or smaller negative $D_t$) will reduce $S_\beta$ relative to the uniform-credit baseline $S_1$.
It does not uniquely determine the functional form of $\beta_t$.
HiCCR instantiates $\beta_t$ via visual attention concentration $C_{\mathcal V}(t)$ as a cheap, differentiable proxy available at every rollout step without additional forward passes.
This design choice is not a consequence of the theorem but an empirical one:
Section~\ref{sec:experiments} (Table~\ref{tab:table4_credit_assignment}) compares the attention-based $\beta_t$ against random, uniform, and gradient-magnitude alternatives under identical $\alpha^{(i)}$, and shows that the attention-based proxy yields the strongest downstream gains.
Thus the role of the theoretical analysis is to identify \emph{the direction of a useful reweighting}, while the particular proxy is selected by ablation.
\end{remark}

\begin{remark}[Positive-advantage asymmetry]
\label{rem:positive_advantage}
The above correction applies only to $\hat A>0$, matching the design of HiCCR.
Applying the same amplification to negative-advantage trajectories would also amplify penalties on visually engaged but incorrect rollouts, which can discourage visual use.
This is why HiCCR reassigns credit asymmetrically: it amplifies visually engaged successful trajectories while leaving unsuccessful trajectories under the base RLVR signal. The related experiments have confirm this conclusion~\ref{tab:asymmetric_shaping}.
\end{remark}

\subsection{Multi-Head and Multilayer Scope}
\label{sec:theory:saliency}

The preceding analysis is a single-head, single-layer local component analysis.
In a full transformer, attention-gradient contributions add across heads and layers but are coupled through output projections, residual streams, and subsequent nonlinear blocks.
Therefore, we do not claim a headwise theorem automatically becomes a full-network theorem without additional assumptions.

The mechanism nevertheless extends qualitatively under limited cross-head and cross-layer cancellation.
If a non-vanishing subset of heads/layers satisfies the centered-dominance and local key-metric conditions, their contributions create a text-favored component in the aggregate update.
Conversely, heads/layers that do not satisfy these conditions can weaken or cancel the effect.
This is why the empirical analysis of $\kappa_{\mathrm{emp}}$ across layers is used as support for where the mechanism is likely active, rather than as a proof that every layer satisfies the same condition.

In practice, HiCCR computes its context-credit score from late decoder layers, where the measured non-visual activation dominance remains consistently above one and where token predictions are closer to the final policy decision.
This motivates the choice of the target layer set $\mathcal L^\star$ for computing $C_{\mathcal V}(t)$ in Section~\ref{sec:method}.

\section{On the Validity of the Context Credit Score}
\label{sec:saliency_proxy}

This section clarifies why $C_{\mathcal{V}}(t)$ is used as the token-level credit signal in HiCCR.
We do not claim that it is a gradient-based saliency score or a token-level causal explanation.
Instead, we use it as an operational measure of visual-attention concentration and evaluate whether it provides a useful signal for visually grounded credit reassignment.

\subsection{Definition}
\label{subsec:cv_definition}

Recall that
\begin{equation}
\label{eq:cv_appendix_b}
    C_{\mathcal{V}}(t)
    =
    1 - \frac{H\!\bigl(p_{\mathcal{V}}(\cdot\mid t)\bigr)}{\log|\mathcal{V}|},
    \qquad
    p_{\mathcal{V}}(j\mid t)\propto \exp(e_{tj}/\tau),\;\; j\in\mathcal{V}.
\end{equation}
Here $p_{\mathcal{V}}(\cdot\mid t)$ is the attention distribution from generation token $t$ to visual positions.
Thus, $C_{\mathcal{V}}(t)$ is high when a token selectively attends to a small subset of visual tokens, and low when its visual attention is diffuse.
We interpret this score as a lightweight proxy for selective visual querying, rather than as a direct measure of causal saliency.

\subsection{Ablation Evidence}
\label{subsec:cv_ablation}

Table~\ref{tab:table4_credit_assignment} tests whether the proposed score is more useful than alternative credit signals under the same HiCCR pipeline.
We compare Uniform Allocation, Random Allocation, Time-decay, Entropy-mask, Text-Context-based HiCCR-T, Raw Attention Energy, and the proposed Context-based HiCCR.

The results support three observations.
First, Random Allocation improves over Uniform Allocation, suggesting that uniform token credit is overly coarse.
Second, Time-decay, Entropy-mask, and HiCCR-T are weaker or unstable, indicating that generic non-uniformity, hard token selection, and textual-context concentration are insufficient.
Third, Raw Attention Energy improves over several heuristics but remains below $C_{\mathcal{V}}(t)$, showing that raw visual-attention magnitude is less effective than normalized concentration within the visual token subset.

These comparisons support a limited conclusion: among the tested alternatives, visual-attention concentration provides the most effective credit signal.
They do not imply that $C_{\mathcal{V}}(t)$ perfectly measures token-level visual dependence.

\subsection{Semantic and Counterfactual Evidence}
\label{app:qualitative_causal}

Beyond credit-score ablations, we further examine whether the learned credit weights align with visually related tokens.
As shown in Figure~\ref{fig:credit_analysis}(e), semantically visual tokens receive higher credit weights than non-visual tokens, with a median weight of $1.303$ versus $1.236$ ($p<0.001$, Cohen's $d=0.547$).
\textbf{This result suggests that the proposed credit score is not merely assigning arbitrary non-uniform weights, but tends to amplify tokens that are semantically associated with visual reasoning}.
We emphasize that this is a semantic alignment test rather than a token-level causal proof.

We also conduct a model-level visual ablation in Appendix~\ref{app:counterfactual}.
During inference, we zero out the visual embeddings and measure the resulting accuracy drop across benchmarks.
As shown in Figure~\ref{fig:visual_ablation}, HiCCR variants show larger degradation than their corresponding uniform-credit RLVR baselines across both 4B and 8B scales.
This suggests that HiCCR's gains are accompanied by stronger functional reliance on visual input at the model level.

Together, the credit-score ablations, semantic token analysis, and visual counterfactual results support $C_{\mathcal{V}}(t)$ as a practical proxy for visually grounded credit reassignment.
They do not require the stronger claim that attention concentration is equivalent to gradient-based saliency or token-level causal dependence.



\section{Mechanistic Interpretability: From Attention to Functional Dependence}
\label{app:empirical_grounding}

A fundamental critique of attention-guided methods is the well-known ``attention is not explanation'' debate~\citep{jain2019attention}.
To rigorously verify whether HiCCR genuinely alters the model's functional dependence on visual inputs---rather than merely manipulating superficial attention patterns---we conduct a three-stage mechanistic and causal analysis with increasing evidential strength:
(i)~\textbf{gradient-based attribution} quantifies how much each modality contributes to the policy gradient (\S\ref{app:gradient_attribution});
(ii)~\textbf{counterfactual ablation} removes visual input entirely and measures performance collapse, establishing a causal link (\S\ref{app:counterfactual});
and (iii)~\textbf{attention dynamics} traces the temporal evolution of visual attention across generation steps, revealing how different training objectives reshape the model's reliance on image tokens over the course of a rollout (\S\ref{app:attention_dynamics}).

\subsection{Gradient-Based Attribution Analysis}
\label{app:gradient_attribution}

We ask: \textit{Does the preservation of visual attention in HiCCR translate to actual gradient contribution?}
Using the Input$\times$Gradient method~\citep{sundararajan2017axiomaticattributiondeepnetworks}, we compute, for each generated token, the fraction of the gradient's L2 norm derived from visual embeddings relative to the entire context, then average over all tokens and prompts per benchmark.

As shown in Table~\ref{tab:gradient_attribution}, uniform RLVR baselines consistently \emph{diminish} visual gradient attribution relative to the Base model (e.g., DAPO reduces the 4B average from 25.0\% to 22.7\%, and the 8B average from 22.0\% to 20.8\%), confirming the gradient-budget bias predicted by Theorem~\ref{thm:grad_bias}.
HiCCR reverses this degradation: HiCCR-G restores the average to 25.8\% (4B) and 26.1\% (8B), with the largest gains on geometry- and diagram-intensive benchmarks where continuous visual extraction is indispensable---Geo3K improves by +5.3\,pp (4B) and +6.1\,pp (8B) over DAPO, and WeMath by +4.0\,pp and +6.6\,pp respectively.
Notably, the attribution gains are consistent across both HiCCR variants and both model scales, confirming that the credit reassignment mechanism reliably redirects optimization pressure toward the visual modality.

\begin{table}[htbp]
\centering
\caption{\textbf{Gradient Attribution Analysis.}
Image Attr.\,(\%): average fraction of total Input$\times$Gradient
attribution originating from visual tokens.
Best results compared to respective uniform baselines are in \textbf{bold}.}
\label{tab:gradient_attribution}
\resizebox{\textwidth}{!}{
\begin{tabular}{l cccccccccc @{\hskip 1.5em} c}
\toprule
\textbf{Method} & \textbf{Geo3K} & \textbf{MathV} & \textbf{MVista} & \textbf{WeMath} & \textbf{MMK12} & \textbf{CLEVR} & \textbf{Logic} & \textbf{MMMU} & \textbf{Hallus.} & \textbf{MMStar} & \textbf{Avg} \\
\midrule
\multicolumn{12}{c}{\textit{Qwen3-VL-4B Backbone}} \\
Base    & 18.7 & 21.6 & 21.2 & 16.1 & 13.5 & \textbf{28.5} & 26.0 & 53.0 & \textbf{32.3} & 18.8 & 25.0 \\
GRPO   & 17.0 & 20.3 & 20.9 & 15.0 & 12.8 & 25.8 & 25.8 & 52.5 & 31.8 & 17.4 & 23.9 \\
DAPO   & 16.7 & 19.6 & 20.6 & 14.2 & 13.5 & 27.5 & 24.1 & 46.6 & 28.5 & 15.8 & 22.7 \\
\rowcolor{gray!10} \textbf{HiCCR-G} & 22.0 & \textbf{23.4} & \textbf{22.0} & \textbf{18.2} & \textbf{15.3} & 27.9 & 26.9 & \textbf{56.1} & 27.1 & \textbf{19.5} & \textbf{25.8} \\
\rowcolor{gray!10} \textbf{HiCCR-D} & \textbf{22.6} & 23.3 & 21.8 & 17.5 & 14.5 & 27.5 & \textbf{27.6} & \textbf{56.2} & 31.6 & 18.4 & \textbf{26.1} \\
\midrule
\multicolumn{12}{c}{\textit{Qwen3-VL-8B Backbone}} \\
Base    & 17.5 & 20.8 & 18.9 & 13.6 & 11.8 & 21.4 & 23.8 & 52.4 & 25.3 & 14.9 & 22.0 \\
GRPO   & 20.4 & 22.2 & 20.3 & 17.0 & 13.3 & 21.4 & 28.0 & 55.6 & 29.4 & 14.6 & 24.2 \\
DAPO   & 15.8 & 18.6 & 17.8 & 13.7 & 11.2 & 19.8 & 23.4 & 49.2 & 24.9 & 14.1 & 20.8 \\
\rowcolor{gray!10} \textbf{HiCCR-G} & \textbf{21.9} & \textbf{22.8} & \textbf{22.3} & \textbf{20.3} & \textbf{14.7} & \textbf{25.3} & \textbf{30.5} & \textbf{55.7} & \textbf{31.0} & \textbf{16.7} & \textbf{26.1} \\
\rowcolor{gray!10} \textbf{HiCCR-D} & 17.2 & 17.9 & 18.9 & 15.4 & 13.3 & 24.8 & 26.7 & 46.2 & 29.7 & 14.2 & \textbf{22.4} \\
\bottomrule
\end{tabular}}
\end{table}
\subsection{Counterfactual Visual Ablation}
\label{app:counterfactual}

Gradient attribution reveals how the model \emph{uses} visual information, but does not establish whether visual input is \emph{necessary}.
To establish stronger causal evidence between our credit reassignment and functional visual dependence, we perform a counterfactual visual ablation study.
During inference, we corrupt the visual embeddings by applying \textit{Zero-Ablation} (zeroing out visual tokens), forcing the model to rely entirely on its textual priors and previously generated context.
The rationale is straightforward: if a model genuinely grounds its reasoning in the visual input, destroying that input should result in a catastrophic performance drop.
Conversely, if a model relies on textual shortcuts or hallucination, the performance drop will be minimal (or even negative).

\begin{figure}[htbp]
    \centering
    \includegraphics[width=\linewidth]{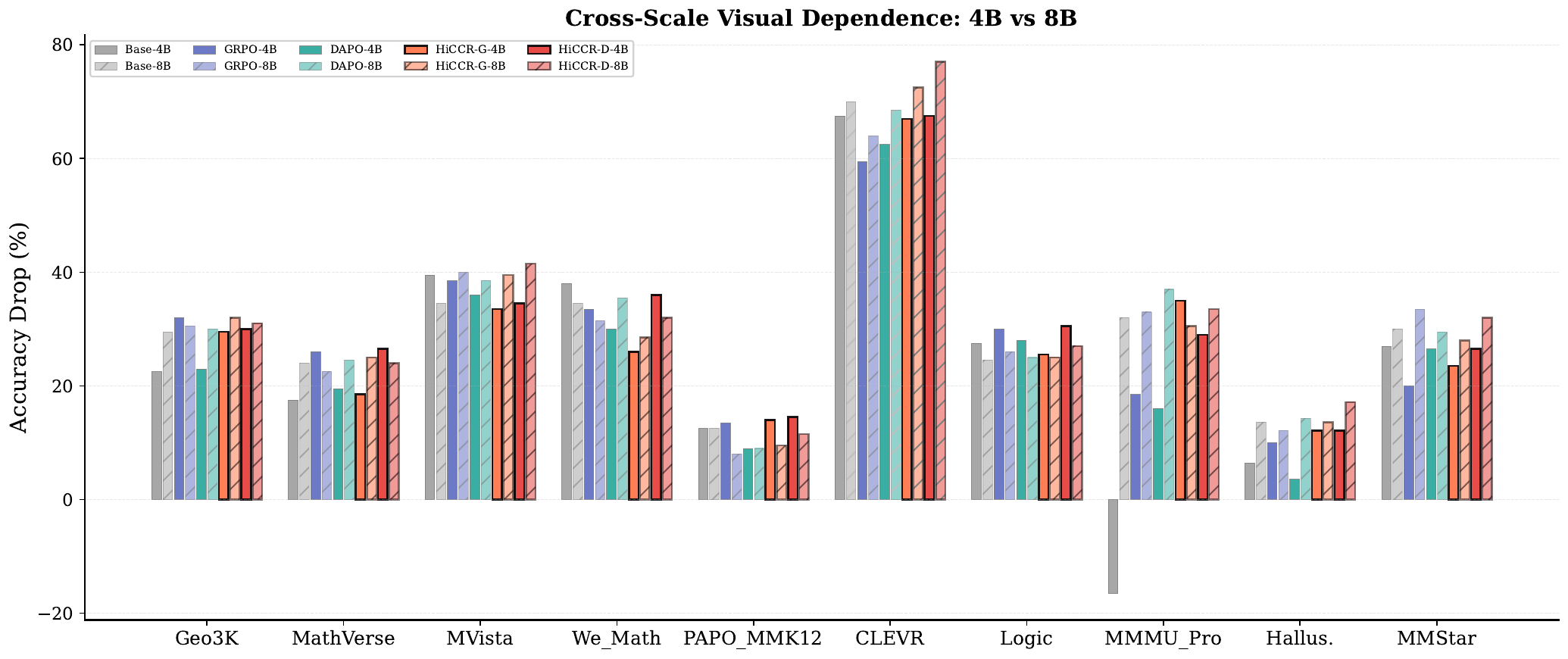}
    \caption{\textbf{Cross-Scale Counterfactual Visual Ablation.} We report the accuracy drop (\%) when visual embeddings are zeroed out during inference. A larger drop indicates stronger functional dependence on visual inputs. Across both 4B (solid) and 8B (hatched) scales, HiCCR variants consistently exhibit the most severe performance degradation upon visual corruption, confirming that our method substantially enforces stronger visual functional dependence.}
    \label{fig:visual_ablation}
\end{figure}

As illustrated in Figure~\ref{fig:visual_ablation}, while most models suffer to some extent when visual evidence is destroyed, \textbf{HiCCR variants (HiCCR-G and HiCCR-D) exhibit a steeper accuracy drop ($\Delta$) compared to both the Base models and uniform baselines across both 4B and 8B scales}.

This disparity is most striking in highly complex benchmarks.
For instance, on MMMU\_Pro, the Base-4B model paradoxically shows a negative drop (i.e., performance slightly improves when the image is removed), exposing a severe reliance on textual hallucination rather than visual comprehension.
In contrast, HiCCR models consistently suffer the largest accuracy penalties when blinded, strongly suggesting that their superior performance stems from functional visual grounding.
This profound vulnerability to visual corruption provides stronger causal evidence: HiCCR's optimization explicitly forces the policy to ground its reasoning in the actual image.

Together with the gradient evidence in \S\ref{app:gradient_attribution}, this ablation establishes both a correlational and a causal link between HiCCR's credit-reassignment mechanism and enhanced visual functional dependence.

\subsection{Attention Dynamics During Generation}
\label{app:attention_dynamics}

We examine how visual attention evolves during generation.
As predicted by Theorem~\ref{thm:drift}, uniform credit assignment can progressively shift policy updates away from image tokens, leading to attention drift.
This subsection provides temporal evidence for this drift and shows that HiCCR counteracts it by reallocating credit toward visually engaged tokens and trajectories.

\paragraph{Protocol.}
For each model and method, we sample 500 instances and track the normalized attention mass at every decoding step over three context components: image tokens, prompt tokens, and prefix tokens.
All trajectories are interpolated onto a uniform 200-point grid and visualized with mean and standard deviation bands.
We evaluate six representative benchmarks covering mathematical reasoning, vision-dependent reasoning, and general multimodal understanding: Geometry3K, MathVerse, MathVerse-VD, CLEVR-Count, MMMU-Pro, and HallusionBench.
Results are reported under four settings: RLVR baseline comparisons at 8B and 4B, and base-model comparisons at 8B and 4B.
In all figures, solid lines denote image attention, dashed lines denote prompt attention, and dotted lines denote prefix attention; shaded regions mark intervals where HiCCR image attention exceeds all compared baselines.

\paragraph{Main findings.}
Three findings are consistent across benchmarks and model scales.
First, the pretrained model already shows decreasing image attention as generation proceeds, indicating that visual under-utilization is an inherited tendency of the backbone.
Second, standard RLVR methods such as GRPO and DAPO amplify this tendency: they improve answer accuracy but further reduce sustained attention to image tokens and increase reliance on generated prefix tokens.
Third, HiCCR reverses this pattern.
Across all settings, HiCCR maintains higher image attention than GRPO, DAPO, and other RLVR baselines, often forming a stable visual attention plateau throughout generation.
The gain mainly comes from reduced prefix reliance while prompt attention remains comparable, suggesting that HiCCR encourages the model to repeatedly consult visual evidence rather than reinforce its own textual continuation.

\subsubsection{RLVR Baselines at 8B}
\label{app:attn_rlvr_8b}

\begin{figure}[htbp]
  \centering
  \begin{subfigure}[b]{0.32\textwidth}
    \includegraphics[width=\textwidth]{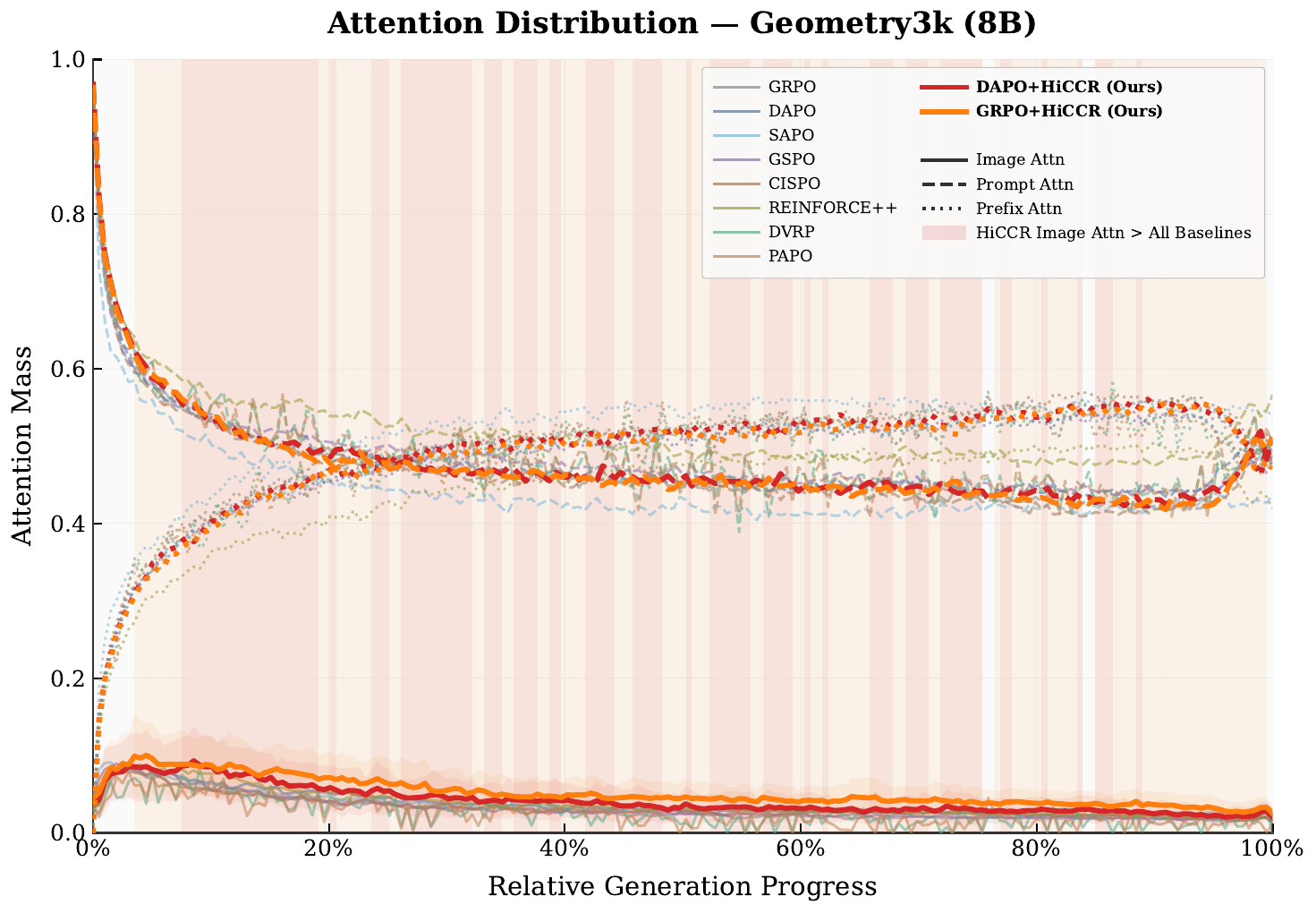}
    \caption{Geometry3K}
  \end{subfigure}\hfill
  \begin{subfigure}[b]{0.32\textwidth}
    \includegraphics[width=\textwidth]{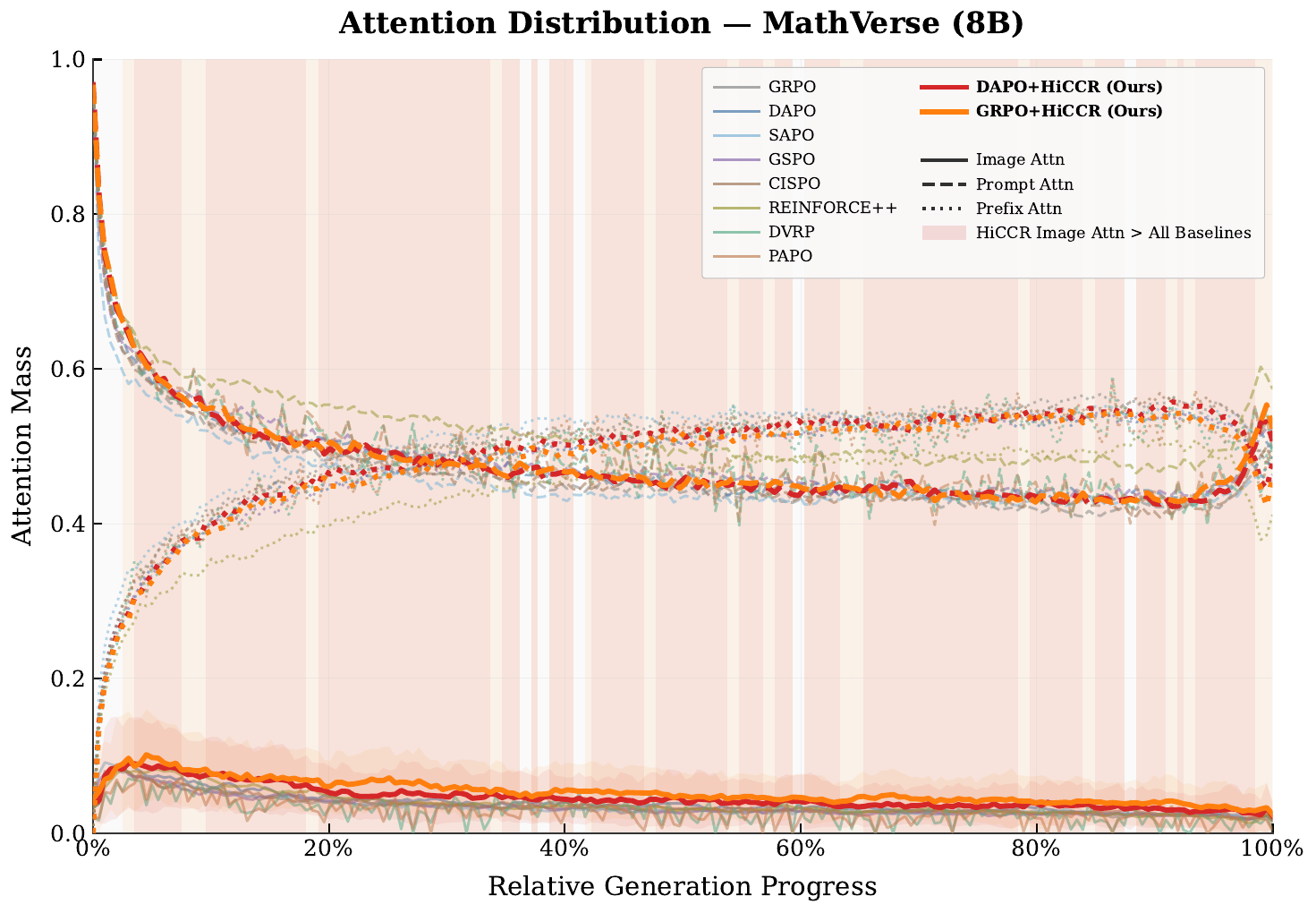}
    \caption{MathVerse}
  \end{subfigure}\hfill
  \begin{subfigure}[b]{0.32\textwidth}
    \includegraphics[width=\textwidth]{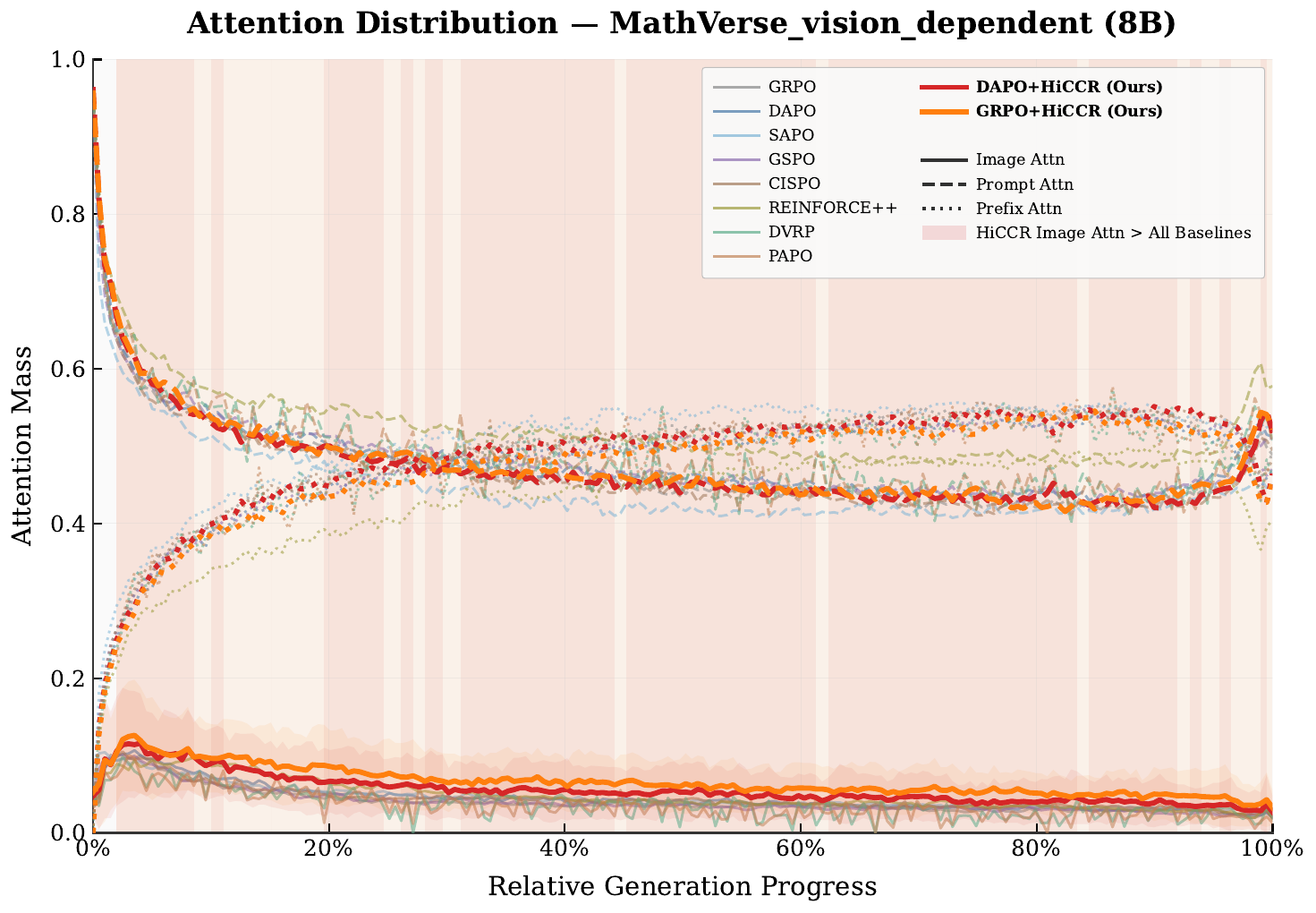}
    \caption{MathVerse-VD}
  \end{subfigure}

  \vspace{0.2cm}
  \begin{subfigure}[b]{0.32\textwidth}
    \includegraphics[width=\textwidth]{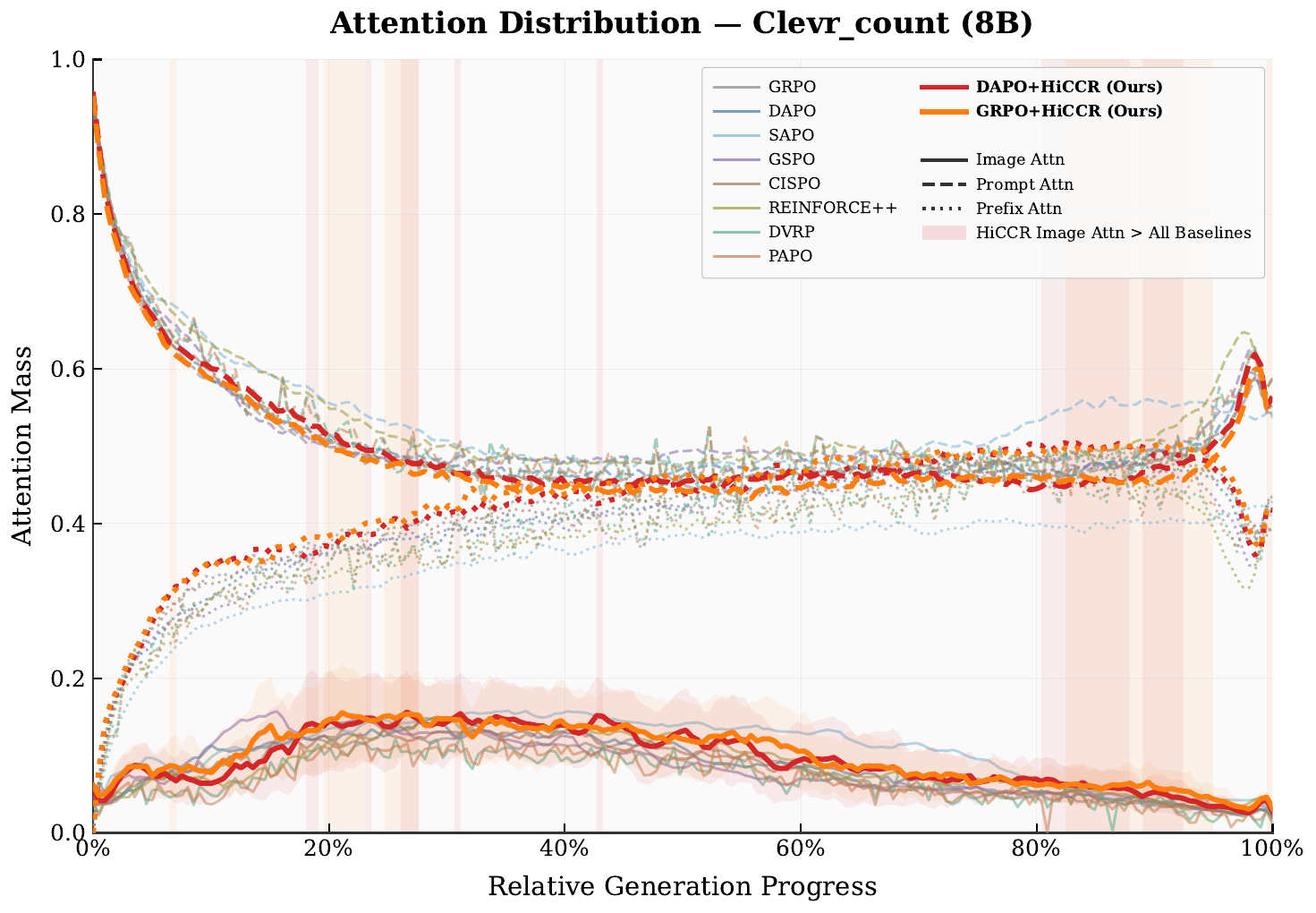}
    \caption{CLEVR-Count}
  \end{subfigure}\hfill
  \begin{subfigure}[b]{0.32\textwidth}
    \includegraphics[width=\textwidth]{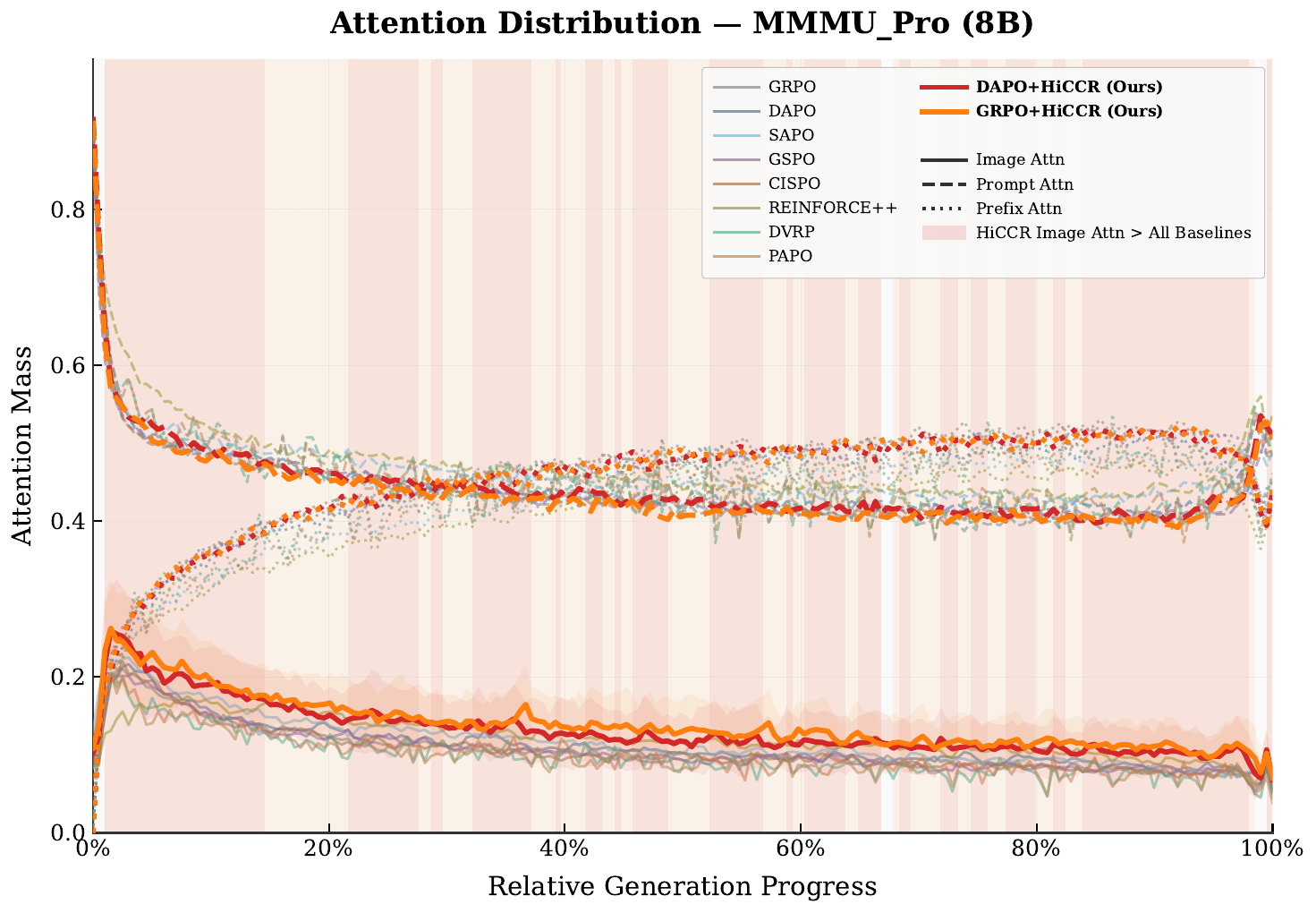}
    \caption{MMMU-Pro}
  \end{subfigure}\hfill
  \begin{subfigure}[b]{0.32\textwidth}
    \includegraphics[width=\textwidth]{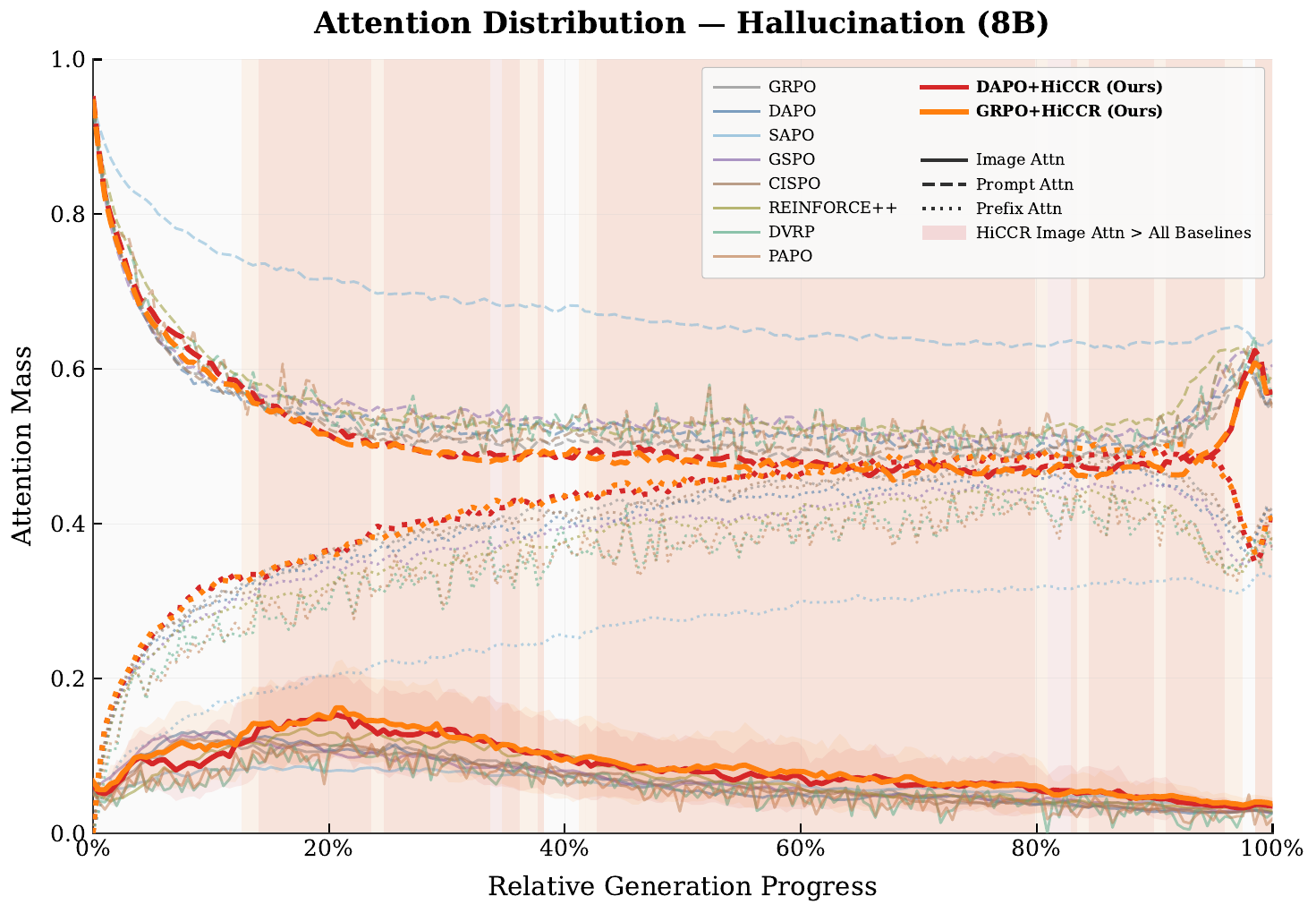}
    \caption{HallusionBench}
  \end{subfigure}
  \caption{\textbf{8B RLVR attention dynamics.}
  HiCCR sustains higher image attention across benchmarks, while GRPO, DAPO, and related RLVR baselines exhibit progressive visual-attention drift.}
  \label{fig:attn_rlvr_8b_all}
\end{figure}

Figure~\ref{fig:attn_rlvr_8b_all} compares HiCCR with nine RLVR baselines on Qwen3-VL-8B.
Across all six benchmarks, GRPO, DAPO, and related RLVR methods show a clear decline in image attention from early to late generation.
In contrast, HiCCR variants sustain higher image attention and exceed all baselines over long decoding intervals.
The advantage is strongest on vision-dependent tasks such as MathVerse-VD and CLEVR-Count, where correct reasoning requires repeated extraction of visual evidence.
Prompt attention remains comparable across methods, while prefix attention is lower under HiCCR.
This indicates that HiCCR improves visual grounding not by weakening prompt processing, but by reducing self-reinforcing dependence on generated text.

\subsubsection{RLVR Baselines at 4B}
\label{app:attn_rlvr_4b}

\begin{figure}[htbp]
  \centering
  \begin{subfigure}[b]{0.32\textwidth}
    \includegraphics[width=\textwidth]{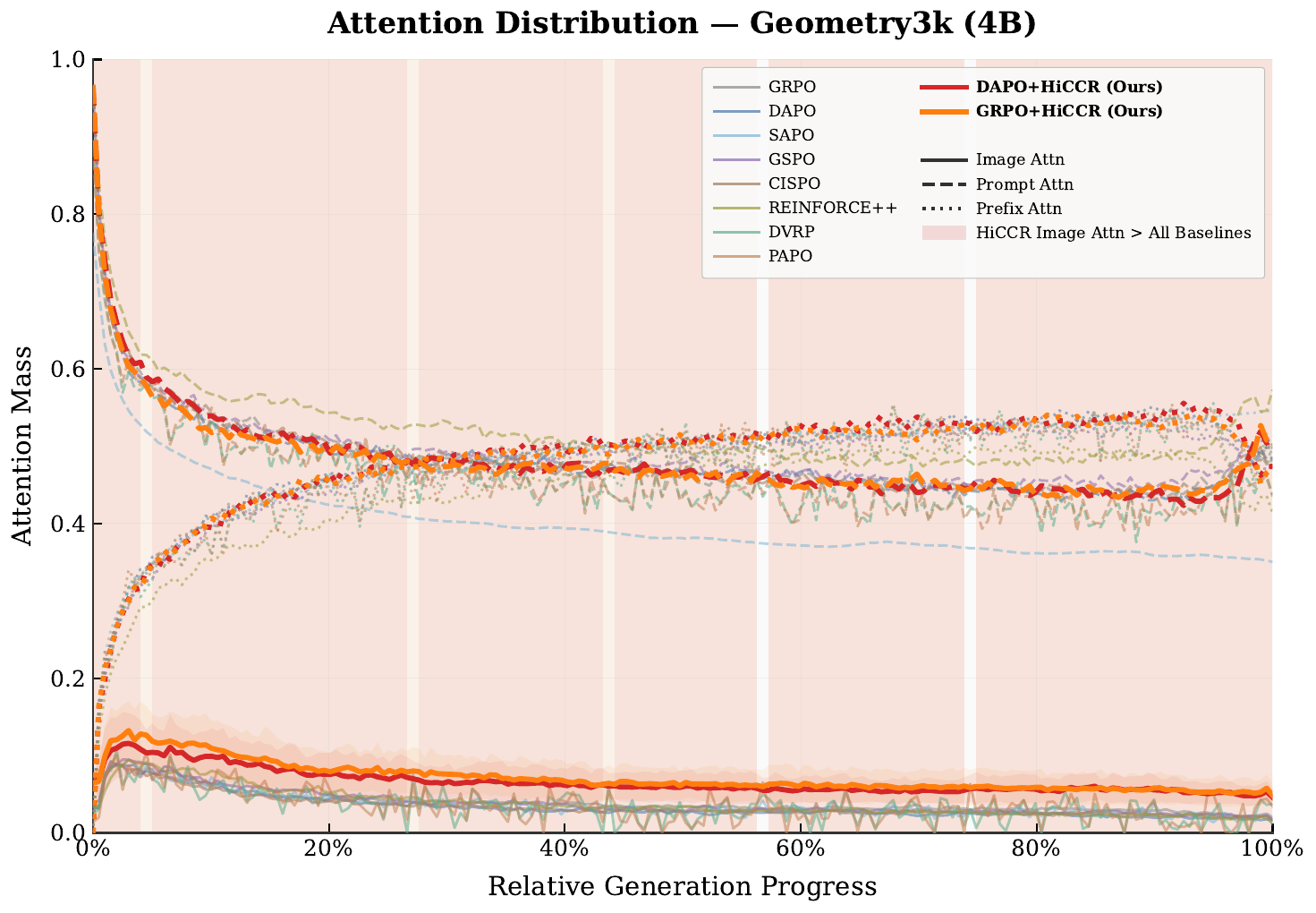}
    \caption{Geometry3K}
  \end{subfigure}\hfill
  \begin{subfigure}[b]{0.32\textwidth}
    \includegraphics[width=\textwidth]{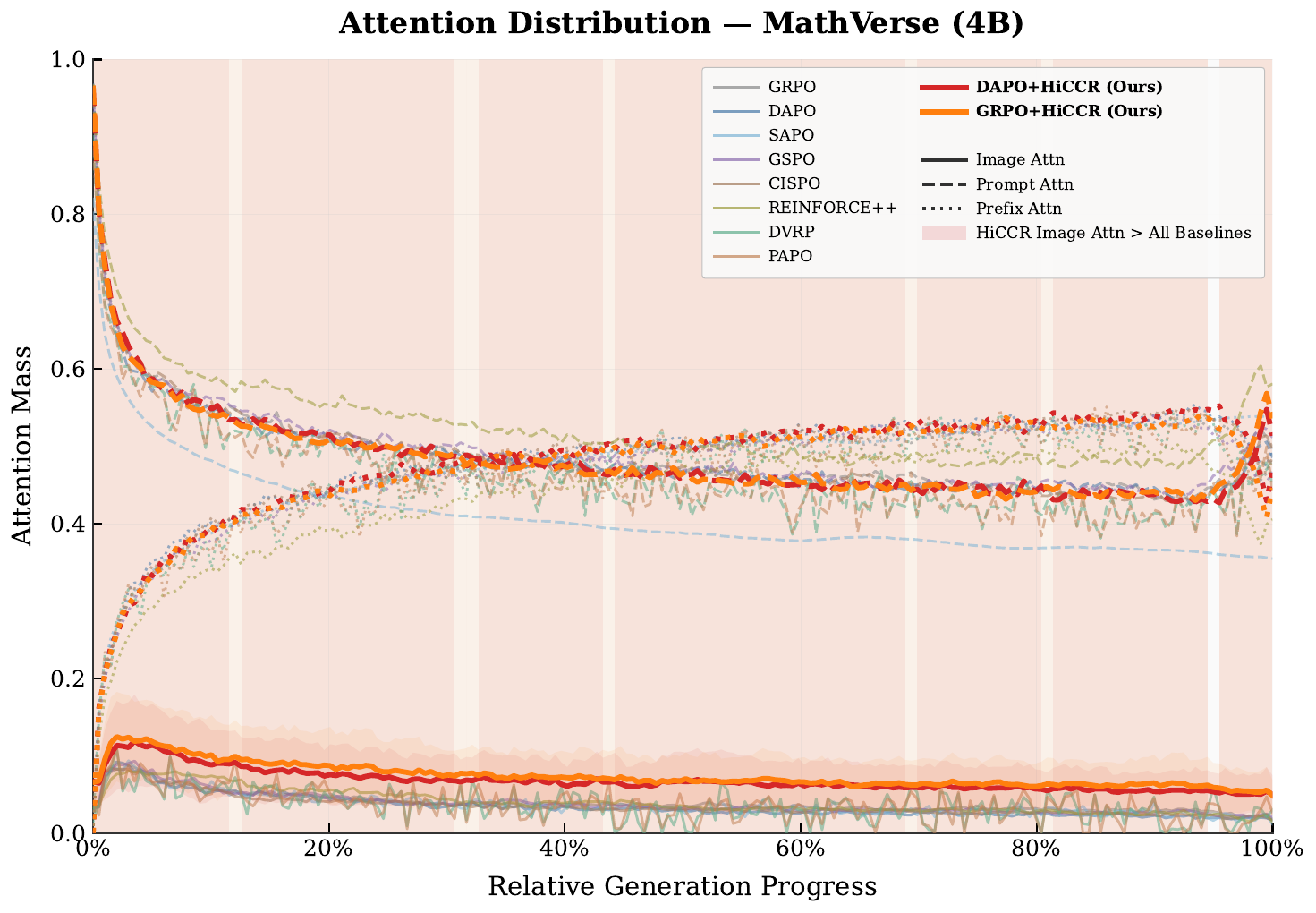}
    \caption{MathVerse}
  \end{subfigure}\hfill
  \begin{subfigure}[b]{0.32\textwidth}
    \includegraphics[width=\textwidth]{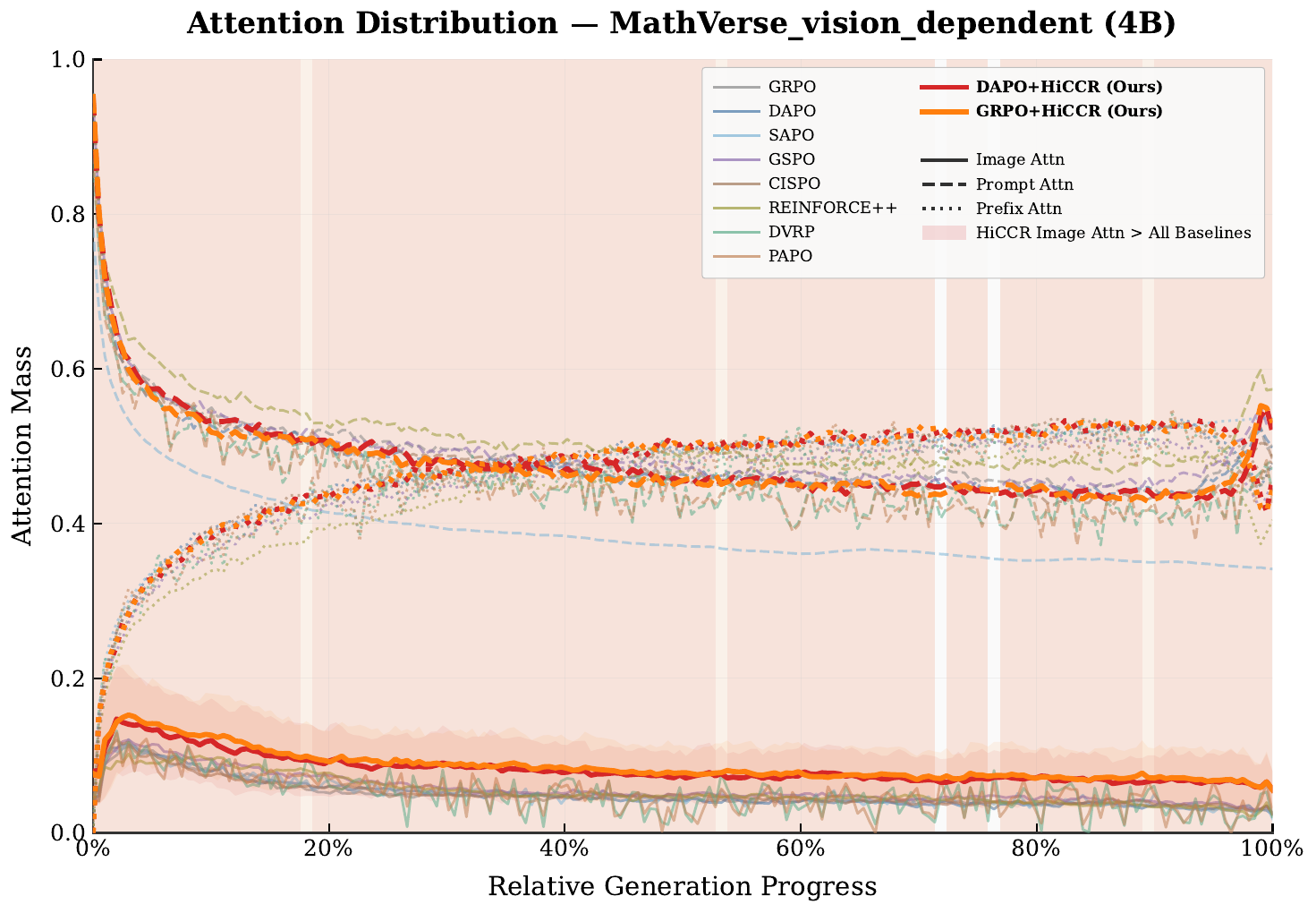}
    \caption{MathVerse-VD}
  \end{subfigure}

  \vspace{0.2cm}
  \begin{subfigure}[b]{0.32\textwidth}
    \includegraphics[width=\textwidth]{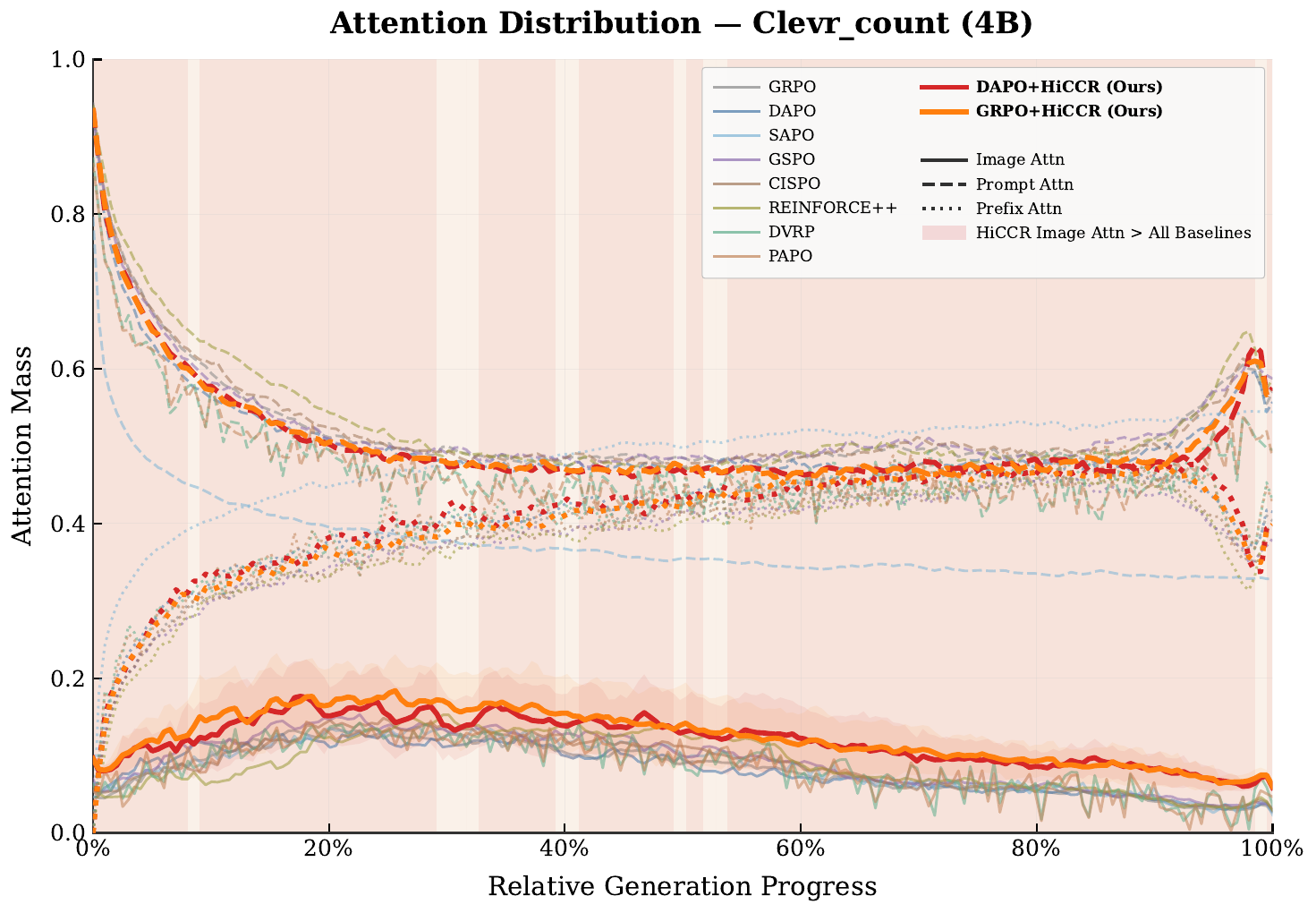}
    \caption{CLEVR-Count}
  \end{subfigure}\hfill
  \begin{subfigure}[b]{0.32\textwidth}
    \includegraphics[width=\textwidth]{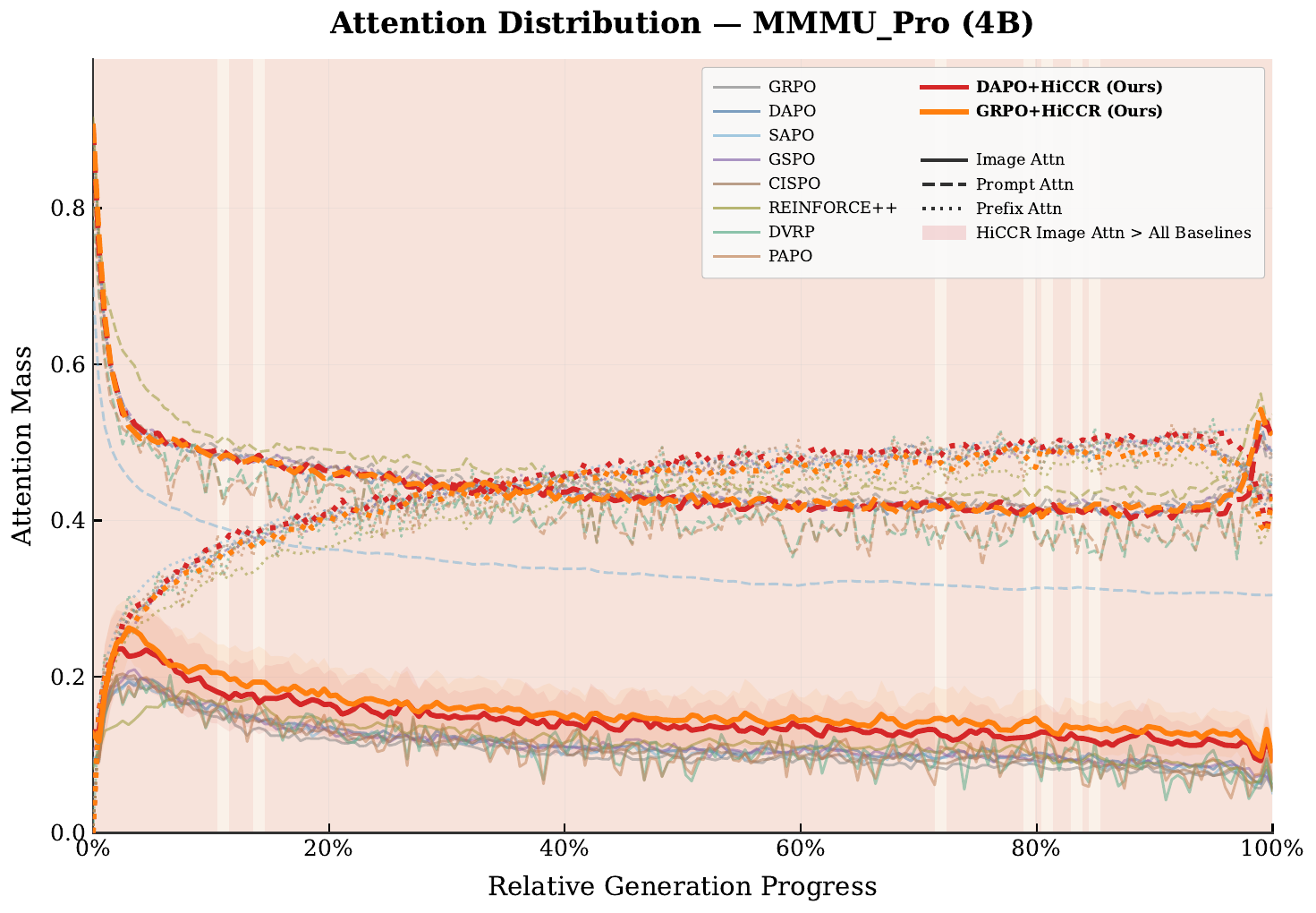}
    \caption{MMMU-Pro}
  \end{subfigure}\hfill
  \begin{subfigure}[b]{0.32\textwidth}
    \includegraphics[width=\textwidth]{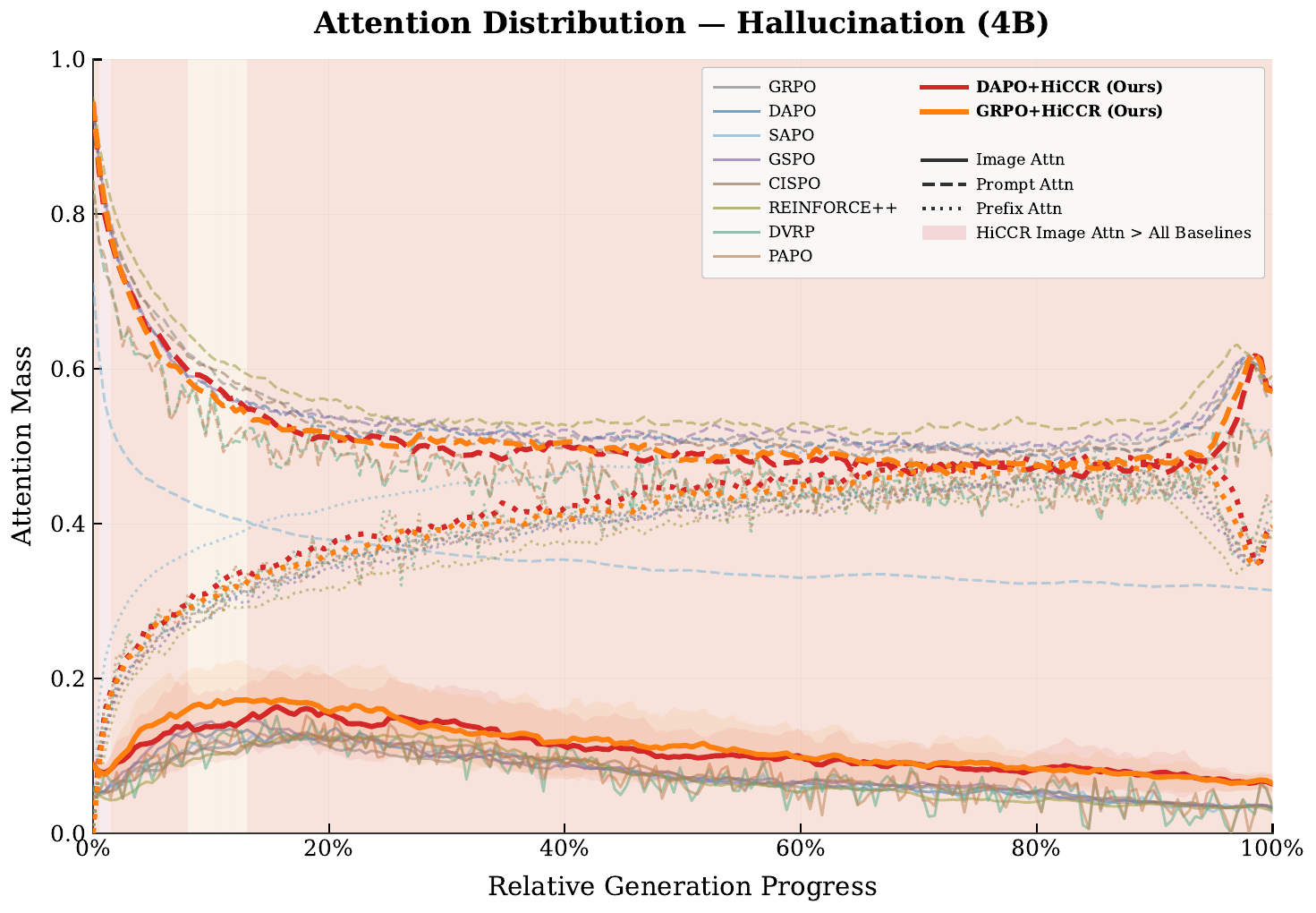}
    \caption{HallusionBench}
  \end{subfigure}
  \caption{\textbf{4B RLVR attention dynamics.}
  Attention drift becomes stronger at the smaller scale, but HiCCR still preserves a stable visual attention plateau beyond all RLVR baselines.}
  \label{fig:attn_rlvr_4b_all}
\end{figure}

We repeat the comparison on Qwen3-VL-4B to test whether the same dynamics hold under tighter capacity constraints.
Figure~\ref{fig:attn_rlvr_4b_all} shows that attention drift becomes more severe at the smaller scale.
The 4B RLVR baselines lose image attention more rapidly than their 8B counterparts, suggesting that limited capacity makes the model more vulnerable to language-prior amplification during RLVR.
Despite this stronger drift, HiCCR again maintains a persistent visual attention plateau and surpasses the RLVR baselines across benchmarks.
This consistency indicates that sustained visual grounding is not a byproduct of model capacity, but a direct effect of credit reassignment.

\subsubsection{Base Model Comparisons at 8B}
\label{app:attn_base_8b}

\begin{figure}[htbp]
  \centering
  \begin{subfigure}[b]{0.32\textwidth}
    \includegraphics[width=\textwidth]{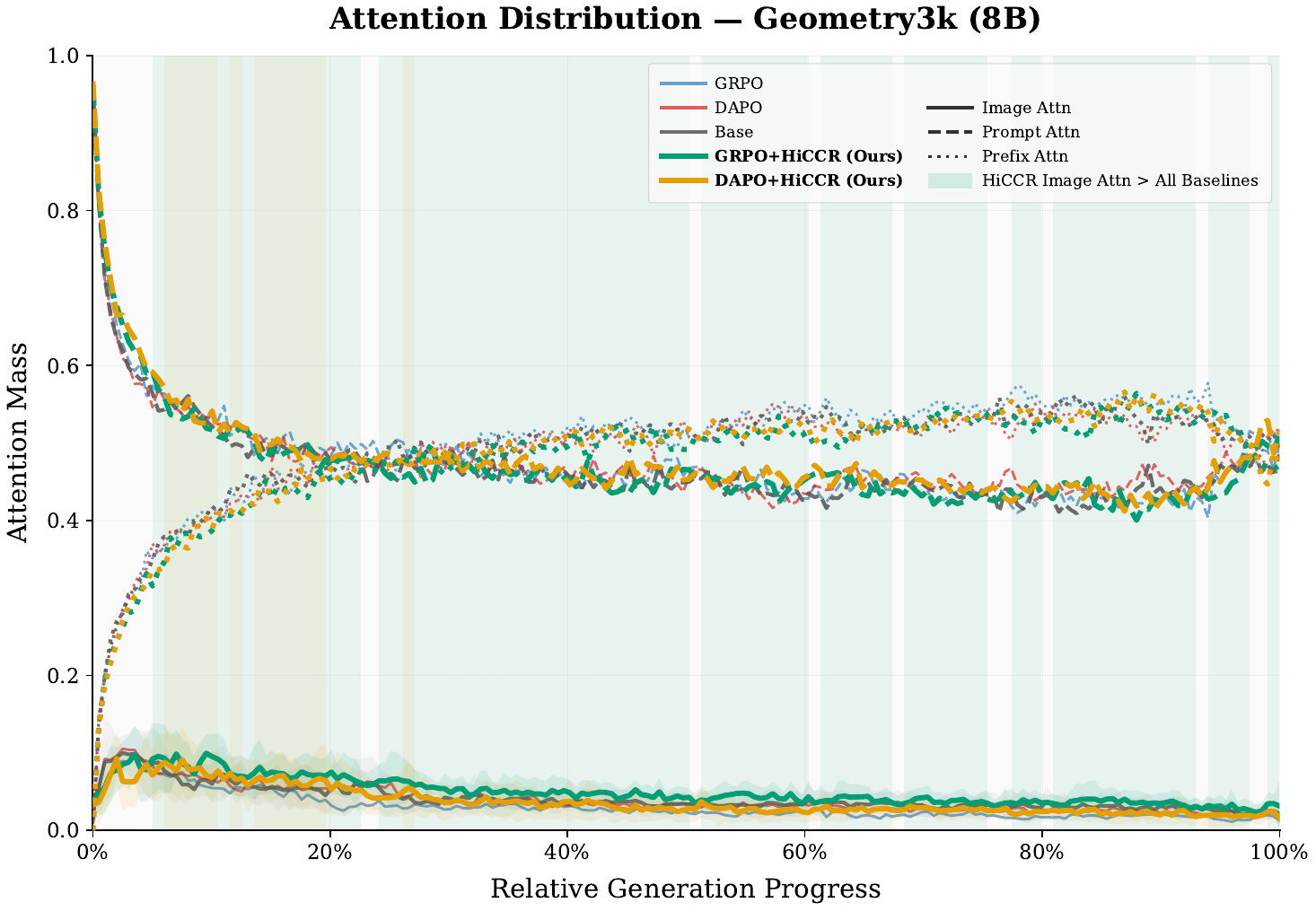}
    \caption{Geometry3K}
  \end{subfigure}\hfill
  \begin{subfigure}[b]{0.32\textwidth}
    \includegraphics[width=\textwidth]{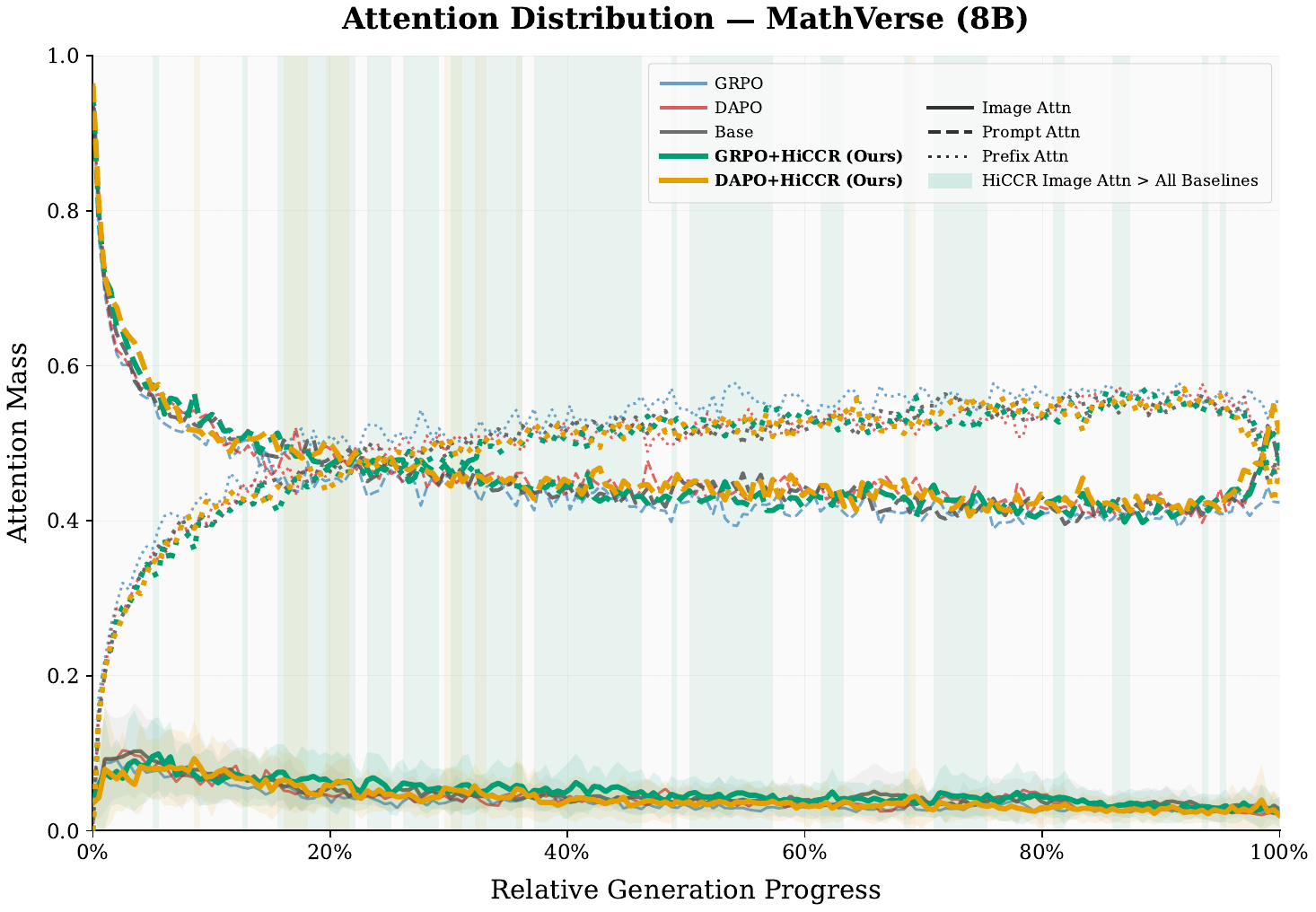}
    \caption{MathVerse}
  \end{subfigure}\hfill
  \begin{subfigure}[b]{0.32\textwidth}
    \includegraphics[width=\textwidth]{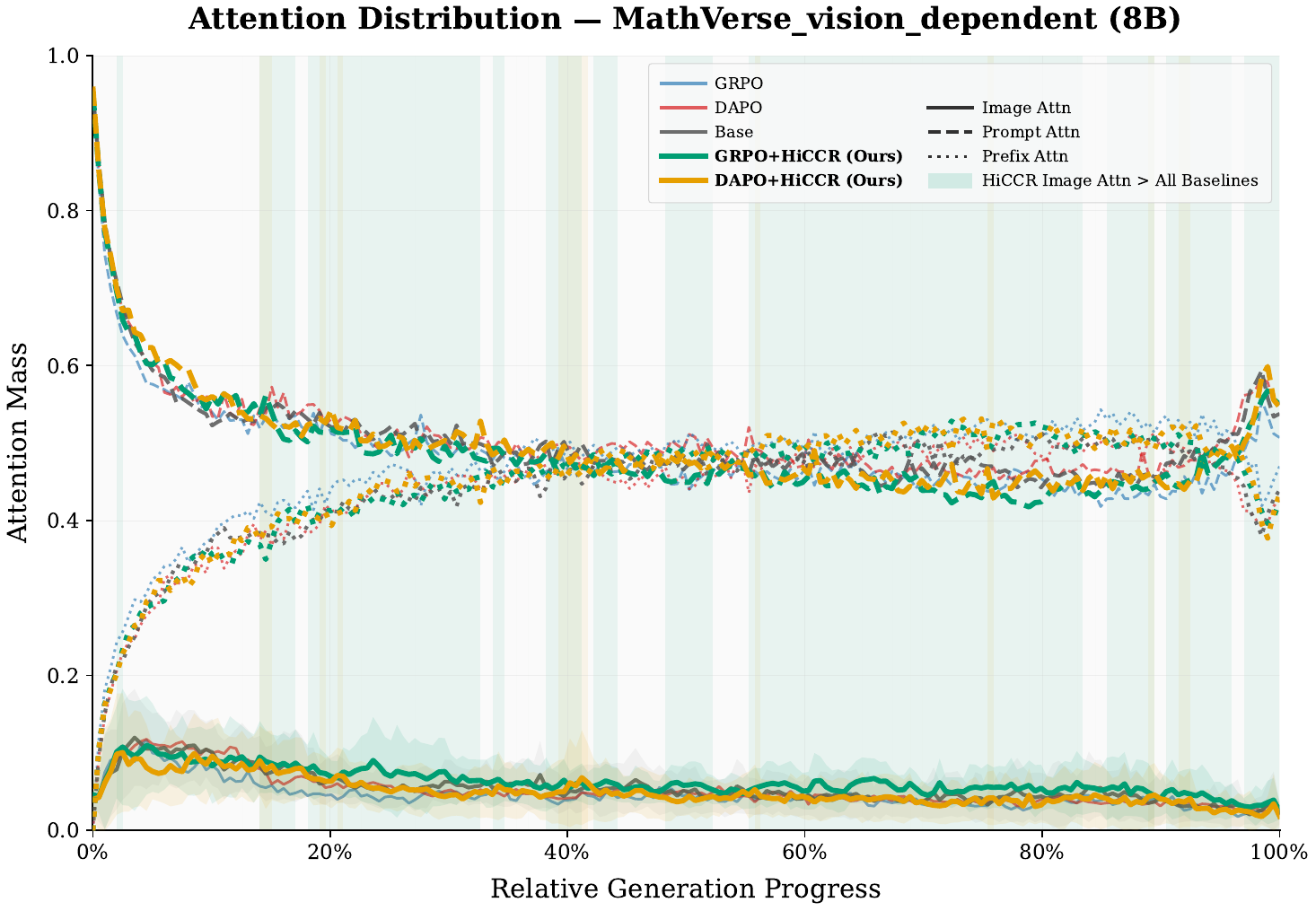}
    \caption{MathVerse-VD}
  \end{subfigure}

  \vspace{0.2cm}
  \begin{subfigure}[b]{0.32\textwidth}
    \includegraphics[width=\textwidth]{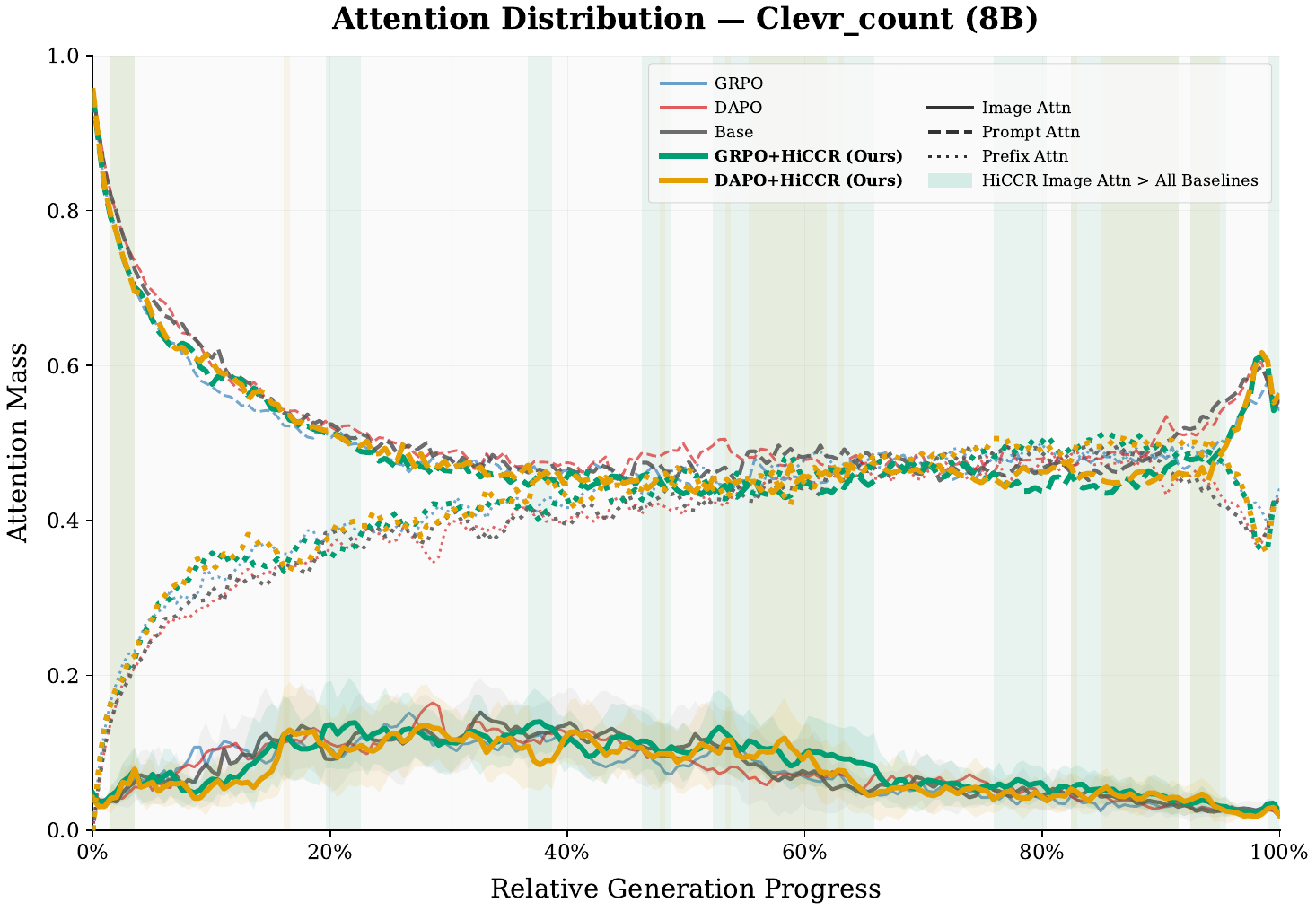}
    \caption{CLEVR-Count}
  \end{subfigure}\hfill
  \begin{subfigure}[b]{0.32\textwidth}
    \includegraphics[width=\textwidth]{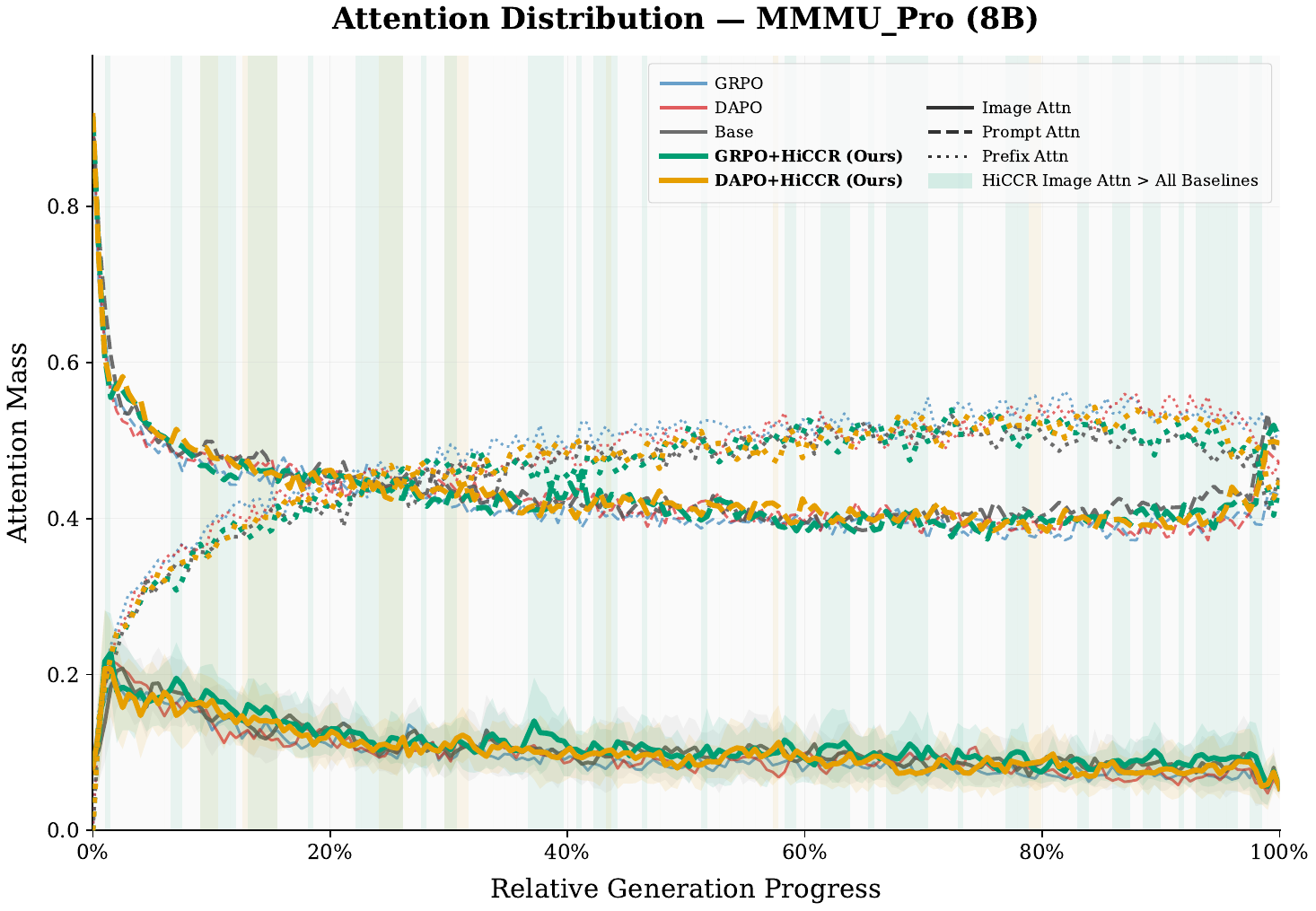}
    \caption{MMMU-Pro}
  \end{subfigure}\hfill
  \begin{subfigure}[b]{0.32\textwidth}
    \includegraphics[width=\textwidth]{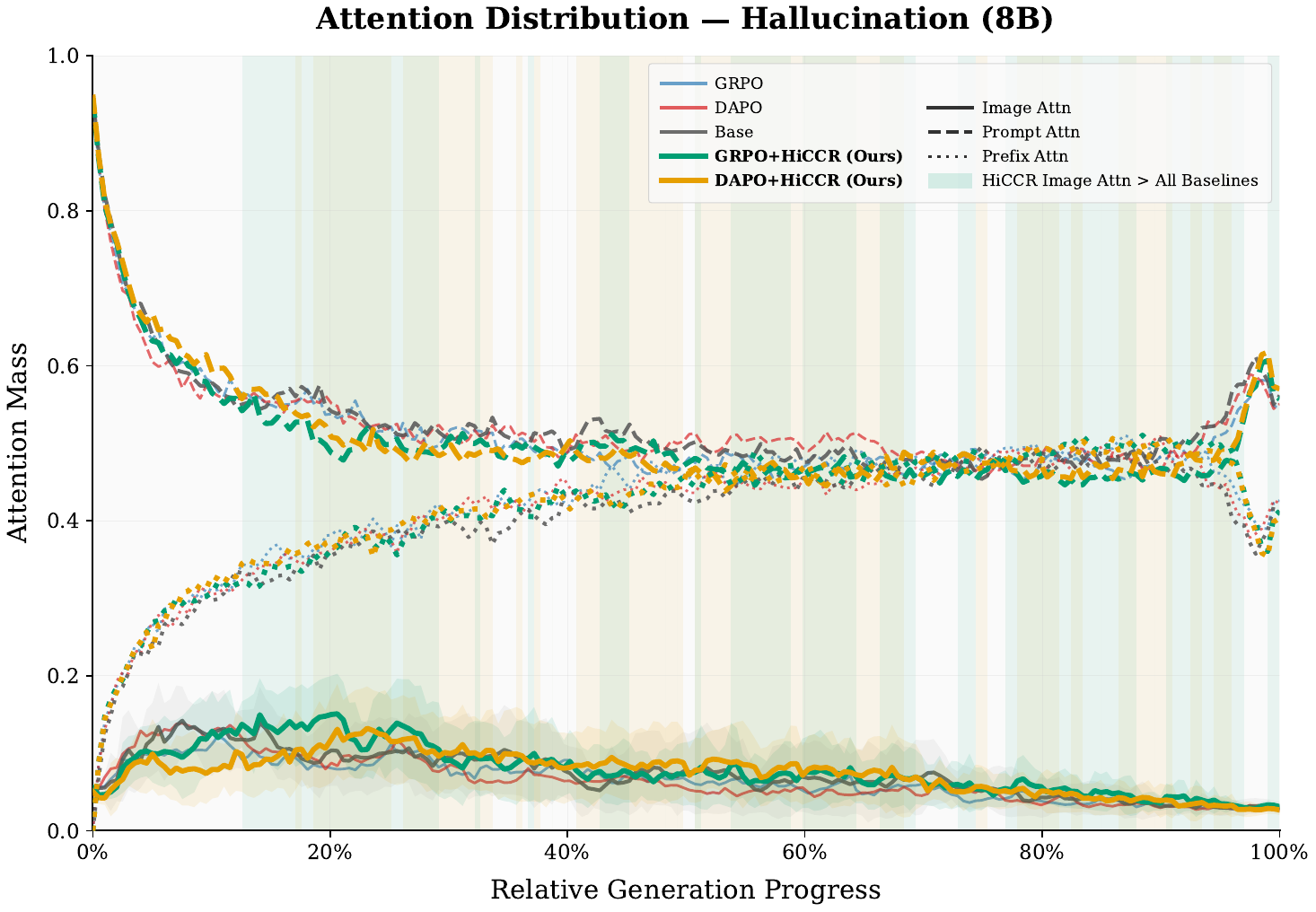}
    \caption{HallusionBench}
  \end{subfigure}
  \caption{\textbf{8B base-model attention comparison.}
  The pretrained base model already shows visual-attention decay, GRPO and DAPO amplify it, and HiCCR reverses the trend.}
  \label{fig:attn_base_8b_all}
\end{figure}

To isolate the effect of HiCCR from the pretrained model's intrinsic attention pattern, we compare HiCCR with the original Qwen3-VL-8B-Instruct model, GRPO, and DAPO.
Figure~\ref{fig:attn_base_8b_all} shows that the base model already reduces image attention as generation progresses.
GRPO and DAPO further intensify this tendency, confirming that standard RLVR amplifies the attention drift inherited from pretraining.
HiCCR substantially reshapes the trajectory by raising image attention above both the base model and RLVR baselines.
Since prompt attention remains comparable, the improvement reflects targeted visual grounding rather than a general redistribution away from text.

\subsubsection{Base Model Comparisons at 4B}
\label{app:attn_base_4b}

\begin{figure}[htbp]
  \centering
  \begin{subfigure}[b]{0.32\textwidth}
    \includegraphics[width=\textwidth]{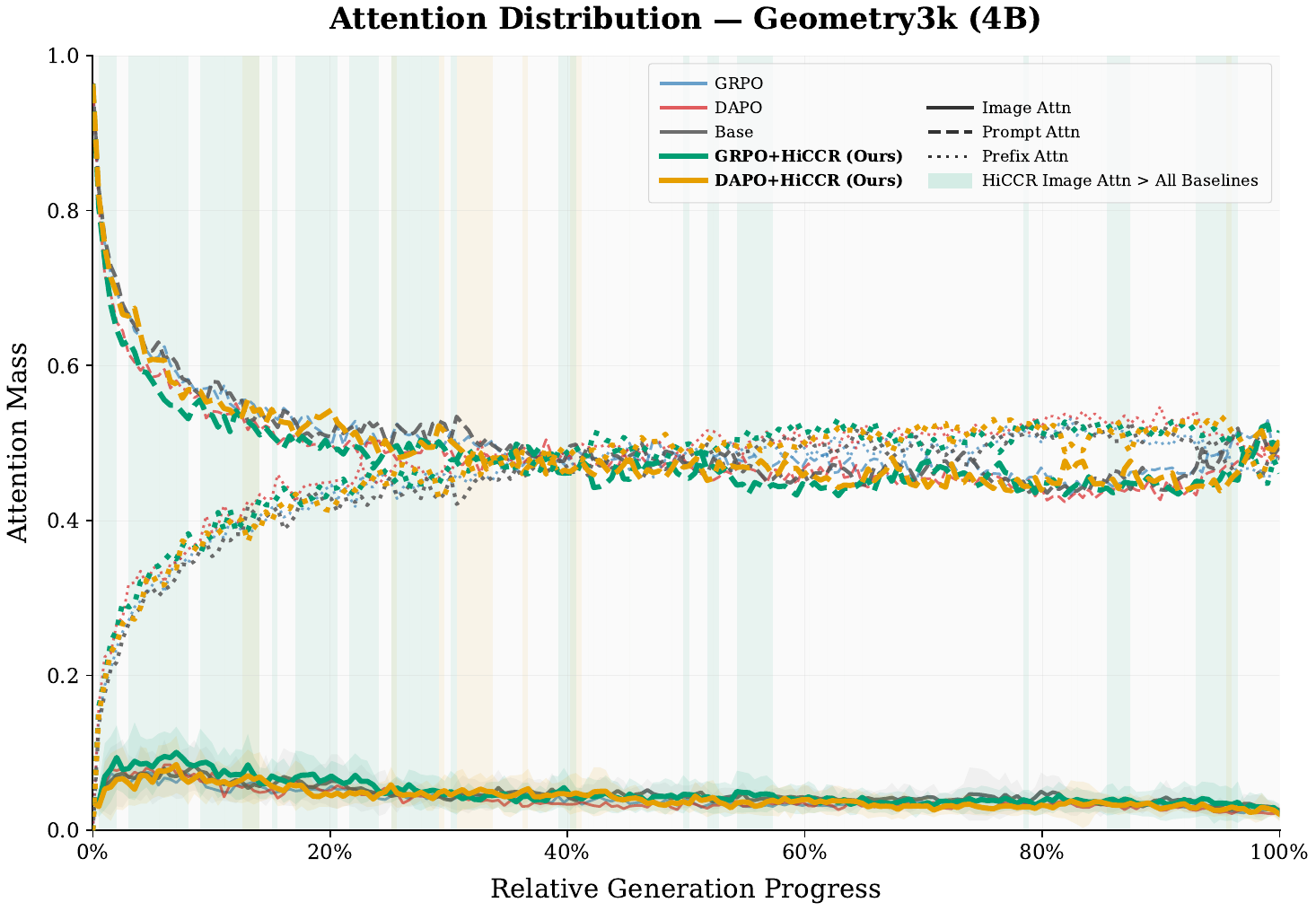}
    \caption{Geometry3K}
  \end{subfigure}\hfill
  \begin{subfigure}[b]{0.32\textwidth}
    \includegraphics[width=\textwidth]{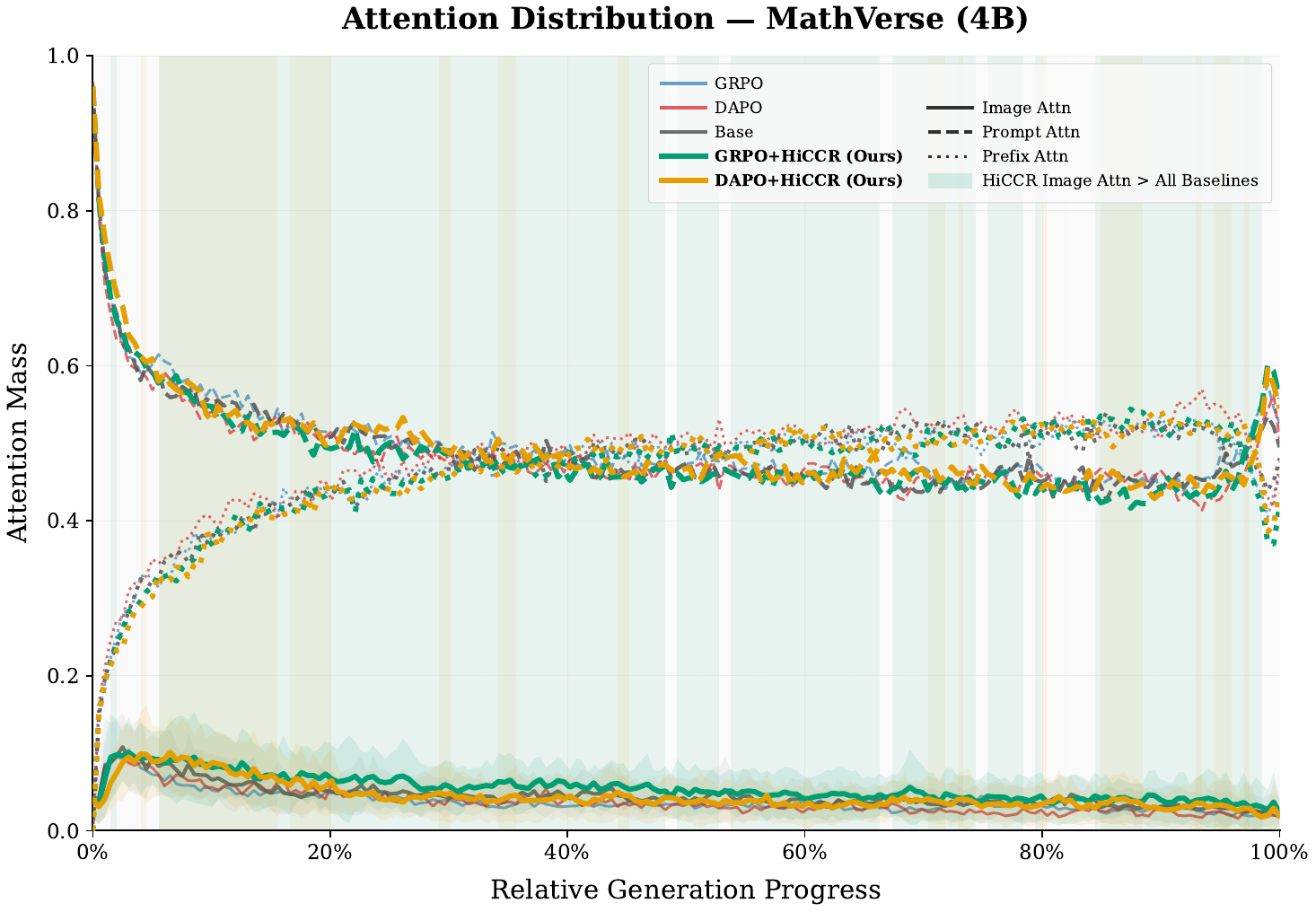}
    \caption{MathVerse}
  \end{subfigure}\hfill
  \begin{subfigure}[b]{0.32\textwidth}
    \includegraphics[width=\textwidth]{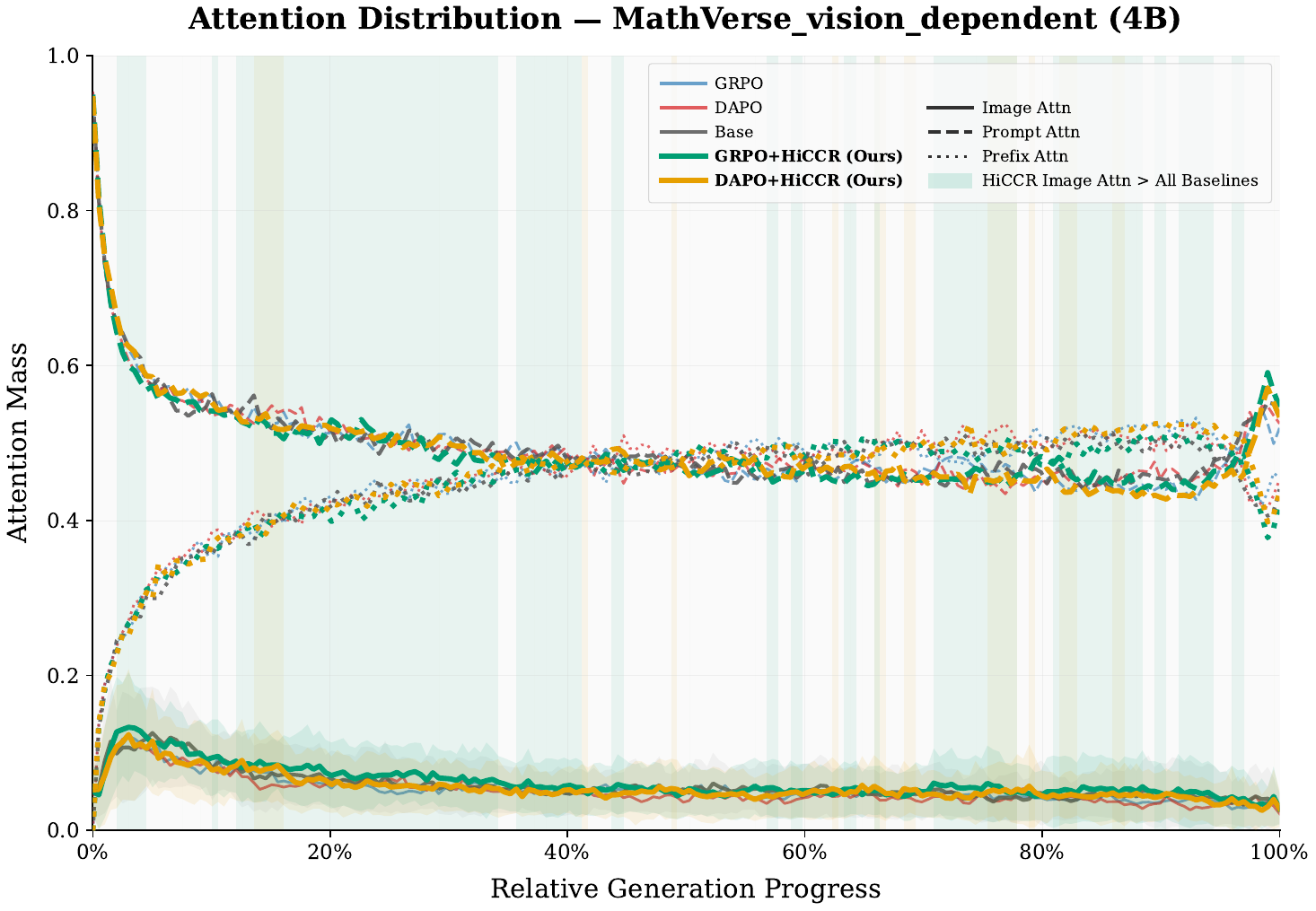}
    \caption{MathVerse-VD}
  \end{subfigure}

  \vspace{0.2cm}
  \begin{subfigure}[b]{0.32\textwidth}
    \includegraphics[width=\textwidth]{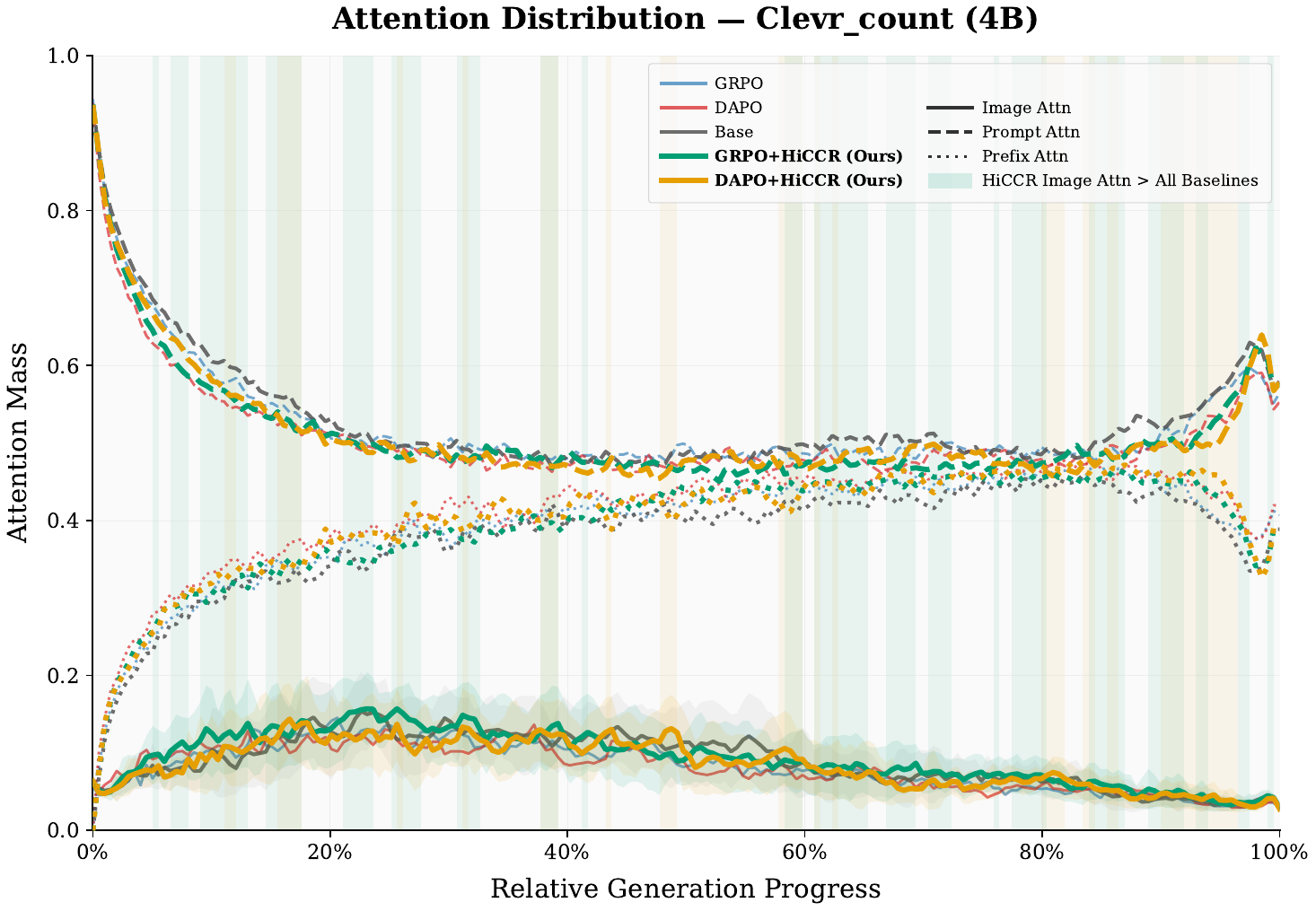}
    \caption{CLEVR-Count}
  \end{subfigure}\hfill
  \begin{subfigure}[b]{0.32\textwidth}
    \includegraphics[width=\textwidth]{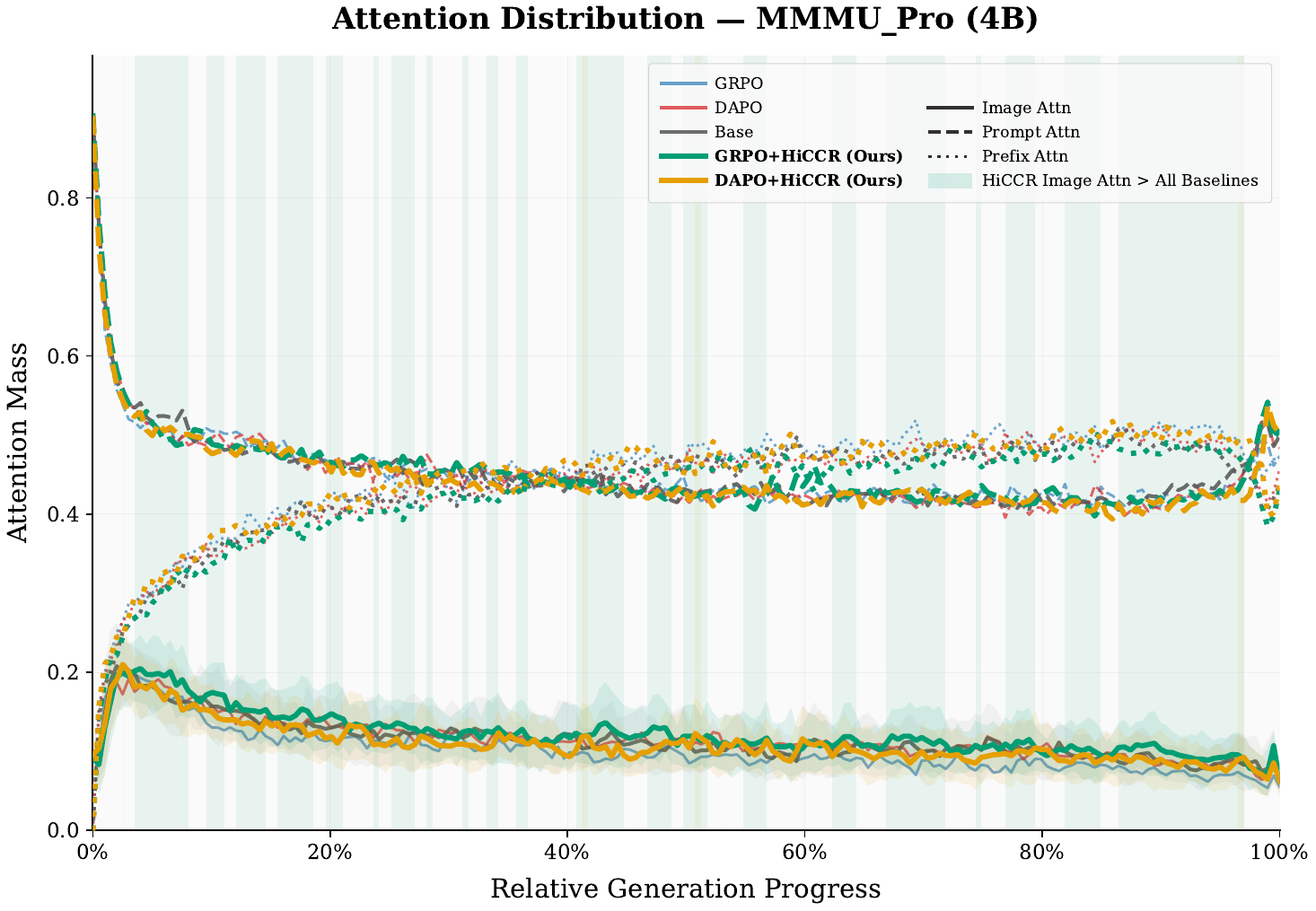}
    \caption{MMMU-Pro}
  \end{subfigure}\hfill
  \begin{subfigure}[b]{0.32\textwidth}
    \includegraphics[width=\textwidth]{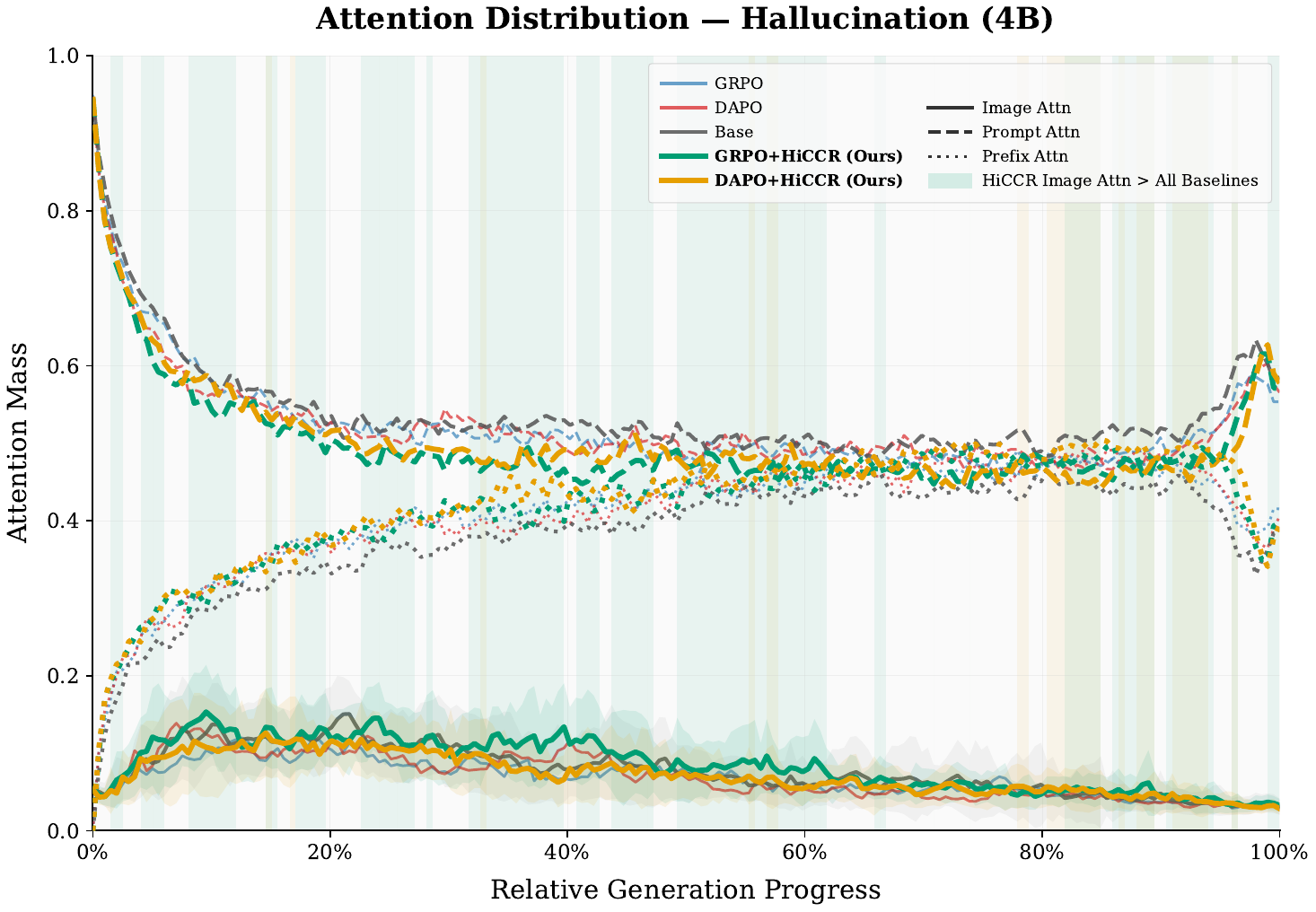}
    \caption{HallusionBench}
  \end{subfigure}
  \caption{\textbf{4B base-model attention comparison.}
  The 4B base model suffers stronger visual-attention collapse, which GRPO and DAPO further reinforce while HiCCR effectively restores sustained image attention.}
  \label{fig:attn_base_4b_all}
\end{figure}

We further compare against the Qwen3-VL-4B-Instruct base model.
As shown in Figure~\ref{fig:attn_base_4b_all}, the smaller base model exhibits a sharper decline in image attention than the 8B base model, often approaching very low visual attention in later decoding steps.
GRPO and DAPO do not correct this collapse and instead reinforce the same drift pattern.
HiCCR effectively rescues visual grounding at this scale, keeping image attention consistently higher throughout generation while maintaining balanced prompt and prefix attention.
This result shows that visual credit reassignment is valuable when standard RLVR is more likely to reinforce language-prior shortcuts.

\subsection{Summary of Findings}
\label{app:empirical_summary}
Taken together, these results provide converging evidence, from correlational to causal, that HiCCR's hierarchical credit reassignment induces functional visual grounding rather than superficial attention patterns.
The gradient attribution confirms where the learning signal flows (\S\ref{app:gradient_attribution}); the counterfactual ablation confirms that visual input is functionally necessary for the learned policy (\S\ref{app:counterfactual}); and the attention dynamics reveal the temporal mechanism through which HiCCR counteracts the drift predicted by Theorem~\ref{thm:drift} (\S\ref{app:attention_dynamics}).



\section{Related Work}
\label{sec:related_work}

\subsection{Multimodal Large Language Models}

Multimodal large language models (MLLMs) extend the reasoning and instruction-following capabilities of LLMs to visual inputs, typically by aligning a vision encoder with a language backbone through projection layers or cross-attention modules~\citep{liu2023visualinstructiontuning,wang2024qwen2vlenhancingvisionlanguagemodels,internvl25,qwen3vl, ivq}.
Recent work has scaled MLLMs along both the model and data axes.
LLaVA~\citep{liu2023visualinstructiontuning} demonstrated that a simple linear projection suffices for effective visual instruction tuning, while subsequent models such as InternVL-2.5~\citep{internvl25} and Qwen2-VL~\citep{wang2024qwen2vlenhancingvisionlanguagemodels} advanced state-of-the-art performance through improved training recipes, dynamic resolution strategies, and curated multimodal corpora.
On the reasoning front, efforts like Insight-V~\citep{insightV} and Vision-R1~\citep{visionr1} explore scalable approaches for generating long-chain reasoning data and applying reinforcement learning to multimodal settings, respectively.

\subsection{Reinforcement Learning with Verifiable Rewards}

Reinforcement learning from human feedback (RLHF)~\citep{ouyang2022traininglanguagemodelsfollow,stiennon2022learningsummarizehumanfeedback} has become the dominant paradigm for aligning LLMs with human preferences, yet it relies on a learned reward model that is expensive to train and prone to reward hacking.
Reinforcement learning with verifiable rewards (RLVR) offers a compelling alternative for tasks where correctness can be automatically verified---such as mathematics, coding, and visual question answering---by replacing the reward model with a deterministic verifier that provides binary (correct/incorrect) feedback~\citep{deepseekr1,lambert2026reinforcementlearninghumanfeedback}.
DeepSeek-R1~\citep{deepseekr1} demonstrated that RLVR alone can elicit sophisticated chain-of-thought reasoning in large-scale models.
On the algorithmic side, GRPO~\citep{grpo} eliminates the critic network by estimating baselines from group-sampled trajectories, substantially reducing memory overhead.
DAPO~\citep{dapo} further improves training stability through clip-higher, dynamic sampling, and token-level loss normalization.
Recent studies have begun probing the boundaries and failure modes of RLVR: \citep{yue2025doesreinforcementlearningreally} show that RLVR gains largely stem from amplifying capabilities already present after supervised fine-tuning, and that performance saturates without sufficiently diverse training distributions~\citep{allroadsleadromeincentivizing_mupo}.
In the multimodal domain, R1-V~\citep{chen2025r1v} and Vision-R1~\citep{visionr1} apply RLVR to MLLMs but retain the uniform per-token advantage formulation, leaving the credit assignment problem unaddressed.
Despite these advances, how to effectively reinforce \emph{visual grounding} during post-training remains understudied~\citep{visualrft}.
Most RLVR pipelines treat visual and textual tokens uniformly~\citep{avar_narrow, Wang2025BeyondNA}, overlooking the unique role of visual context in grounding the reasoning process---a gap that motivates our work.

\subsection{Token-Level Credit Assignment in RLVR}
The credit assignment problem---determining which tokens in a generated sequence are responsible for the final reward---is central to efficient policy optimization.
Standard RLVR algorithms such as GRPO~\citep{grpo} dispense with the critic model and broadcast the trajectory-level advantage uniformly to every token, implicitly assuming equal contribution.
This assumption is particularly problematic for vision-centric understanding and long-form reasoning, where pivotal tokens are grounded on visual content or only a sparse subset actually drives the outcome~\citep{meng2026sparsecriticaltokenlevelanalysis, Gao2026LiveAL}.
A growing body of work seeks finer-grained credit signals; we organize these efforts by modality setting.

\noindent\textbf{Credit assignment in text-only RLVR.}
One line of work introduces critic-free temporal-difference corrections.
GRPO-$\lambda$~\citep{grpolambdacredit} propagates credit backward via an eligibility trace reparameterised from the GAE $\lambda$-return, while TEMPO~\citep{tempo} derives nonparametric token-level TD corrections from the implicit prefix tree formed by grouped rollouts.
A second line leverages entropy for reward shaping: GTPO~\citep{gtpo} redistributes trajectory reward proportionally to per-token policy entropy, treating it as a proxy for cognitive effort, and Entropy-RL~\citep{entropyrl} provides a theoretical account linking entropy dynamics to the probability-advantage covariance.
A third category employs external models for dense credit—Grad2Reward~\citep{zhang2026grad2rewardsparsejudgmentdense} extracts token-level rewards from a judge model's gradient attribution, and CAPO~\citep{capo-llm-as-gen} uses an LLM-as-GenPRM to generate step-wise critiques with asymmetric outcome–process reward shaping.
A fourth line learns explicit value functions: Q-RM~\citep{qrm} optimizes a discriminative Q-function over preference data for token-level value estimates, and PURE~\citep{pure} replaces the canonical summation-form credit with a min-form value function $V(s_t)=\min_{k\ge t}r_k$ to alleviate PRM-induced reward hacking.
All the above are designed for text-only LLMs and do not account for the interplay between generated tokens and \emph{visual context}.

\noindent\textbf{Credit assignment in multimodal RLVR.}
Several recent efforts extend token-level credit to the multimodal setting.
PEPO~\citep{pepo}, AT-RL~\citep{atrl}, and VGPO~\citep{vgpo_acl} derive
visual grounding signals from hidden-state similarity between response and
image tokens. These approaches operate in the \emph{representation} space
and rely on the assumption that high cosine similarity implies visual
dependence---an assumption that is not formally justified.
In contrast, HiCCR measures the \emph{concentration} of the attention
distribution over visual positions, which directly reflects how selectively
the model extracts visual information during generation.
A second family identifies perception-critical tokens via counterfactual perturbation: VPPO~\citep{vppo_spotlight}, ToR-GRPO~\citep{tor-grpo}, and PGPO~\citep{notalltokens} compare next-token distributions with and without visual input to score per-token visual dependency, while Video-KTR~\citep{videoktr} extends this to the video domain with masked and frame-shuffled branches.
All such counterfactual methods require one or more additional forward passes, incurring non-trivial overhead.
TW-GRPO~\citep{twgrpo} avoids perturbation by measuring intra-group distributional divergence at each token position as a proxy for information density.
APPO~\citep{appo} is also based on intra-group similarities, by calculating attention-to-frame scores to target perception tokens and reinforce them with more credits.

Despite recent progress, existing multimodal credit-assignment methods largely leave the training-time evolution of visual grounding under RLVR underexplored. In particular, they do not diagnose the grounding drift we identify: standard uniform-credit RLVR can reduce image attention and the gradient share routed to visual tokens even while improving task accuracy, making it unclear whether gains come from stronger visual use or greater exploitation of textual/reasoning priors. HiCCR is designed around this diagnosis: it estimates per-token visual engagement through the information concentration of visual attention, derives both token- and trajectory-level credit weights from this lightweight forward-pass score, and pairs the mechanism with attention, gradient, and visual counterfactual diagnostics to verify that accuracy gains are accompanied by mitigated grounding drift.

\section{Additions about Training and Evaluation}
\label{sec:details_train_eval}

\subsection{Experiments Details}
\label{app:exp_details}

\paragraph{Implementation framework.}
All experiments are built on veRL~\citep{verl} with PyTorch 2.8.0~\citep{pytorch} and CUDA 12.8, and are conducted on $8\times$ High-Performance GPUs. We fine-tune the Qwen3-VL-4B and 8B backbones~\citep{qwen3vl} following the RLVR paradigm of DeepSeek-R1~\citep{deepseekr1}.

\paragraph{Training details \& computational efficiency.}
We integrate HiCCR with two representative RLVR algorithms---GRPO~\citep{grpo} and DAPO~\citep{dapo}---yielding \textbf{HiCCR-G} and \textbf{HiCCR-D}, respectively. Key training hyperparameters are shared across all runs: $G{=}5$ rollouts per prompt, temperature $1.0$, maximum generation length $4096$, learning rate $2\times10^{-6}$ with AdamW~\citep{adamw} and cosine decay, and PPO clip ratio $\epsilon{=}0.2$. For HiCCR-specific settings, we fix $\lambda_{\text{tok}}{=}0.5$, $\tau{=}0.5$, and select the last two transformer layers as $\mathcal{L}_{\text{target}}$ throughout. Ablation on the target layer has been presented in Table~\ref{tab:layer_ablation_grpo_4b}.

\paragraph{Implementation.}
Eq.~\eqref{eq:concentration} does \emph{not} require materializing the full $T \times T$ attention matrix.
We register lightweight forward hooks on the query and key projection modules of the target layers, cache their detached outputs, and compute the score via a two-pass chunked algorithm: (i)~accumulate the global log-sum-exp over visual positions using \texttt{logaddexp}, and (ii)~compute $\sum_{j} p \log p$ chunk-by-chunk with the cached normalizer.
For Grouped Query Attention, keys are expanded to match the query head count.
The resulting $\mathcal{O}(S)$ token scores are immediately offloaded to CPU, so GPU memory overhead is near zero.
The entire procedure runs in a single \texttt{no\_grad} forward pass before the policy update, retaining only the $[B, T]$-shaped score tensor, and adds less than 3\% wall-clock time per iteration (see Table~\ref{tab:compute_overhead} for detailed profiling).
Since the computation operates solely on intercepted Q/K projections, it is fully compatible with FlashAttention and other fused attention backends.
Full hyperparameter configurations are provided in Appendix~\ref{app:hyperparams}.

\paragraph{Compared Multimodal Models.}
To contextualize HiCCR beyond controlled RLVR baselines, we compare against a broad set of proprietary, general-purpose, and domain-specific multimodal models. Proprietary baselines include GPT-4o, GPT-4o-mini, and Gemini-2.0-Flash~\citep{openai2024gpt4ocard}. General-purpose open models include Qwen2.5-VL-72B~\citep{qwen25vl}, InternVL2.5-8B/78B~\citep{internvl25}, LLaVA-OV-7B/72B and LLaVA-OV-1.5-8B~\citep{llavaonev,llavaonev15}, LLaVA-Critic-R1-7B~\citep{wang2025llavacriticr1criticmodelsecretly}, and R1-OneVision-7B~\citep{r1onevision}. We further include domain-specific multimodal reasoning models, including MM-Eureka-7B and MM-Eureka-7B-CPGD~\citep{mmeurka}, R1-VL-7B~\citep{chen2025r1v}, VLAA-Thinker-7B~\citep{vlaa-thinker}, VL-Rethinker-7B~\citep{vl-rethinker}, RACRO-7B-CRO-GRPO~\citep{racro}, ThinkLite-7B-VL~\citep{thinklite}, and AVAR-Thinker-7B~\citep{avar_narrow}. This comparison evaluates whether HiCCR-trained Qwen3-VL models~\citep{qwen3vl} remain competitive beyond matched RLVR baselines under the same benchmark suite.

\paragraph{Prompt Template.}
Evaluation shares the chat template below, applied identically to every baseline and HiCCR variant, so that differences are attributable to credit reassignment rather than prompting.

\begin{tcolorbox}[
    colback=gray!4, colframe=black!55, boxrule=0.4pt,
    left=5pt, right=5pt, top=3pt, bottom=3pt, arc=1pt,
    fonttitle=\bfseries\small, title=Rollout Prompt
]
\small
\textbf{System.} You first think about the reasoning process internally, then provide the final answer. Enclose them in \texttt{<think>...</think>} and \texttt{<answer>...</answer>} respectively.\\[2pt]
\textbf{User.} \texttt{<image>}\ \{question\}\\[2pt]
\textbf{Assistant.} \texttt{<think>}
\end{tcolorbox}

\subsection{Hyperparams}
\label{app:hyperparams}

Table~\ref{tab:hyperparams} lists all hyperparameters used in our experiments.
All runs share the same training, rollout and optimizer settings unless
otherwise stated; only HiCCR-specific entries are introduced in the dedicated
block. Values confirmed in the main text are kept identical here, and any
unspecified defaults follow the corresponding open-source implementation of
veRL~\citep{verl}, GRPO~\citep{grpo} and
DAPO~\citep{dapo}. Values marked with $\dagger$ follow the default
setting of the corresponding base algorithm and are kept untouched for fair
comparison.

\definecolor{groupblue}{RGB}{70,130,200}   
\begin{table*}[t]
\centering
\footnotesize
\setlength{\tabcolsep}{3pt}
\setlength{\aboverulesep}{0pt}
\setlength{\belowrulesep}{0pt}
\renewcommand{\arraystretch}{0.9}
\caption{Hyperparameter configuration for HiCCR and the baseline RLVR
algorithms. Shared values apply to both HiCCR-G and HiCCR-D.}
\label{tab:hyperparams}
\vspace{-4pt}
\begin{tabular}{@{}p{0.18\textwidth}p{0.26\textwidth}@{\hspace{10pt}}p{0.20\textwidth}p{0.26\textwidth}@{}}
\toprule
\textbf{Hyperparameter} & \textbf{Value} &
\textbf{Hyperparameter} & \textbf{Value} \\
\midrule
\multicolumn{2}{@{}l}{\textcolor{groupblue}{\textit{Model \& Data}}} &
\multicolumn{2}{@{}l}{\textcolor{groupblue}{\textit{RL Objective (shared)}}} \\
Backbone                     & Qwen3-VL, InternVL3           &
PPO clip ratio $\varepsilon$ & $0.2$ \\
Training dataset             & ViRL39K~\citep{vl-rethinker} &
KL coefficient $\beta_{\text{KL}}$ & $0.0$$^{\dagger}$ \\
Max prompt length            & $8{,}192$ tokens           &
Advantage norm.              & group-relative (Eq.~1) \\
Max response length          & $4{,}096$ tokens           &
Reward type                  & $\{0,1\}$ verifiable \\
Image resolution             & up to $1024^2$             &
                             & \\
\addlinespace[1pt]
\multicolumn{2}{@{}l}{\textcolor{groupblue}{\textit{Optimization}}} &
\multicolumn{2}{@{}l}{\textcolor{groupblue}{\textit{DAPO-specific (HiCCR-D)}}} \\
Optimizer                    & AdamW ($\beta_1{=}.9,\beta_2{=}.999$) &
Clip-higher $\varepsilon_{\text{hi}}$ & $0.28$$^{\dagger}$ \\
Learning rate                & $2{\times}10^{-6}$         &
Clip-lower $\varepsilon_{\text{lo}}$  & $0.20$$^{\dagger}$ \\
LR schedule                  & cosine, $10$ warm-up       &
Dynamic sampling             & enabled$^{\dagger}$ \\
Weight decay                 & $1{\times}10^{-2}$         &
Overlong soft penalty        & enabled$^{\dagger}$ \\
Gradient clipping            & $1.0$                      &
                             & \\
Training steps               & $198$                      &
\multicolumn{2}{@{}l}{\textcolor{groupblue}{\textit{HiCCR-specific}}} \\
Global batch (prompts)       & $256$                      &
Token weight $\lambda_{\text{tok}}$ & $0.5$ \\
Mini-batch (per update)      & $128$                       &
Softmax temperature   & $0.5$ \\
PPO epochs ($\mu$)           & $2$                        &
Target layers $\mathcal{L}_{\text{target}}$ & last $2$ decoder layers \\
Precision                    & bf16               &
Attention heads              & all (mean) \\
                             &                            &
Stability $\epsilon$         & $1{\times}10^{-8}$ \\
                             &                            &
Reassignment cond.           & $\hat{A}^{(i)}{>}0$ only \\
\addlinespace[1pt]
\multicolumn{2}{@{}l}{\textcolor{groupblue}{\textit{Rollout \& Sampling}}} &
\multicolumn{2}{@{}l}{\textcolor{groupblue}{\textit{Hardware \& Runtime}}} \\
Rollouts per prompt ($G$)    & $5$                        &
Framework                    & veRL~\citep{verl} \\
Sampling temperature         & $1.0$                      &
Backend                      & PyTorch 2.8, CUDA 12.8 \\
Top-$p$ / Top-$k$            & $1.0$ / $-1$               &
GPUs                         & $8{\times}$ High-Performance GPUs \\
Rollout engine               & vLLM 0.11                  &
Attention kernel             & FlashAttention-2 \\
                             &                            &
Wall-clock overhead          & $<3\%$ \\
\addlinespace[1pt]
\multicolumn{2}{@{}l}{\textcolor{groupblue}{\textit{Evaluation}}} &
                             & \\
Decode temperature         & $1.0$                      &
                             & \\
Samples per query            & $8$ (AVG@8)                 &
                             & \\
Max gen.\ length      & $4{,}096$                   &
                             & \\
Checkpoint choice            & last training ckpt          &
                             & \\
\bottomrule
\end{tabular}
\vspace{-6pt}
\end{table*}

\paragraph{Notes on robustness to hyperparameter choices.}
We verified that HiCCR is insensitive to moderate perturbations of its two
new hyperparameters: Table~\ref{tab:lambda_tok} shows stable performance in
the range $\lambda_{\text{tok}}\in[0.5,0.8]$, and Table~\ref{tab:layer_ablation_grpo_4b}
reports monotonic improvement as $\mathcal{L}_{\text{target}}$ moves toward
the last layers. All remaining entries follow the defaults of the
corresponding base algorithm and were \emph{not} re-tuned for HiCCR, which
isolates the gains to credit reassignment rather than to hyperparameter
search.

\subsection{Algorithm Pipeline}
\label{app:algorithm}

HiCCR integrates into mainstream group-based RLVR algorithm as a pre-processing step that transforms advantages before the standard policy update. The procedure is summarized in Alg.~\ref{alg:hiccr}.

\begin{algorithm}[t]
\caption{HiCCR: Hierarchical Context Credit Reassignment}
\label{alg:hiccr}
\begin{algorithmic}[1]
\REQUIRE Policy $\pi_\theta$, dataset $\mathcal{D}$, target layers $\mathcal{L}_{\text{target}}$, hyperparameter $\lambda_{\text{tok}}$, temperature $\tau$
\FOR{each training iteration}
    \STATE Sample prompts $\{\mathbf{x}^{(i)}\}$ from $\mathcal{D}$; generate $G$ trajectories per prompt
    \STATE Compute verifiable rewards $R^{(i)}$ and group advantages $\hat{A}^{(i)}$ via Eq.~\eqref{eq:group_adv}
    \STATE \textcolor{blue}{\texttt{/* Stage 1: Context Credit Scoring */}}
    \STATE Register forward hooks on $\{\texttt{q\_proj}, \texttt{k\_proj}\}$ in layers $\mathcal{L}_{\text{target}}$
    \FOR{each micro-batch}
        \STATE Forward pass through $\pi_\theta$ (\texttt{eval}, \texttt{no\_grad})
        \STATE Compute $p_{\mathcal{V}}(\cdot \mid t)$ via Eq.~\eqref{eq:intra_vis_attn}; compute $C_{\mathcal{V}}(t)$ via Eq.~\eqref{eq:concentration}
    \ENDFOR
    \STATE Remove all hooks
    \STATE \textcolor{blue}{\texttt{/* Stage 2: Hierarchical Advantage Reassignment */}}
    \STATE Compute token weights $\beta_t^{(i)}$ via Eqs.~\eqref{eq:norm_score}--\eqref{eq:beta}
    \STATE Compute trajectory weights $\alpha^{(i)}$ via Eqs.~\eqref{eq:traj_score}--\eqref{eq:alpha}
    \STATE $\tilde{A}_t^{(i)} \leftarrow \alpha^{(i)} \cdot \beta_t^{(i)} \cdot \hat{A}^{(i)}$ if $\hat{A}^{(i)}{>}0$;\; $\tilde{A}_t^{(i)} \leftarrow \hat{A}^{(i)}$ otherwise
    \STATE \textcolor{blue}{\texttt{/* Stage 3: Standard Policy Update (unchanged) */}}
    \FOR{each PPO epoch}
        \STATE Update $\theta$ using Eq.~\eqref{eq:ppo_loss} with $\tilde{A}_t^{(i)}$ replacing $\hat{A}^{(i)}$
    \ENDFOR
\ENDFOR
\end{algorithmic}
\end{algorithm}

\paragraph{Design principles.}
HiCCR is built on four properties that jointly underpin its robustness. (a) \textit{Information-theoretic credit signal}: the context credit score is derived from the normalized entropy of visual attention, offering a principled, shift-invariant measure of grounding that is stable across layers and training stages. (b) \textit{Minimal overhead}: the score is obtained from the policy's own query-key representations in a single forward pass, with no trainable reward or value network and only $\lambda_{\text{tok}}$ to specify (Appendix~\ref{app:overhead}). (c) \textit{Bounded modulation}: $\alpha \in [1, 2]$ and $\beta \in (1, 1{+}\lambda_{\text{tok}})$ by construction, so HiCCR only \emph{amplifies} gradients for visually engaged tokens without reversing or suppressing any signal, preserving the convergence guarantees of the base algorithm (\S\ref{sec:theory:correction}, \S\ref{sec:exp_ablation}). (d) \textit{Asymmetric application}: reassignment is applied to positive-advantage trajectories only, since negative advantages already serve as corrective signals and modulating them would risk penalizing exploratory rollouts that attend to the image yet reach an incorrect conclusion through otherwise valid reasoning (\S\ref{sec:exp_ablation}).

\section{Case Study}
\label{sec:case_study}

To complement the quantitative results in Section~\ref{sec:experiments}
and the aggregate credit analysis in Appendix~\ref{sec:context_credits_analysis},
we visualize four rollouts sampled from the training set
(ViRL39K~\citep{vl-rethinker}) produced by HiCCR-G on Qwen3-VL-8B, to
illustrate how the context credit weight $\beta_t$ is allocated at the
token level. Each figure pairs the input image with the generated
rationale, where every token is color-coded by its final credit weight
$\beta_t$ (blue: low visual dependency; red: high visual dependency).
All five rollouts yield correct predictions.

\begin{figure}[htbp]
\centering
\includegraphics[width=\linewidth]{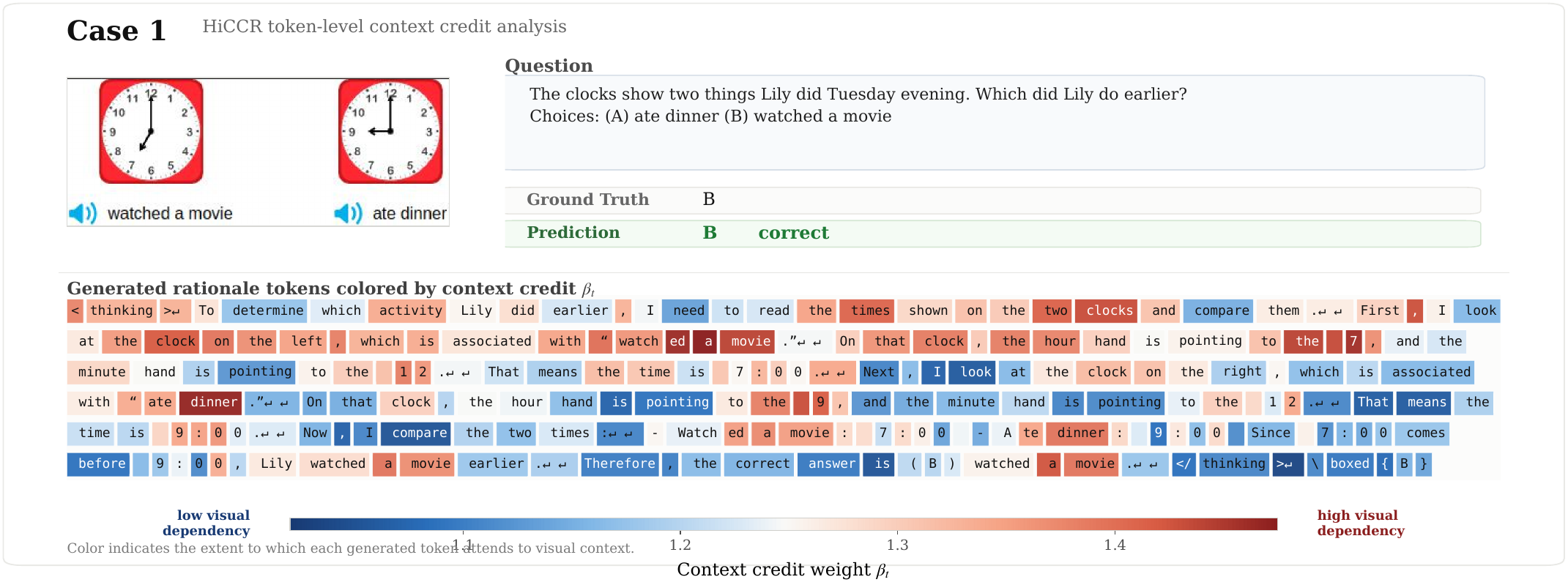}
\caption{\textbf{Case 1: clock reading (ViRL39K training rollout).}
Tokens referring to clock positions (``\textit{7}'', ``\textit{9}'',
``\textit{pointing}'', ``\textit{clock}'', ``\textit{movie/dinner}'')
receive the highest credit, while temporal comparison tokens
(``\textit{before}'', ``\textit{Since}'', ``\textit{Therefore}'') drop to
near-baseline weight, showing that HiCCR amplifies visually grounded
time reading rather than purely textual reasoning.}
\label{fig:case1}
\end{figure}

\begin{figure}[htbp]
\centering
\includegraphics[width=\linewidth]{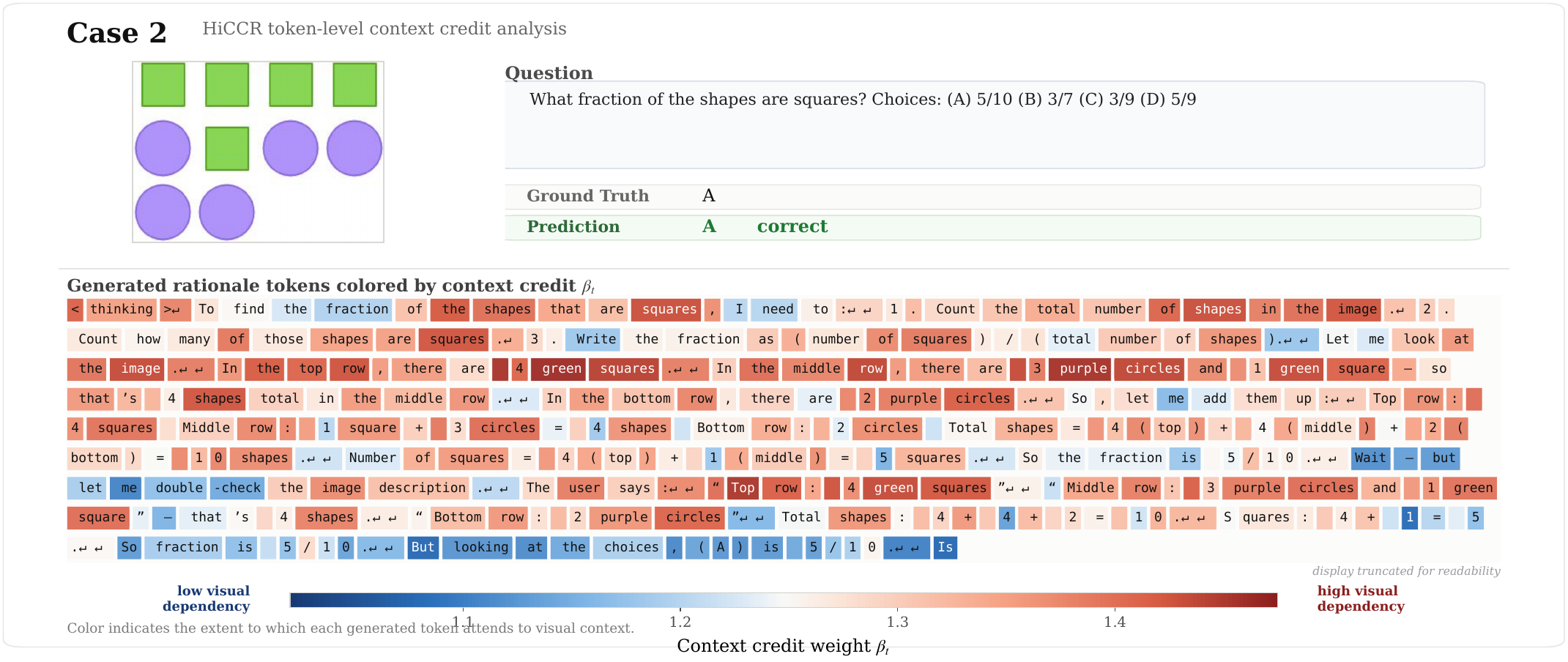}
\caption{\textbf{Case 2: fraction of shapes (ViRL39K training rollout).}
Shape- and color-descriptive tokens (``\textit{green squares}'',
``\textit{purple circles}'', row-wise counts) carry the largest weights,
while arithmetic tokens (``$5/10$'', ``\textit{Is}'') fall into the
low-dependency band, indicating that credit follows perceptual evidence
rather than downstream calculation.}
\label{fig:case2}
\end{figure}

\begin{figure}[htbp]
\centering
\includegraphics[width=\linewidth]{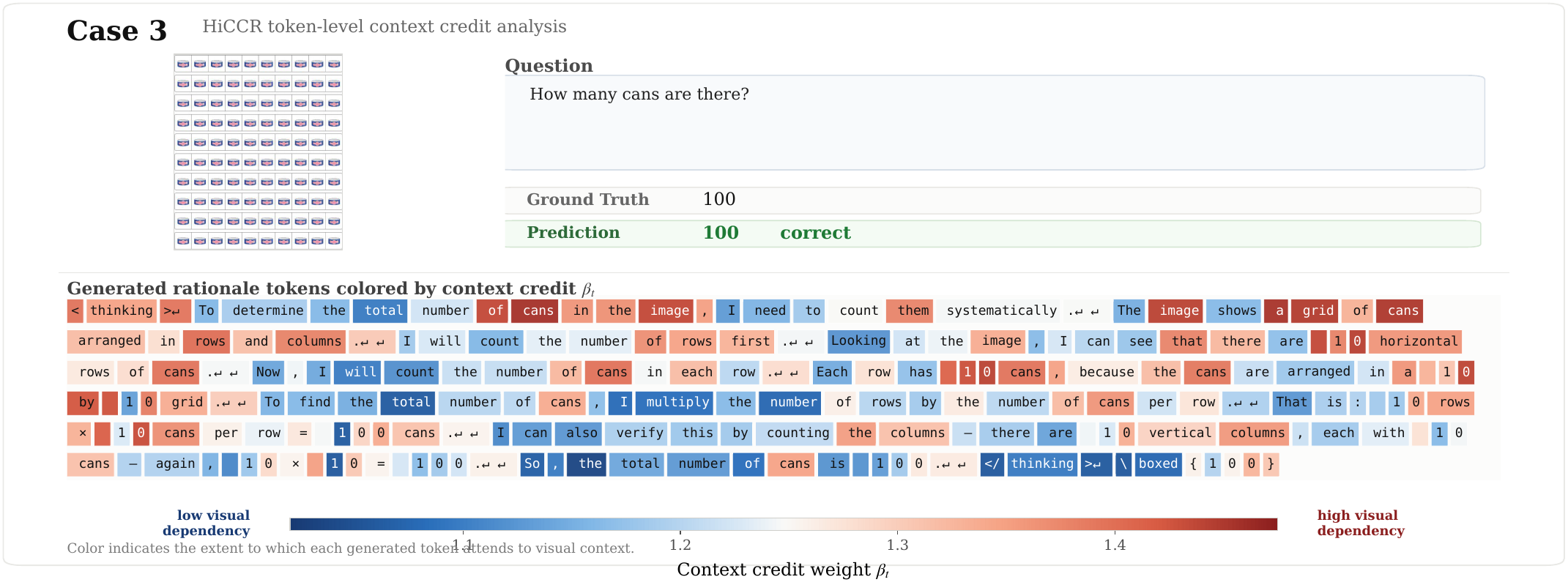}
\caption{\textbf{Case 3: counting $10{\times}10$ cans
(ViRL39K training rollout).}
HiCCR assigns high credit to structural-perception tokens
(``\textit{grid}'', ``\textit{rows}'', ``\textit{columns}'',
``\textit{10 horizontal rows}''), while the multiplicative closure
(``$10\times10=100$'') and the answer tag receive near-unity weight,
a clean separation between ``looking'' and ``computing''.}
\label{fig:case3}
\end{figure}

\begin{figure}[htbp]
\centering
\includegraphics[width=\linewidth]{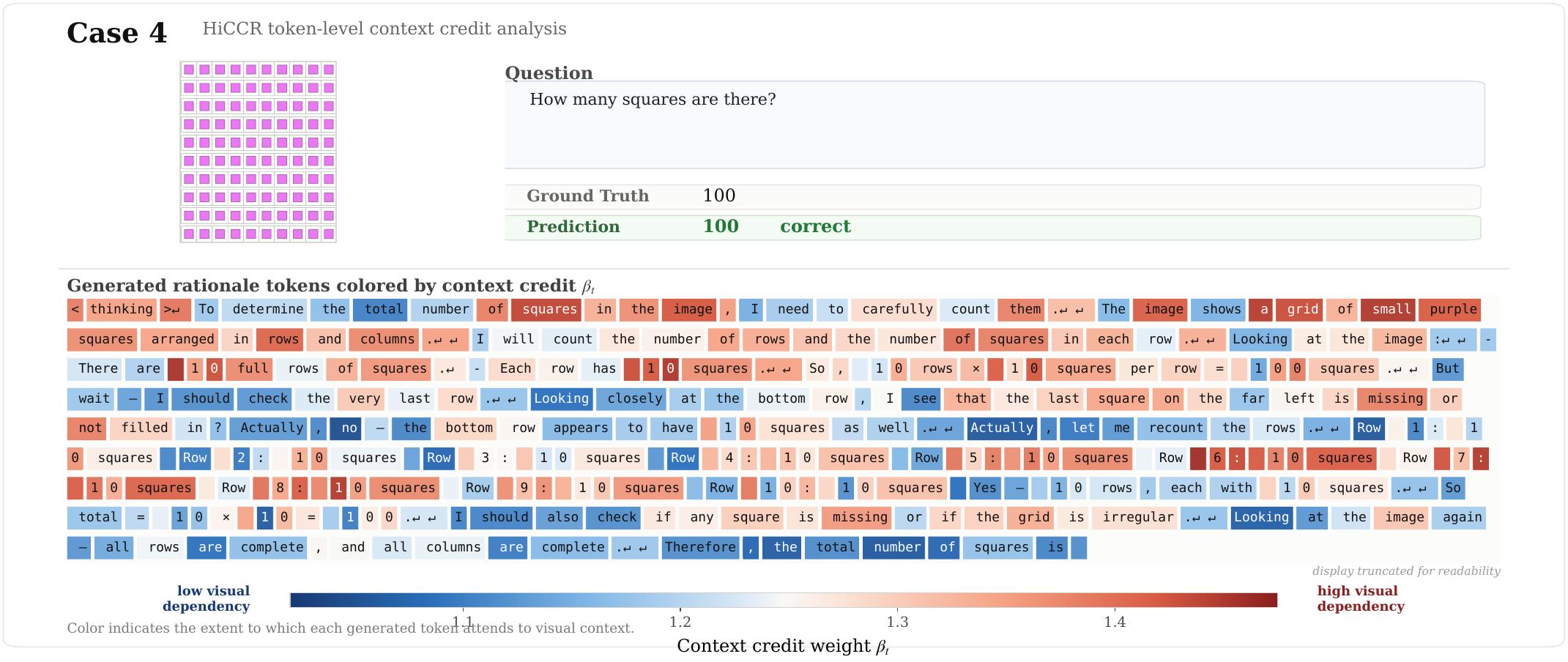}
\caption{\textbf{Case 4: counting $10{\times}10$ squares with
self-verification (ViRL39K training rollout).}
Even in a rollout containing a self-check sub-trajectory (``\textit{But
wait $-$ I should check the very last row}''), HiCCR consistently
upweights the row-wise enumeration tokens (Row~1--Row~10,
``\textit{squares}''), showing that the credit signal is robust to
reasoning detours.}
\label{fig:case4}
\end{figure}

Three consistent patterns emerge across cases.
\textbf{(i)} High-credit tokens concentrate on \emph{visually grounded
semantic units}---clock hands and numerical readings in
Figure~\ref{fig:case1}, color-shape descriptors in
Figure~\ref{fig:case2}, object counts and grid structure in
Figures~\ref{fig:case3}--\ref{fig:case4}---exactly the tokens whose generation
requires reading the image. \textbf{(ii)} Low-credit tokens cluster on
\emph{language-driven connectives and meta-reasoning}
(``\textit{Therefore}'', ``\textit{That means}'',
``\textit{$\langle$/thinking$\rangle$}'', arithmetic closers such as
``$=10$''), where the next token is determined by textual priors rather
than visual evidence. \textbf{(iii)} Credit decays along the response:
early perceptual tokens receive the strongest amplification, while later
symbolic-manipulation tokens receive weights close to unity, matching the
quantitative trend in Figure~\ref{fig:credit_analysis}. Together, these
patterns confirm that HiCCR reshapes gradient flow in a vision-aware,
rather than uniform, manner---without any explicit token-level supervision.

\section{Limitations}
\label{sec:limitation}

HiCCR is designed for RLVR settings where verifiable outcome rewards are available and the model exposes attention statistics that can be efficiently reused during training. While our theory identifies gradient-based saliency as a principled signal for visual dependence, computing it requires an additional backward pass, so HiCCR adopts visual-attention concentration as a forward-only proxy; Table~\ref{tab:table4_credit_assignment} and Appendix~\ref{sec:saliency_proxy} show that this proxy outperforms raw attention energy and task-agnostic heuristics, but richer causal or semantic estimators may further improve credit reassignment. Extending the idea to open-ended rewards, video and audio inputs, or architectures with substantially different visual-token representations may therefore require adapting the credit estimator. Our experiments focus on image-based multimodal reasoning and hallucination-oriented benchmarks, with additional validation on a smaller-data setting and another MLLM family; broader evaluation on more diverse real-world multimodal tasks would further clarify the generality of the approach. Finally, our theoretical analysis focuses on the credit-assignment mechanism itself and abstracts from some engineering details of large-scale RL training; nevertheless, the resulting predictions are consistent with the attention, gradient, and counterfactual diagnostics observed empirically.

\section{Broader Impacts}
\label{sec:broader_impacts}

This work aims to improve the visual grounding of RLVR-trained MLLMs by reducing excessive reliance on language priors when visual evidence is required. It may benefit educational, scientific, and assistive applications where multimodal answers should be supported by image content. However, HiCCR does not guarantee faithful visual understanding, and stronger multimodal reasoning can introduce risks in high-stakes domains such as medical, legal, surveillance, or safety-critical settings. Since RL fine-tuning large models also incurs computational costs, HiCCR should be deployed with task-specific evaluation, privacy safeguards, and appropriate human oversight.

\newpage

\section*{NeurIPS Paper Checklist}

\begin{enumerate}

\item {\bf Claims}
    \item[] Question: Do the main claims made in the abstract and introduction accurately reflect the paper's contributions and scope?
    \item[] Answer: \answerYes{} 
    \item[] Justification: The Abstract and Introduction state the main diagnosis, theoretical explanation, HiCCR design, and empirical claims, which are supported by Figure~\ref{fig:teaser}, Theorems~\ref{thm:grad_bias}--\ref{thm:drift}, and Table~\ref{tab:table1_rlvr}. The claims are scoped to RLVR-trained MLLMs with verifiable rewards, rather than arbitrary multimodal tasks.

\item {\bf Limitations}
    \item[] Question: Does the paper discuss the limitations of the work performed by the authors?
    \item[] Answer: \answerYes{} 
    \item[] Justification: The paper discusses limitations in \S\ref{sec:limitation}.

\item {\bf Theory assumptions and proofs}
    \item[] Question: For each theoretical result, does the paper provide the full set of assumptions and a complete (and correct) proof?
    \item[] Answer: \answerYes{} 
    \item[] Justification: The theoretical results are stated under explicit assumptions in Appendix~\ref{sec:pg_theory}, including language-prior dominance and key-correlation conditions. Theorems~\ref{thm:grad_bias} and~\ref{thm:drift} are accompanied by proof sketches, complete proofs, and empirical verification of central assumptions.

    \item {\bf Experimental result reproducibility}
    \item[] Question: Does the paper fully disclose all the information needed to reproduce the main experimental results of the paper to the extent that it affects the main claims and/or conclusions of the paper (regardless of whether the code and data are provided or not)?
    \item[] Answer: \answerYes{} 
    \item[] Justification: The paper describes the algorithmic objective, credit score, token- and trajectory-level reweighting, implementation hooks, datasets, baselines, benchmarks, hyperparameters, and compute setup in \S\ref{sec:method}, \S\ref{sec:exp_setup}, Appendix~\ref{sec:details_train_eval}, and Appendix~\ref{app:overhead}. These details provide a reasonable path to reproduce the main experimental conclusions.

\item {\bf Open access to data and code}
    \item[] Question: Does the paper provide open access to the data and code, with sufficient instructions to faithfully reproduce the main experimental results, as described in supplemental material?
    \item[] Answer: \answerYes{} 
    \item[] Justification: Code is zipped in supplement files.

\item {\bf Experimental setting/details}
    \item[] Question: Does the paper specify all the training and test details (e.g., data splits, hyperparameters, how they were chosen, type of optimizer) necessary to understand the results?
    \item[] Answer: \answerYes{} 
    \item[] Justification: The paper specifies the training data, benchmarks, evaluation protocol, baselines, optimizer, learning rate, rollout configuration, decoding setup, checkpoint selection, and HiCCR-specific hyperparameters in \S\ref{sec:exp_setup} and Appendix~\ref{sec:details_train_eval}. Additional implementation details for attention hooks and memory-efficient credit computation are also provided.

\item {\bf Experiment statistical significance}
    \item[] Question: Does the paper report error bars suitably and correctly defined or other appropriate information about the statistical significance of the experiments?
    \item[] Answer: \answerNo{} 
    \item[] Justification: Similar to prior large-scale RL studies~\cite{papo, dvrp, beyond28, deepseekr1, avar_narrow}, we do not report error bars due to the high computational cost of repeatedly training all baselines. Instead, we report AVG@8, where each benchmark accuracy is averaged over eight independent forward passes to reduce stochastic decoding variance, and further support the conclusions with ablation and diagnostic analyses as described in \S\ref{sec:exp_setup}. Fair evaluation protocol and metrics are ensured in our experiments.

\item {\bf Experiments compute resources}
    \item[] Question: For each experiment, does the paper provide sufficient information on the computer resources (type of compute workers, memory, time of execution) needed to reproduce the experiments?
    \item[] Answer: \answerYes{} 
    \item[] Justification: The paper reports the hardware and software setup configurations in Appendix~\ref{sec:details_train_eval}. Appendix~\ref{app:overhead} and Table~\ref{tab:compute_overhead} further provide step time, peak memory, wall-clock overhead.
    
\item {\bf Code of ethics}
    \item[] Question: Does the research conducted in the paper conform, in every respect, with the NeurIPS Code of Ethics \url{https://neurips.cc/public/EthicsGuidelines}?
    \item[] Answer: \answerYes{} 
    \item[] Justification: The research uses publicly datasets and complies with the NeurIPS Code of Ethics.

\item {\bf Broader impacts}
    \item[] Question: Does the paper discuss both potential positive societal impacts and negative societal impacts of the work performed?
    \item[] Answer: \answerYes{} 
    \item[] Justification: The paper discusses potential impacts in educational, scientific, and assistive applications in \S\ref{sec:broader_impacts}.
    
\item {\bf Safeguards}
    \item[] Question: Does the paper describe safeguards that have been put in place for responsible release of data or models that have a high risk for misuse (e.g., pre-trained language models, image generators, or scraped datasets)?
    \item[] Answer: \answerNA{} 
    \item[] Justification: The current submission does not release a high-risk pretrained model, image generator, or newly scraped dataset requiring special release safeguards.

\item {\bf Licenses for existing assets}
    \item[] Question: Are the creators or original owners of assets (e.g., code, data, models), used in the paper, properly credited and are the license and terms of use explicitly mentioned and properly respected?
    \item[] Answer: \answerYes{} 
    \item[] Justification: All datasets and pretrained models are cited with their original sources throughout the paper

\item {\bf New assets}
    \item[] Question: Are new assets introduced in the paper well documented and is the documentation provided alongside the assets?
    \item[] Answer: \answerNA{} 
    \item[] Justification: The paper introduces a new algorithmic method, but the current submission does not release a new dataset, benchmark, model checkpoint, or software package as a documented asset.

\item {\bf Crowdsourcing and research with human subjects}
    \item[] Question: For crowdsourcing experiments and research with human subjects, does the paper include the full text of instructions given to participants and screenshots, if applicable, as well as details about compensation (if any)? 
    \item[] Answer: \answerNA{} 
    \item[] Justification: This work does not involve crowdsourcing or human subjects.

\item {\bf Institutional review board (IRB) approvals or equivalent for research with human subjects}
    \item[] Question: Does the paper describe potential risks incurred by study participants, whether such risks were disclosed to the subjects, and whether Institutional Review Board (IRB) approvals (or an equivalent approval/review based on the requirements of your country or institution) were obtained?
    \item[] Answer: \answerNA{} 
    \item[] Justification: This work does not involve human subjects research.

\item {\bf Declaration of LLM usage}
    \item[] Question: Does the paper describe the usage of LLMs if it is an important, original, or non-standard component of the core methods in this research? Note that if the LLM is used only for writing, editing, or formatting purposes and does \emph{not} impact the core methodology, scientific rigor, or originality of the research, declaration is not required.
    \item[] Answer: \answerNA{} 
    \item[] Justification: LLMs are only used for editing.

\end{enumerate}

\end{document}